\documentclass{article}

\usepackage{amsmath,amsfonts,bm}

\def\eqref#1{equation~\ref{#1}}

\def\1{\bm{1}}

\def\rva{{\mathbf{a}}}

\def\rvr{{\mathbf{r}}}
\def\rvs{{\mathbf{s}}}

\def\rmA{{\mathbf{A}}}
\def\rmB{{\mathbf{B}}}

\def\rmR{{\mathbf{R}}}
\def\rmS{{\mathbf{S}}}

\def\rmV{{\mathbf{V}}}

\DeclareMathAlphabet{\mathsfit}{\encodingdefault}{\sfdefault}{m}{sl}
\SetMathAlphabet{\mathsfit}{bold}{\encodingdefault}{\sfdefault}{bx}{n}

\newcommand{\ours}{\textbf{ACE}~}

\usepackage{microtype}
\usepackage{graphicx}
\usepackage{booktabs} % for professional tables

\usepackage{hyperref}

\usepackage[accepted]{icml2024}

\usepackage{mathtools}
\usepackage{amsthm}
\usepackage{dashrule}
\usepackage{amsmath}
\usepackage{bbm}
\usepackage{amsthm}   
\usepackage{amssymb}
\usepackage[utf8]{inputenc} % allow utf-8 input
\usepackage[T1]{fontenc}    % use 8-bit T1 fonts
\usepackage{hyperref}       % hyperlinks
\usepackage{url}            % simple URL typesetting
\usepackage{amsfonts}       % blackboard math symbols
\usepackage{nicefrac}       % compact symbols for 1/2, etc.
\usepackage{microtype}      % microtypography
\usepackage{xcolor}         % colors
\usepackage{tikz}
\usepackage{caption}
\usepackage{subcaption}
\usepackage{wrapfig}
\usepackage{float}
\usepackage{array}
\usepackage{makecell}
\usepackage{enumitem}
\usepackage{multirow}
\usepackage[export]{adjustbox}
\usepackage{mdframed}
\usepackage{colortbl}

\usepackage[capitalize,noabbrev]{cleveref}

\definecolor{mydarkblue}{rgb}{0,0.08,0.45}
\definecolor{myred}{rgb}{0.84,0.17,0.11}
\definecolor{myyellow}{rgb}{0.94,0.71,0.18}
\definecolor{mygreen}{rgb}{0.17,0.63,0.17}
\definecolor{myorange}{rgb}{0.99,0.50,0.05}
\definecolor{myblue}{rgb}{0.12,0.46,0.70}
\definecolor{mypurple}{rgb}{0.58,0.40,0.71}
\definecolor{mylightblue}{rgb}{0.78,0.90, 0.96}
\definecolor{mygray}{rgb}{0.45,0.45, 0.45}
\definecolor{mypink}{HTML}{d8548a}

\usepackage{soul}

\theoremstyle{plain}
\newtheorem{theorem}{Theorem}[section]
\newtheorem{proposition}[theorem]{Proposition}

\theoremstyle{definition}
\newtheorem{definition}[theorem]{Definition}
\newtheorem{assumption}[theorem]{Assumption}
\theoremstyle{remark}

\newcommand{\ourshort}{\textbf{ACE}}
\usepackage[textsize=tiny]{todonotes}

\icmltitlerunning{\ours: Off-Policy \textbf{A}ctor-critic with \textbf{C}ausality-Aware \textbf{E}ntropy Regularization}

\begin{document}
\sloppy{}
\twocolumn[
\icmltitle{\ours: Off-Policy Actor-Critic with Causality-Aware Entropy Regularization}

% It is OKAY to include author information, even for blind
% submissions: the style file will automatically remove it for you
% unless you've provided the [accepted] option to the icml2024
% package.

% List of affiliations: The first argument should be a (short)
% identifier you will use later to specify author affiliations
% Academic affiliations should list Department, University, City, Region, Country
% Industry affiliations should list Company, City, Region, Country

% You can specify symbols, otherwise they are numbered in order.
% Ideally, you should not use this facility. Affiliations will be numbered
% in order of appearance and this is the preferred way.
\icmlsetsymbol{equal}{*}

\begin{icmlauthorlist}
\icmlauthor{Tianying Ji}{equal,thu}
\icmlauthor{Yongyuan Liang}{equal,qizhi,umd}
\icmlauthor{Yan Zeng}{btbu}
\icmlauthor{Yu Luo}{thu}
\icmlauthor{Guowei Xu}{thu}
\icmlauthor{Jiawei Guo}{thu}
\icmlauthor{Ruijie Zheng}{umd}
\icmlauthor{Furong Huang}{umd}
\icmlauthor{Fuchun Sun}{thu}
\icmlauthor{Huazhe Xu}{thu,qizhi}
\end{icmlauthorlist}

\icmlaffiliation{thu}{Tsinghua University}
\icmlaffiliation{umd}{University of Maryland}
\icmlaffiliation{btbu}{Beijing Technology and Businesses University}
\icmlaffiliation{qizhi}{Shanghai Qi Zhi Institute}
% \icmlaffiliation{sch}{School of ZZZ, Institute of WWW, Location, Country}

\icmlcorrespondingauthor{Tianying Ji}{jity20@mails.tsinghua.edu.cn}
\icmlcorrespondingauthor{Yongyuan Liang}{cheryllLiang@outlook.com}

% You may provide any keywords that you
% find helpful for describing your paper; these are used to populate
% the "keywords" metadata in the PDF but will not be shown in the document
\icmlkeywords{Machine Learning, ICML}

\vskip 0.3in
]

% this must go after the closing bracket ] following \twocolumn[ ...

% This command actually creates the footnote in the first column
% listing the affiliations and the copyright notice.
% The command takes one argument, which is text to display at the start of the footnote.
% The \icmlEqualContribution command is standard text for equal contribution.
% Remove it (just {}) if you do not need this facility.

% \printAffiliationsAndNotice{}  % leave blank if no need to mention equal contribution
\printAffiliationsAndNotice{\icmlEqualContribution} % otherwise use the standard text.
% \printAffiliationsAndNotice{\icmlEqualContribution} % otherwise use the standard text.

\begin{abstract}
The varying significance of distinct primitive behaviors during the policy learning process has been overlooked by prior model-free RL algorithms. Leveraging this insight, we explore the causal relationship between different action dimensions and rewards to evaluate the significance of various primitive behaviors during training. We introduce a causality-aware entropy term that effectively identifies and prioritizes actions with high potential impacts for efficient exploration. Furthermore, to prevent excessive focus on specific primitive behaviors, we analyze the gradient dormancy phenomenon and introduce a dormancy-guided reset mechanism to further enhance the efficacy of our method. Our proposed algorithm, \ourshort: Off-policy \textbf{A}ctor-critic with \textbf{C}ausality-aware \textbf{E}ntropy regularization, demonstrates a substantial performance advantage across 29 diverse continuous control tasks spanning 7 domains compared to model-free RL baselines, which underscores the effectiveness, versatility, and efficient sample efficiency of our approach. Benchmark results and videos are available at \url{https://ace-rl.github.io/}.

\end{abstract}

\section{Introduction}
\label{sec:intro}

Reinforcement Learning (RL) has made remarkable strides in addressing complex decision-making problems, ranging from video games~\citep{mnih2013playing, silver2016mastering} to robot control~\citep{trpo, ppo, lee2020learning}.
Despite this, a persistent challenge in RL is high sample complexity, which poses a formidable impediment to the practical application of deep RL in real-world scenarios.
Effective exploration lies in the core of reinforcement learning~(RL) for optimal decision-making as well as sample efficiency~\citep{lopes2012exploration,sutton2018reinforcement,ladosz2022exploration}.  Ineffective exploration would lead to unsatisfactory sample efficiency, as the agent may spend excessive interactions in low-value or irrelevant areas that do not contribute to performance improvement.
\begin{figure}[t]
    \centering
        \includegraphics[width=\linewidth]{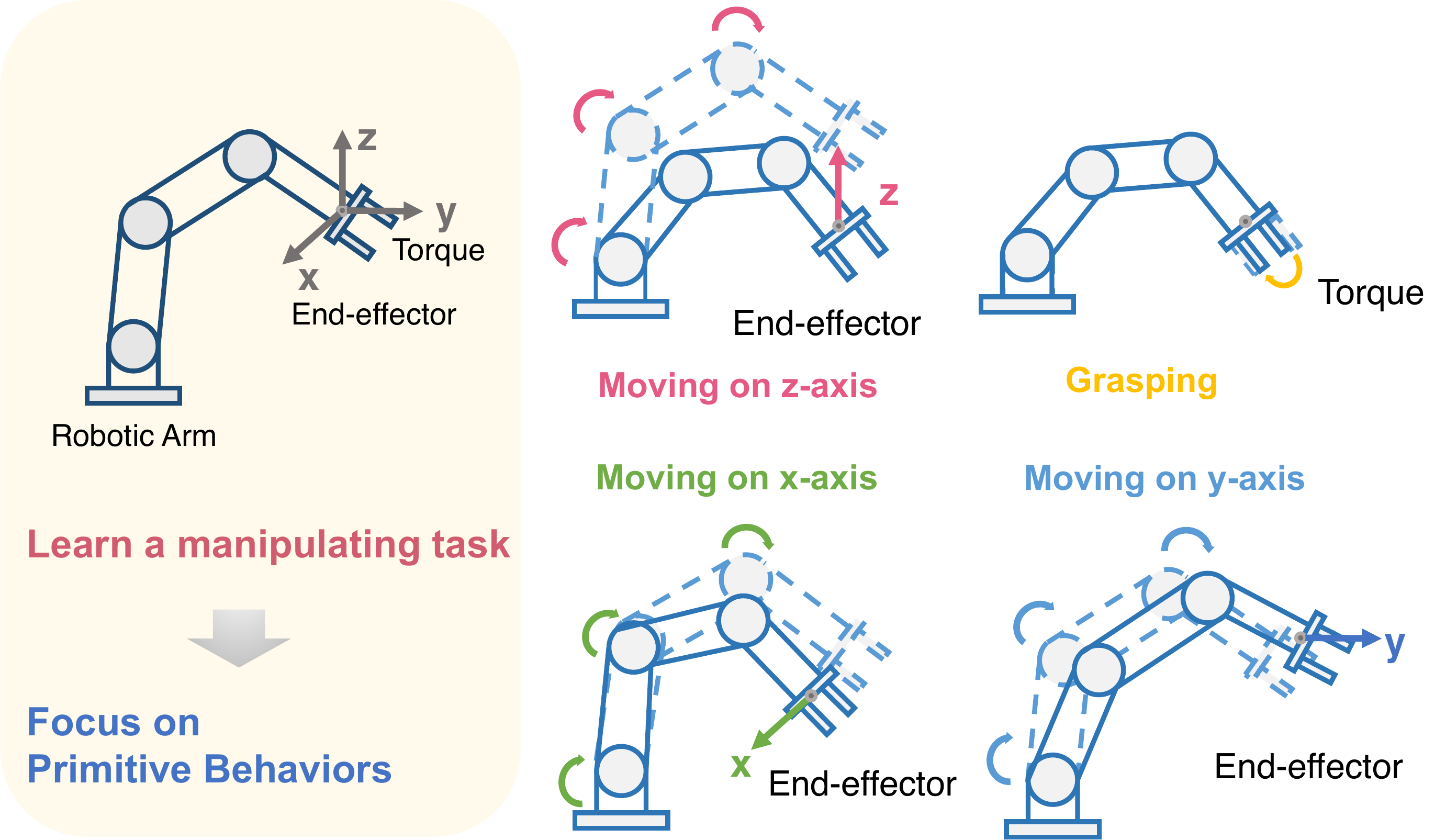}
        \vspace{1mm}
    \textcolor{mygray}{\hdashrule[0.5ex]{\linewidth}{1pt}{5pt 3pt}}
    \vspace{2mm}
        \includegraphics[width=\linewidth]{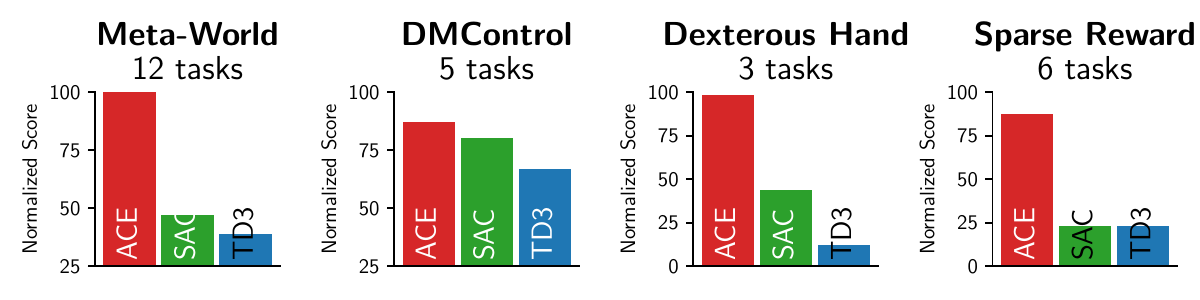}
    \vspace{-2.5em}
    \caption{\small \textit{(Top)}: \textbf{Learning process of a manipulator.} A robotic arm learns to manipulate objects in a manner akin to human learning. This arm would be programmed with four primitive behaviors for its end-effector: vertical movements along the \textcolor{mypink}{z-axis (up and down)}, \textcolor{mygreen}{horizontal movements along the x-axis (left and right)}, \textcolor{myblue}{depth movements along the y-axis (forward and backward)}, and \textcolor{myyellow}{grasping (apply torque)}.  \textit{(Bottom)}: \textbf{Comparison of normalized score.} Our \ours demonstrates a significant superiority over the widely used model-free RL baselines SAC and TD3 with a single set of hyperparameters.}
    \vspace{-1.5em}
    \label{fig:manipulate}
\end{figure}
In order to enhance sample efficiency, previous works have proposed various exploration strategies, such as upper-confidence bounds (UCB) based exploration~\citep{chen2017ucb}, curiosity-driven exploration for sparse reward tasks~\citep{pathak2017curiosity}, random network distillation~\cite{burda2018exploration}, maximum entropy RL~\citep{haarnoja2018soft}, etc. 

Intriguingly, existing exploration methods typically simply aggregate uncertainty across all action dimensions, failing to account for the varying significance of each primitive behavior in the policy optimization process over the course of training. These methods may neglect a fundamental aspect: mastering a motor task often involves multiple stages, each requiring proficiency in different primitive behaviors, similar to how humans learn~\citep{roy2022brain}. Consider a simple example, as shown in Figure~\ref{fig:manipulate}: a manipulator should initially learn to lower its arm and grasp the object, then shift attention to learning the movement direction of the arm towards the final goal. 
Therefore, it is crucial to emphasize the exploration of the most significant primitive behaviors at distinct stages of policy learning. 
The deliberate focus on various primitive behaviors during exploration can accelerate the agent's learning of essential primitive behaviors at each stage, thus improving the efficiency of mastering the complete motor task.

\begin{figure*}[ht]
    \centering
    \includegraphics[width=0.87\textwidth]{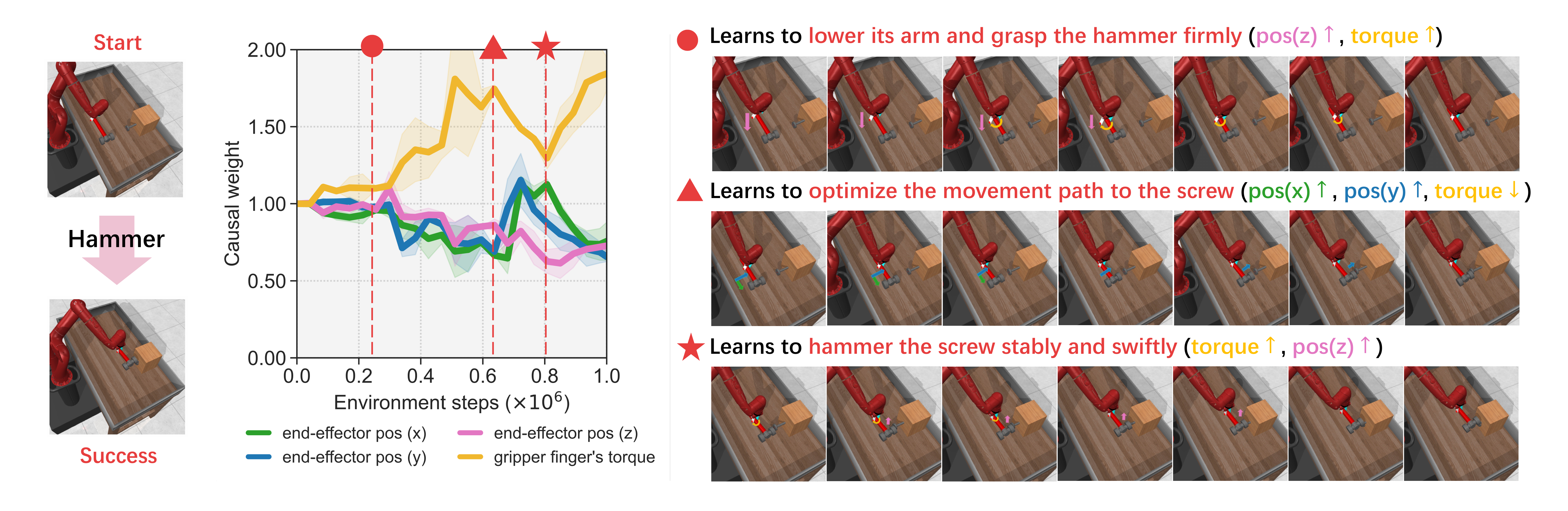}
    \vspace{-1em}
    \caption{\small \textbf{Motivating example.} This task involves a robotic arm hammering a screw into a wall. 
    \textcolor{red}{\large $\bullet$} 
    Initially, the robotic arm approaches the desk moving on the \textcolor{mypink}{z-axis} and struggles with torque \textcolor{myyellow}{grasping}, making z-axis positioning \textcolor{mypink}{$\uparrow$} and torque exploration \textcolor{myyellow}{$\uparrow$} a priority. 
    \textcolor{red}{$\blacktriangle$} As the training advances, the agent's focus shifts to optimizing movement, prioritizing end-effector position (\textcolor{mygreen}{x-axis $\uparrow$} and \textcolor{myblue}{y-axis $\uparrow$}). \textcolor{red}{$\bigstar$} Finally, potential improvements lie in the stable and swift hammering, shifting focus back to torque \textcolor{myyellow}{$\uparrow$} and placing down the object \textcolor{mypink}{$\uparrow$}. The evolving causal weights, depicted on the left, reflect these changing priorities. See more examples in Appendix~\ref{sec:example}.}
    \label{fig:motivation}
    \vspace{-3mm}
\end{figure*}

\textit{How can we identify the most crucial primitive behaviors at each stage of policy learning?} To answer this question, we evaluate the significance of individual primitive behaviors by quantifying their impact on rewards through the analysis of causal relationships. Within each task, the action spaces of RL agents comprise dimensions representing individual primitive behaviors—such as the 3D positioning of the end-effector and manipulation of gripper fingers. When a particular action dimension demonstrates a larger influence on rewards, it indicates its higher importance in the ongoing learning stage, and vice versa.
We introduce a causal policy-reward structural model to compute the causal weights on action spaces and provide theoretical analyses to ensure the identifiability of the causal structure. In Figure~\ref{fig:motivation}, we illustrate the changes in causal weights on four action dimensions in a manipulator task, contrasting them with the agent's behaviors at different time points. This effectively elucidates how the agent's focus on different primitive behaviors changes during the learning process, as reflected in the causal weights.

Therefore, causal weights naturally guide agents to conduct more efficient exploration, encouraging exploration for action dimensions with larger causal weights, indicating greater significance on rewards, and reducing exploration for those with smaller causal weights. Our methodology builds on the maximum entropy framework~\citep{haarnoja2018soft, ziebart2008maximum, zhao2019maximum}, incorporating a regularization term for policy entropy in the objective function. The general maximum entropy objective lacks awareness of the importance of distinctions between primitive behaviors at different learning stages, potentially resulting in inefficient exploration. To address this limitation, we introduce a policy entropy weighted by causal weights as a causality-aware entropy maximization objective, effectively strengthening the exploration of significant primitive behaviors and leading to more efficient exploration.

To mitigate the risk of overfitting due to excessive focus on certain behaviors, we introduce a gradient-dormancy-based reset mechanism based on our analysis of the gradient dormancy phenomenon. This mechanism intermittently perturbs the agent's neural networks with a factor determined by the gradient dormancy degree. The integration of causality-aware exploration with this novel reset mechanism aims to facilitate more efficient and effective exploration, ultimately enhancing the agent's overall performance.

We conduct a comprehensive evaluation of our method across 28 diverse continuous control tasks, spanning 7 domains, including tabletop manipulation~\citep{metaworld, panda}, locomotion control~\citep{mujoco, dmc, robel}, and dexterous hand manipulation tasks~\citep{adroit, shadow}, covering a broad spectrum of task difficulties including sparse reward tasks. In comparison with popular model-free RL algorithms - TD3~\citep{td3} and SAC~\citep{haarnoja2018soft} and exploration method RND~\citep{burda2018exploration}, our results consistently demonstrate that \ours outperforms all the baselines across various task types. Specifically, our method achieves performance improvements of \textit{2.1x} on highly challenging manipulator tasks, \textit{1.1x} on locomotion tasks, \textit{2.2x} on dexterous hand tasks, and \textit{3.7x} on tasks with sparse rewards as shown in Figure~\ref{fig:manipulate}. Our work introduces a novel perspective to RL by deconstructing the learning process and analyzing the varying significance of primitive behaviors over the course of policy optimization.  Our method leverages these insights to enhance sample efficiency via efficient exploration, and we empirically validate its effectiveness through an extensive set of experiments.

Our \textbf{contributions} can be summarized as follows:
\begin{itemize}[leftmargin=10pt]
\vspace{-1em}
\item We propose a causal policy-reward structural model to calculate 
the impact on rewards from different primitive behaviors. 
This provides insight that agents purposefully focus on various primitive behaviors during RL training.
\vspace{-0.5em}
\item We introduce a technology that employs causal weights on policy entropy as a causality-aware entropy objective to enhance exploration efficiency.
\vspace{-0.5em}
\item Additionally, we analyze the gradient dormancy phenomenon and propose a gradient-dormancy-guided reset mechanism to prevent agents from excessively overfitting to certain behaviors.
\vspace{-0.5em}
\item Extensive experiments across a wide range of tasks and domains consistently achieve superior performance against model-free RL baselines with a single
set of hyperparameters, showing the effectiveness of our method. 
Our work offers valuable insights into enhancing sample efficiency more systematically by focusing on the efficient exploration of significant primitive behaviors.
\end{itemize}
\section{Preliminary}
\label{sec:prelim}
\noindent\textbf{Reinforcement Learning~(RL).}\quad
Within the standard framework of the Markov decision process (MDP), RL can be formulated as $\mathcal{M}=\langle\mathcal{S}, \mathcal{A}, \mathcal{P}, \mathcal{R}, \gamma\rangle$. Here, ${\cal S}$  denotes the state space, ${\cal A}$  the action space,  $r: {\cal S\times A}\in [-R_{max}, R_{max}]$  the reward function, and $\gamma\in (0,1)$ the discount factor, and $P(\cdot \mid s,a)$ stands for transition dynamics. 
The objective of an RL agent is to learn an optimal policy $\pi$ that maximizes the expected discounted sum of rewards, formulated as $\mathbb{E}_\pi[\sum_{t=0}^{\infty}\gamma^t r_t]$. For any $s\in\mathcal{S}$ and action $a\in\mathcal{A}$, the value of action $a$ under state $s$ is given by the action-value function $Q^\pi(s,a)=\mathbb{E}[\sum_{t=0}^\infty\gamma^t\mathcal{R}(s_t,a_t)]$. In deep RL, the policy $\pi$ and the value function $Q$ are represented by neural network function approximations.

\noindent\textbf{Soft Actor-Critic~(SAC).}\quad
Soft actor-critic~(SAC) is a popular off-policy maximum entropy deep reinforcement learning algorithm based on soft policy iteration that maximizes the entropy-augmented policy objective function.
Its objective includes a policy entropy regularization term in the objective function with the aim of performing more diverse actions for each given state and visiting states with higher entropy for better exploration, stated as below,
\begin{equation*}
J(\pi) = \sum_{t=0}^{\infty}\mathbb{E}_{(\rvs_t, \rva_t)\sim \rho(\pi)} \left[ \gamma^t (r(\rvs_t, \rva_t) + \alpha \mathcal{H}(\pi(\cdot\vert \rvs_t)))\right]
\end{equation*}

\section{Off-policy Actor-critic with Causality-aware Entropy regularization}
\label{sec:method}

\paragraph{Overview.}
Our approach builds upon maximum entropy RL~\citep{haarnoja2018soft}, with the primary aim of enhancing the sample efficiency of off-policy RL algorithms. Initially, we construct a causal policy-reward structural model, providing a theoretical guarantee for its identifiability. Subsequently, based on this causal model, we introduce causality-aware entropy to present a causality-aware variant of SAC named CausalSAC. To address potential overfitting and further enhance exploration efficiency, we analyze the gradient dormancy phenomenon and propose a gradient-dormancy-guided reset mechanism. The integration of this reset mechanism with CausalSAC constitutes our proposed \ourshort: Off-policy \textbf{A}ctor-critic with \textbf{C}ausality-aware \textbf{E}ntropy regularization algorithm.

\subsection{Causal Discovery on Policy-Reward Relationship}
To explore the causal relationships between each action dimension $\rva_i$ and its potential impact on reward gains $r$,  we ﬁrst establish a causal policy-reward structural model and provide theoretical analyses to ensure the identifiability of the causal structure.

% \noindent\textbf{Causal Policy-Reward Structural Modeling.}\quad
% \paragraph{Causal policy-reward structural modeling.}
\noindent\textbf{Causal policy-reward structural modeling.}\quad
Suppose we have sequences of observations $\{\rvs_t, \rva_t, r_t\}_{t=1}^{T}$, where $\rvs_t = (s_{1,t}, ..., s_{\mathrm{dim}\mathcal{S},t})^T \subseteq {\cal S}$ denote the perceived $\mathrm{dim}\mathcal{S}$-dimensional states at time $t$, $\rva_t = (a_{1,t}, ..., a_{\mathrm{dim}\mathcal{A},t})^T \subseteq {\cal A}$ are the executed $\mathrm{dim}\mathcal{A}$-dimensional actions and $r_t $ is the reward. 
Note that the reward variable $r_t$ may not be influenced by every dimension of $\rvs_t$ or $\rva_t$, and there are causal structural relationships between $\rvs_t$, $\rva_t$ and $r_t$~\citep{huang2022action}. 
To integrate such relationships in MDP, we explicitly encode the causal structures over variables into the reward function
\begin{equation} \label{eq:r_func}
    r_t = r_{\mathcal{M}}\left( \rmB_{{\rvs\to r|\rva}} \odot \rvs_t,  \textcolor{myred}{\rmB_{\rva\to r | \rvs}} \odot \rva_t, \epsilon_t\right),
\end{equation}
where $\rmB_{{\rvs\to r|\rva}} \in \mathbb{R}^{\mathrm{dim}\mathcal{S}\times 1}$ and $\rmB_{\rva\to r|\rvs} \in \mathbb{R}^{\mathrm{dim}\mathcal{A}\times 1} $ are vectors that represent the graph structure~\footnote{Please note that $\rmB_{\cdot \to \cdot}$ encodes information of both causal directions and causal effects. For example, ${B}_{\rva\to r}^{i} = 0$ means there is no edge between $a_{i,t}$ and $r_t$; and ${B}_{\rva\to r}^{i}=c$ implies that $a_{i,t}$ causally influences $r_t$ with effects $c$. Causal effects are called causal weights as well in this paper.} from $\rvs_t$ to $r_{t}$ given $\rva_t$ and from $\rva_t$ to $r_{t}$ given $\rvs_t$, respectively. Here $\odot$ denotes the element-wise product while $\epsilon_t$ are i.i.d. noise terms.

Specifically, under the causal Markov condition and faithfulness assumption~\citep{pearl2009causality}, we establish conditions for the causal relationship existence in Proposition~\ref{prop:noind}, then the true causal graph $\rmB_{\rva \to r|\rvs}$ could be identified from observational data alone, as guaranteed in Theorem~\ref{theo:ident}.
\begin{assumption}[Global Markov Condition~\citep{spirtes2000causation,pearl2009causality}]
    The distribution $p$ over a set of variables $\rmV = (s_{1,t}, ..., s_{\mathrm{dim}\mathcal{S},t}, a_{1,t}, ..., a_{\mathrm{dim}\mathcal{A},t}, r_{t})^T$ satisfies the global Markov condition on the graph if for any partition $(\rmS, \rmA, \rmR)$ in $\rmV$ such that if $\rmA$ d-separates $\rmS$ from $\rmR$, then $p(\rmS, \rmR|\rmA) = p(\rmS|\rmA)p(\rmR|\rmA)$.
\end{assumption}

\begin{assumption}[Faithfulness Assumption~\citep{spirtes2000causation,pearl2009causality}] 
    For a set of variables $\rmV = (s_{1,t}, ..., s_{\mathrm{dim}\mathcal{S},t}, a_{1,t}, ..., a_{\mathrm{dim}\mathcal{A},t}, r_{t})^T$, 
    there are no independencies between variables that are not entailed by the Markovian Condition.
\end{assumption}

With these two assumptions, we provide the following proposition to characterize the condition of the causal relationship existence so that we are able to uncover those key actions from conditional independence relationships.  

\begin{proposition} \label{prop:noind}
    Under the assumptions that the causal graph is Markov and faithful to the observations, there exists an edge from $a_{i,t}$ to $r_t$ if and only if $a_{i,t} \not\!\perp\!\!\!\perp r_t | \rvs_{t}, \rva_{-i, t}$, where $\rva_{-i, t}$ are states of $\rva_t$ except $a_{i, t}$.  
\end{proposition}

We next provide the theorem to guarantee the identifiability of the proposed causal structure.

\begin{theorem} \label{theo:ident}
    Suppose $s_t$, $a_t$, and $r_t$ follow the MDP model with no unobserved confounders, as in Eq.(\ref{eq:r_func}). Under the Markov condition and faithfulness assumption, the structural vectors $\rmB_{\rva \to r|\rvs}$ are identifiable.
\end{theorem}

Note that such a theorem guarantees the identifiability of the correct graph in an asymptotic manner. Additionally, by imposing further assumptions on the data generation mechanism, we could uniquely identify the causal effects. See the proof in Appendix~\ref{app:proof} for details.

\subsection{Causality-aware Bellman Operator}
By infusing the explainable causal weights $\rmB_{\rva \to r|\rvs}$ into policy entropy, we propose the causality-aware entropy $\mathcal{H}_c$ for enhanced exploration. $\mathcal{H}_c$ is defined as 
\begin{equation}
    \begin{aligned}
\mathcal{H}_c(\pi(\cdot\vert \rvs)) &= -\mathbb{E}_{\rva\in \mathcal{A}}\left[\sum\limits_{i=1}^{\mathrm{dim}\mathcal{A}}\rmB_{a_i \to r|\rvs}\pi(a_i\vert \rvs) \log \pi(a_i\vert \rvs) \right], 
 \\ & \quad \rva = (a_1, \ldots, a_{\mathrm{dim}\mathcal{A}}).
\label{eq:causal-policy-reward-entropy}
    \end{aligned}
\end{equation}

Based on the causality-aware entropy, then the $Q$-value for a fixed policy $\pi$ could be computed iteratively by applying a modified Bellman operator $\mathcal{T}_c^{\pi}$ with $\mathcal{H}_c(\pi(\cdot\vert \rvs))$ term as stated below,  
\begin{equation}
\begin{aligned}
    \mathcal{T}_c^{\pi} Q(\rvs_t, \rva_t) \triangleq & r(\rvs_t, \rva_t) + \gamma \mathbb{E}_{\rvs_{t+1}\sim P}[\mathbb{E}_{\rva_t\sim \pi}[Q(\rvs_{t+1}, \rva_{t+1})
    \\
    &+ \alpha\textcolor{myred}{\mathcal{H}_c(\pi(\rva_{t+1}\vert \rvs_{t+1}))}]]. 
\end{aligned}
\end{equation}
For a better understanding of our operator, we conduct a theoretical analysis of its dynamic programming properties in the tabular MDP setting, covering policy evaluation, policy improvement, and policy iteration. All proofs are included in Appendix~\ref{sec:proof-adp-properties}.
\begin{proposition}[Policy evaluation]
Consider an initial $Q_0:\mathcal{S}\times\mathcal{A}\rightarrow\mathbb{R}$ with $\vert \mathcal{A}\vert < \infty$, and $Q$-vlaue  iterates by $Q_{k+1}= \mathcal{T}_c^{\pi}Q_{k}$. Then the sequence $\{Q_{k}\}$ converges to a fixed point $Q^{\pi}$ as $k\rightarrow \infty$.
\end{proposition}

\begin{proposition}[Policy improvement]\label{policy-improvement}
Let $\pi_{k}$ be the policy at iteration $k$, and $\pi_{k+1}$ be the updated policy~( maximize of the $Q$-value). 
Then for all $(s,a)\in \mathcal{S}\times \mathcal{A}$, $\vert \mathcal{A}\vert < \infty$, we have $Q^{\pi_{k+1}}(\rvs,\rva) \geq Q^{\pi_k}(\rvs,\rva)$.
\end{proposition}

\begin{proposition}
[Policy iteration]
Assume $\vert \mathcal{A}\vert < \infty$, by repeating iterations of the policy evaluation and policy improvement, any initial policy converge to the optimal policy $\pi^*$, s.t. $Q^{\pi^*}(\rvs_t,\rva_t)\geq Q^{\pi}(\rvs_t,\rva_t), \forall \pi\in \Pi, \forall (\rvs_t,\rva_t)$.
\end{proposition}

\noindent\textbf{Causality-aware off-policy actor-critic~(CausalSAC).}\quad
% \paragraph{Causality-aware off-policy actor-critic~(CausalSAC).} 
Our causality-aware entropy provides a flexible solution that can be seamlessly incorporated into any Max-Entropy RL framework. For example, as a plug-and-play component, an algorithm instantiation CausalSAC can be implemented within SAC~\citep{haarnoja2018soft} by integrating our $\mathcal{H}_c$ into the policy optimization objective,  $J(\pi) = \sum_{t=0}^{\infty}\mathbb{E}_{(\rvs_t, \rva_t)\sim \rho(\pi)} \left[ \gamma^t (r(\rvs_t, \rva_t) + \alpha \mathcal{H}_c(\pi(\cdot\vert \rvs_t))\right]$.

\subsection{Gradient-dormancy-guided Reset}

Guided by causality-aware entropy, the agent efficiently explores and masters primitive behaviors for different learning stages. However, causality-aware exploration introduces the risk of getting stuck in local optima and overfitting to specific primitive behaviors. To address this challenge, we analyze the gradient dormancy phenomenon during RL training and introduce a soft reset mechanism. This mechanism, guided by gradient dormancy, regularly perturbs the agents' neural networks to maintain network expressivity, thereby improving the agent’s performance. 

The dormancy phenomenon of neural networks in RL, signifying a loss of expressive capacity, has been previously discussed in existing works~\citep{dormant, drm}. However, the dormant phenomenon defined in these works is not evident in state-based RL, and it cannot effectively enhance our algorithm in exploration scheduling. We are the first to investigate dormancy from the perspective of gradients. Here, we introduce definitions for gradient-dormant neurons and the gradient dormancy degree of a neural network.

\begin{definition}[Gradient-dormant Neurons]
For a fully connected layer in a neural network, where $N^l$ represents the number of neurons in layer $l$, the L2 norm of gradients of the weights for neuron $i$ is denoted as $n_{i}^l$. Neuron $i$ is classified as a gradient-dormant neuron if it satisfies
\begin{equation}
\frac{n_i^l(x)}{\frac{1}{N^l}\sum_{k \in l} n_k^l} \leq \tau,
\end{equation}
where $\tau$ is a constant serving as a threshold to determine the gradient dormancy of neurons in each layer.
\end{definition}
\begin{definition}[$\tau$-Dormancy Degree $\alpha_\tau$]
Denote the number of all neurons in the neural network identified as gradient-dormant neurons as $N_{\tau}^l$. The $\alpha_\tau$ for the neural network is defined as:
\begin{equation}
\alpha_\tau = \frac{\sum_{l \in \phi} N_{\tau}^l}{\sum_{l \in \phi} N^l}.
\end{equation}
The $\tau$-dormancy degree $\alpha_\tau$ indicates the percentage of gradient-dormant neurons with the $\tau$ threshold in the fully-connected neural network.
\end{definition}

\begin{figure*}[t]
    \centering
    \begin{subfigure}[b]{0.49\textwidth}
        \includegraphics[width=0.9\columnwidth]{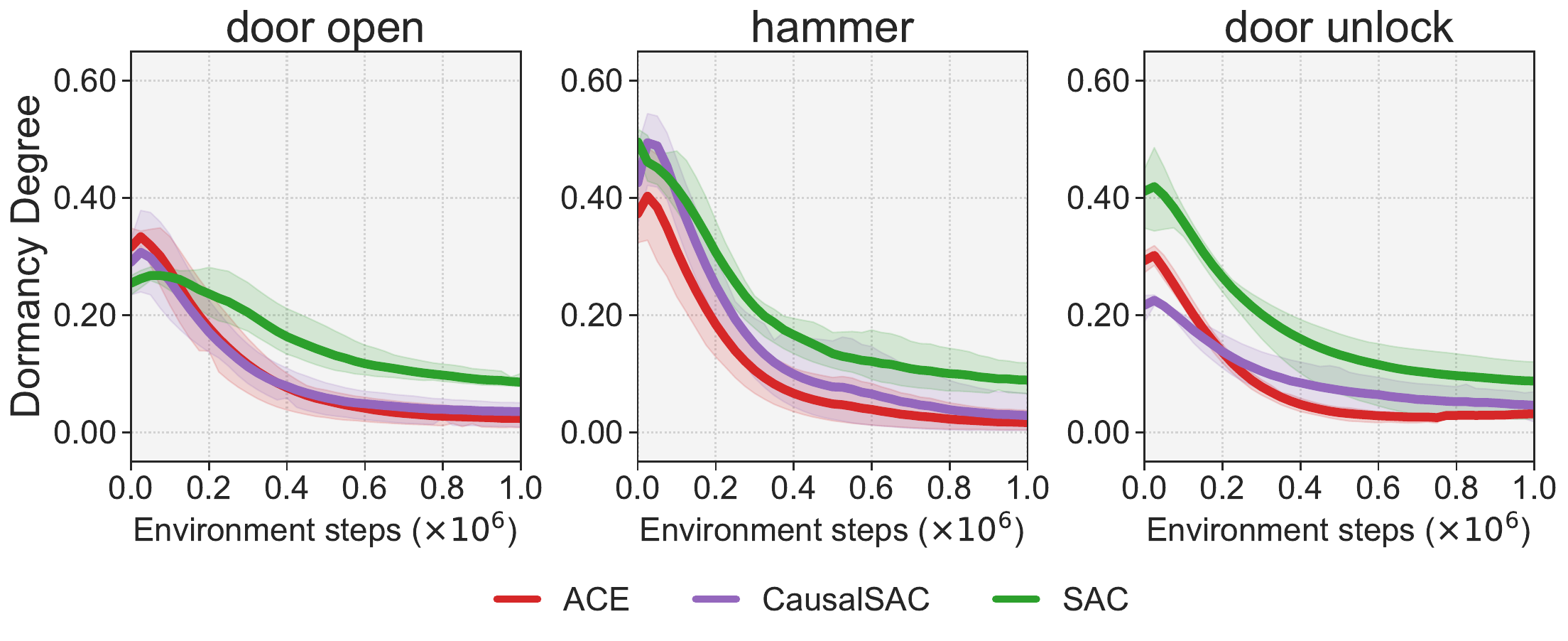}
        \caption{\small In simpler tasks, CausalSAC performs adequately, with observed gradient dormancy degrees decreasing to near-zero. This indicates that in less complex environments, the issue of gradient dormancy becomes negligible.}
    \end{subfigure}
    \hfill
    \begin{subfigure}[b]{0.49\textwidth}
        \includegraphics[width=0.9\columnwidth]{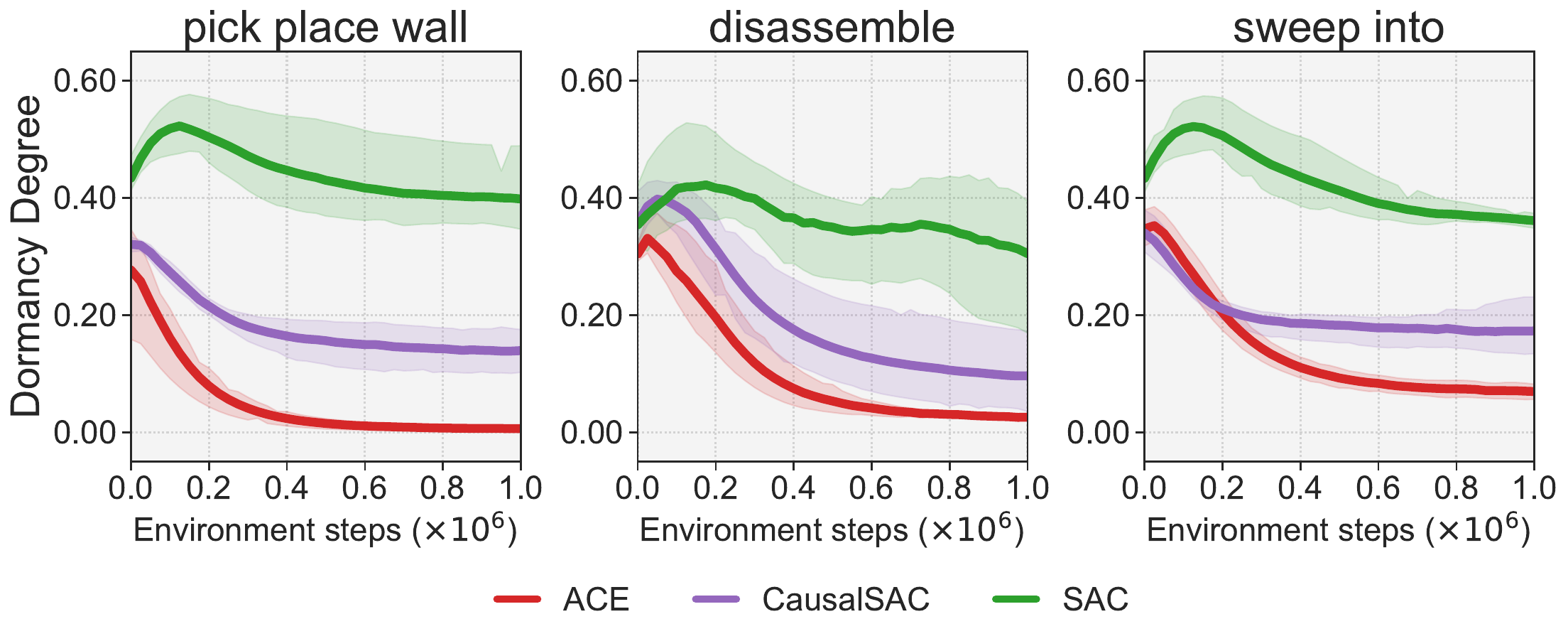}
        \caption{\small In challenging tasks, both SAC and CausalSAC exhibit high dormancy degrees without achieving expected performance. The gradient dormancy-guided reset mechanism in ACE, however, effectively lowers the gradient dormancy degree.}
    \end{subfigure}
    \caption{\small Dormancy degree curves for SAC, CausalSAC, and \ours in MetaWorld tasks, which indicates that the gradient-dormancy-guided reset mechanism effectively reduces gradient dormancy degrees, contributing to the best performance of \ours.}
    \label{fig: dormancy}
    \vspace{-0.5em}
\end{figure*}

In Figure~\ref{fig: dormancy}, we depict the occurrence of the gradient dormancy phenomenon at the initial learning stages of SAC. However, in algorithms with superior sample efficiency like CausalSAC and \ours, there is a notable reduction in the dormancy degree during training. In tasks that CausalSAC can effectively solve, the gradient dormancy degree can decrease to near-zero levels, as shown in Figure~\ref{fig: dormancy}. Particularly, when addressing challenging tasks such as pick place walls, which were previously unsolved by existing baselines, dormancy degrees persist at high levels without reset interventions in SAC and CausalSAC. 
Hence, we speculate that this may represent a potential local optimum for causal-aware exploration.
We consider that dormancy degrees may impact sample efficiency and employ a soft reset method~\citep{drm, ash2020warm} to further decrease dormancy degree by periodically perturbing the policy network and critic network with a reset factor $\eta$, representing the magnitude of weight resetting:
\begin{equation}
\vspace{-0.6em}
\theta_t=(1-\eta) \theta_{t-1} + \eta \phi_{i},\ \phi_{i}\sim\text{initializer}.
\end{equation}
Intuitively, a higher dormancy degree should correspond to a more substantial degree of weight refresh. The value of $\eta$ is determined by the gradient-dormancy degree $\alpha_\tau$ and regulated as $\eta=\text{clip}(\alpha_\tau, 0, \eta_{\text{max}})$, with $\eta_{\text{max}} \leq 1$ as a constant. 
Compared with prior works~\citep{dormant, drm}, our gradient-dormancy-guided reset method employs minimal hyperparameters and is highly adaptable in different RL settings, as analyzed in Appendix~\ref{app:drm}.

\subsection{Algorithm instantiation}
Combining the causality-aware entropy regularization and gradient-dormancy-guided reset mechanism, we propose our algorithm \ours: off-policy \textbf{A}ctor-critic with \textbf{C}ausality-aware \textbf{E}ntropy regularization. The pseudocode and further implementation details are provided in Appendix~\ref{sec:practicalimplementation}.

Instantiating \ourshort\ involves specifying three main components: 1) effectively recognizing the causal weights of $\rva\rightarrow\rvr \vert \rvs$; 2) incorporating causal weights and the corresponding causality-aware entropy term into policy optimization. 3) periodically resetting the network based on our gradient dormancy degree.

To effectively compute $\mathbb{B}_{a_i\rightarrow r\vert s}$, we adopt the well-regarded DirectLiNGAM~\citep{shimizu2011directlingam} method. The main implementation idea of training DirectLiNGAM is as follows. In the first phase, it estimates a causal ordering for all variables of interest (i.e., state, action, and reward variables) based on the independence and non-Gaussianity characteristics of the root variable. The causal ordering is a sequence that implies the latter variable cannot cause the former one. In the second phase, DirectLiNGAM estimates the causal effects between variables using some conventional covariance-based methods. Besides, we formulate a training regime wherein we iteratively adjust the causal weights for the policy at regular intervals $I$ on a local buffer $\mathcal{D}_c$ with fresh transitions to reduce computation cost.

Given the causal weight matrix $\rmB_{\rva\rightarrow\rvr \vert \rvs}$, we could obtain the causality-aware entropy $\mathcal{H}_c(\pi(\cdot\vert \rvs))$
through Eq.(\ref{eq:causal-policy-reward-entropy}).
Based on the causality-aware entropy, then the $Q$-value for a fixed policy $\pi$ could be computed iteratively by applying $\mathcal{T}_c^{\pi}$. Based on the policy evaluation, we can adopt many off-the-shelf policy optimization oracles; we chose SAC as the backbone technique primarily for its simplicity in our primary implementation of CausalSAC and \ourshort.

For each reset interval, we calculate the gradient dormancy degree, initialize a random network with weights $\phi_i$, and soft reset the policy network $\pi_\theta$ and the Q network $Q_{\phi}$. 

In the next section, we empirically substantiate the effectiveness and efficiency of our proposed causality-aware actor-critic, coupled with the gradient-dormancy-guided reset mechanism, through an extensive array of experiments.

\section{Experiments}
\label{sec:exp}
Our experiments aim to investigate the following questions:
1) How effective is the proposed \ours in diverse continuous control tasks, spanning locomotion and manipulation skills and covering both sparse and dense reward settings, compared to model-free RL baselines?
2) What role does each component of our method play in achieving the final performance?
3) How do hyperparameters and the use of different causal inference methods impact our results?

\subsection{Evaluation on various benchmark suites}
We evaluate \ours across 29 diverse continuous control tasks spanning 7 task domains using a single set of hyperparameters: \textbf{MuJoCo}~\citep{mujoco}, \textbf{MetaWorld}~\citep{metaworld}, \textbf{Deepmind Control Suite}~\citep{dmc}, \textbf{Adroit}~\citep{adroit}, \textbf{Shadow Dexterous Hand}~\citep{shadow}, \textbf{Panda-gym}~\citep{panda}, and \textbf{ROBEL}~\citep{dkitty}.Our experimental tasks include high-dimensional state and action spaces, sparse rewards, multi-object manipulation, and locomotion for diverse embodiments, and cover a broad spectrum of task difficulties. \ours\ can be applied to a variety continuous control problems covering high-dimensional state and action spaces, sparse rewards, multi-object manipulation, and locomotion for diverse embodiments, without any need for hyperparameter-tuning.

\noindent\textbf{Baselines.} \quad
We compare our method with three popular data-efficient model-free RL baselines on all the tasks and an efficient exploration method on tasks with sparse rewards: 
1) Soft Actor-Critic (SAC)~\citep{haarnoja2018soft}, a model-free off-policy actor-critic algorithm with maximum entropy regularization.
2) Twin Delayed DDPG (TD3)~\citep{td3}: An advanced version of DDPG, which trains a deterministic policy in an off-policy way, incorporating Double Q-learning tricks.
3) Random Network Distillation (RND)~\citep{burda2018exploration}, an efficient exploration bonus based on the error of predicting features of observations, which is particularly effective for sparse reward tasks.

\begin{figure*}[t]
	\centering
     \includegraphics[width=14cm,keepaspectratio]{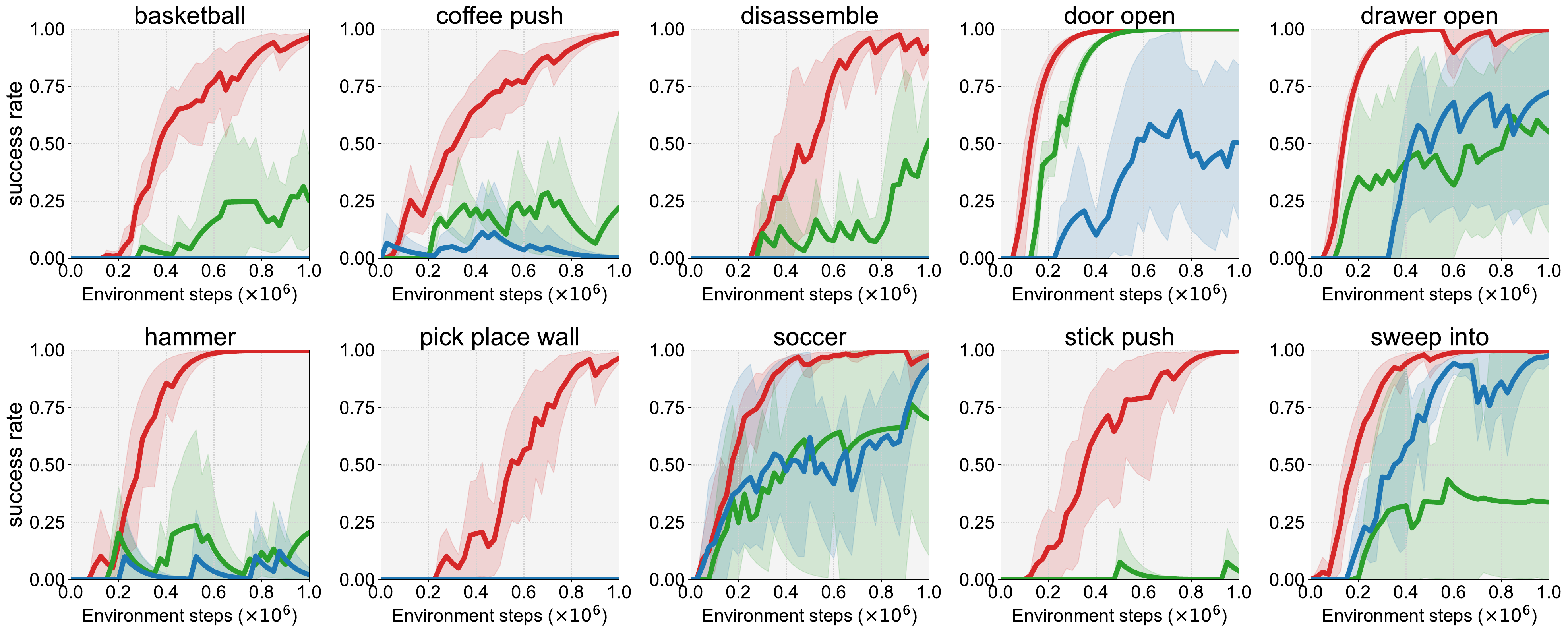}
    \\
    \includegraphics[height=0.5cm,keepaspectratio]{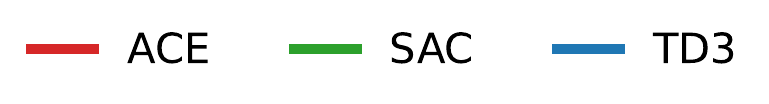}
 \vspace{-0.8em}
 \caption{\small \textbf{Manipulation tasks.} Success rate of \ourshort, SAC, TD3 on manipulation tasks from the MetaWorld benchmark suite. Solid curves depict the mean of six trials, and shaded regions correspond to the one standard deviation. More results are in Appendix Figure~\ref{fig:metaworld}.}
 \vspace{-0.5em}
 \label{fig:metaworld-manipulation}
\end{figure*}

We also provide a comparison with BAC~\citep{bee}, which leverages the value of past successes to enhance Q-value estimation and policy learning, in Appendix~\ref{section:metaworld_benchmark}.  Additionally, we integrate our causality-aware entropy and reset mechanism into the BAC algorithm and find that the ad-hoc ACE-BAC also outperforms the original BAC, as shown in Appendix~\ref{section:shadowhand_benchmark}, which further showcases the generalizability of our method. Detailed performance curves on different benchmark suites are provided in Appendix~\ref{app:morebenchmarkresults}.

\noindent\textbf{Tabletop Manipulation.}\quad
We conducted experiments on tabletop manipulation tasks in MetaWorld, tackling 14 tasks with dense rewards, spanning 4 very hard, 7 hard, and 3 medium tasks, including all types of tasks and all levels of task difficulties, as shown in Figure~\ref{fig:metaworld-manipulation}. Notably, \ours\ exhibited a substantial lead of over 70\% in very hard tasks, coupled with noteworthy performance improvements exceeding 30\% in hard and medium tasks. Traditional model-free RL baselines often struggle to accomplish very hard tasks like pick-place-wall, stick push and disassemble. In contrast, our approach not only demonstrates superior learning efficiency but also achieves a flawless 100\% success rate in these challenging tasks. These results prominently highlight the high sample efficiency of \ours\ in tabletop manipulation tasks, underscoring the crucial role of causality-aware entropy in improving sample efficiency, along with the positive impact of our reset mechanism on effective exploration, particularly in challenging exploration tasks.

\begin{figure*}[h]
    \centering
    \includegraphics[width=14cm,keepaspectratio]{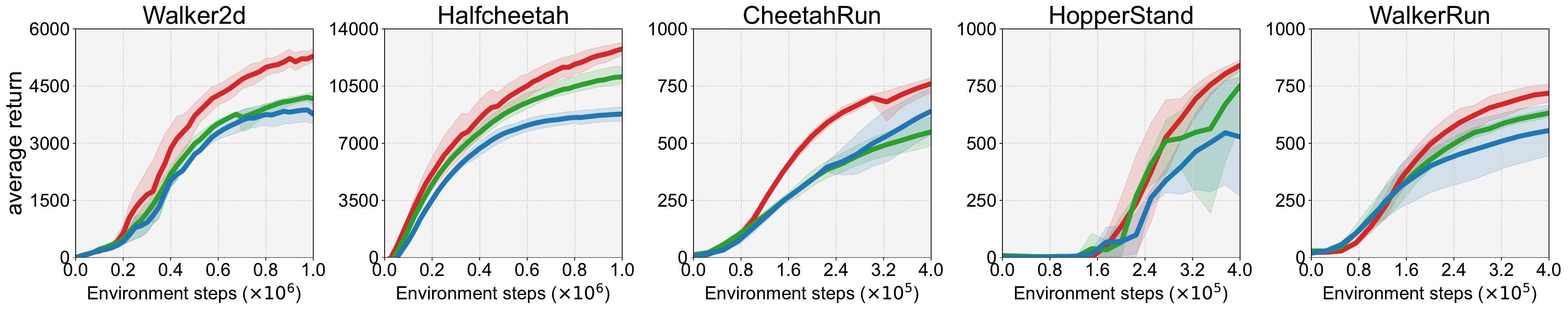}
    \\
    \includegraphics[height=0.55cm,keepaspectratio]{icml2024/figures/main_legend.pdf}
    \vspace{-0.8em}
    \caption{\small \textbf{Locomotion tasks.}  Average return of \ourshort\, SAC, TD3 on locomotion tasks provided by MuJoCo and DMControl benchmark suites. Solid curves depict the mean of six trials, and shaded regions correspond to the one standard deviation. See Figure~\ref{fig:mujoco} and \ref{fig:dmcontrol} in the Appendix for an overall comparison of locomotion tasks.}
    \vspace{-1.0em}
    \label{fig:locomotion}
\end{figure*}

\noindent\textbf{Locomotion.}\quad
Another important task category involves locomotion. We conducted experiments on four MuJoCo tasks and five DeepMind Control Suite tasks, encompassing diverse embodiments as presented in Figure~\ref{fig:locomotion}. Our algorithm achieves state-of-the-art performance and demonstrates improvements across all locomotion tasks, with particularly notable advancements in the Walker2d and HalfCheetah tasks. 
It is noteworthy to observe that, due to the relatively smaller total number of training steps required for locomotion tasks, the gradient dormancy phenomenon is less pronounced compared to manipulation tasks. Consequently, our reset mechanism may not exhibit its optimal effect. To provide further evaluation, we include a detailed comparison between our \ours\, CausalSAC and baselines in Figure~\ref{fig:locomotion_ace_cac_compare}. The results underscore the significant performance enhancement that causality-aware entropy brings to off-policy RL in locomotion tasks.

\begin{figure*}[!h]
\vspace{-2mm}
    \begin{minipage}[t]{0.41\textwidth}
        \centering
        \includegraphics[width=6cm,keepaspectratio]{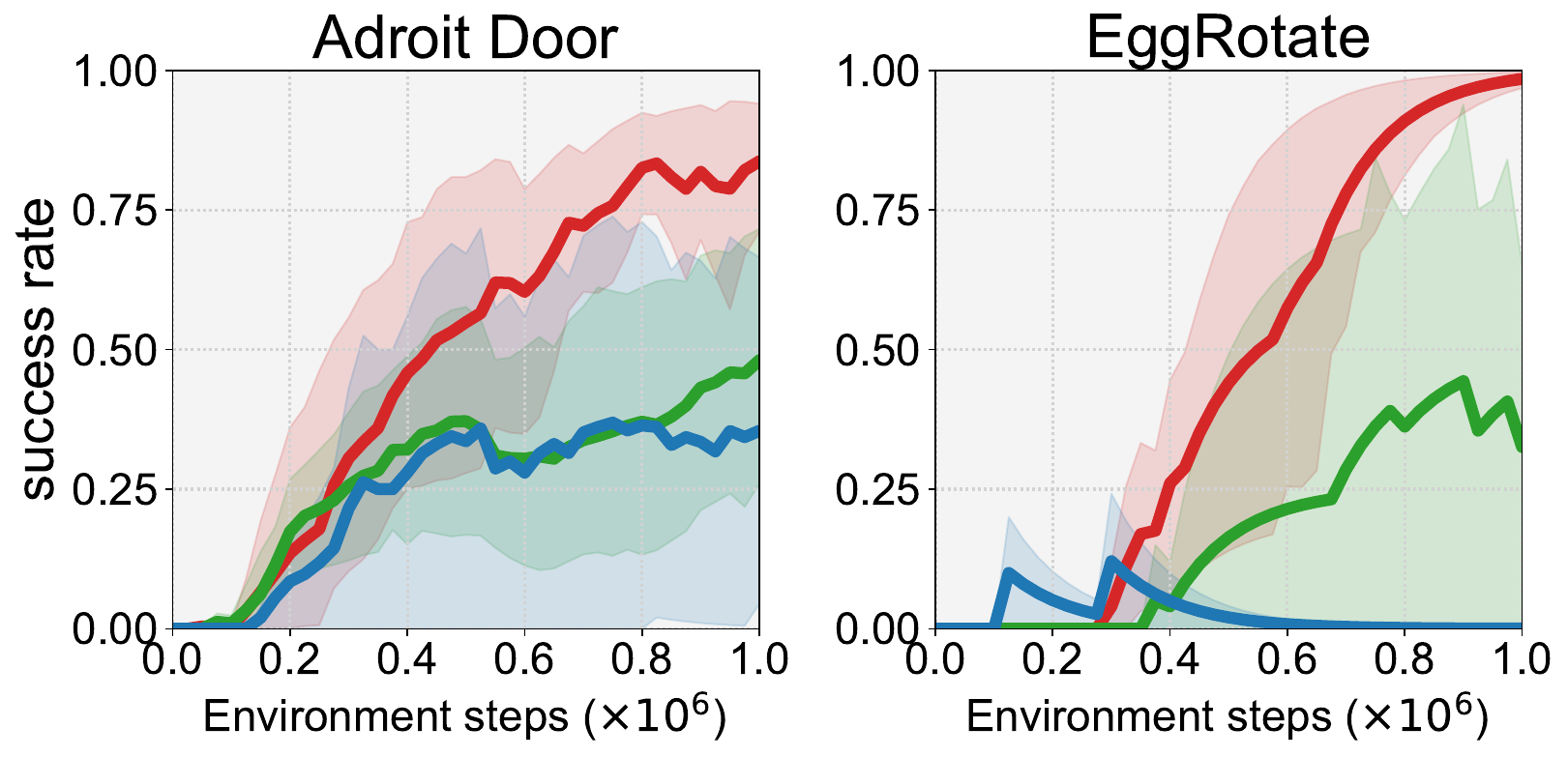}
        \\
        \includegraphics[height=0.55cm,keepaspectratio]{icml2024/figures/main_legend.pdf}
        \vspace{-0.8em}
        \caption{\small \textbf{Shadow hand manipulation tasks.} Success rate of \ourshort\ , SAC, TD3 on challenging shadow hand manipulation tasks from Adroit and Shadow Dexterous Hand suites. 6 seeds. All the results are in Figure~\ref{fig:dex_hand}.}
        \vspace{-0.6em}
        \label{fig:shadowhand_performance}
    \end{minipage}
    \hfill
    \begin{minipage}[t]{0.57\textwidth}
        \centering
        \includegraphics[width=8.0cm,keepaspectratio]{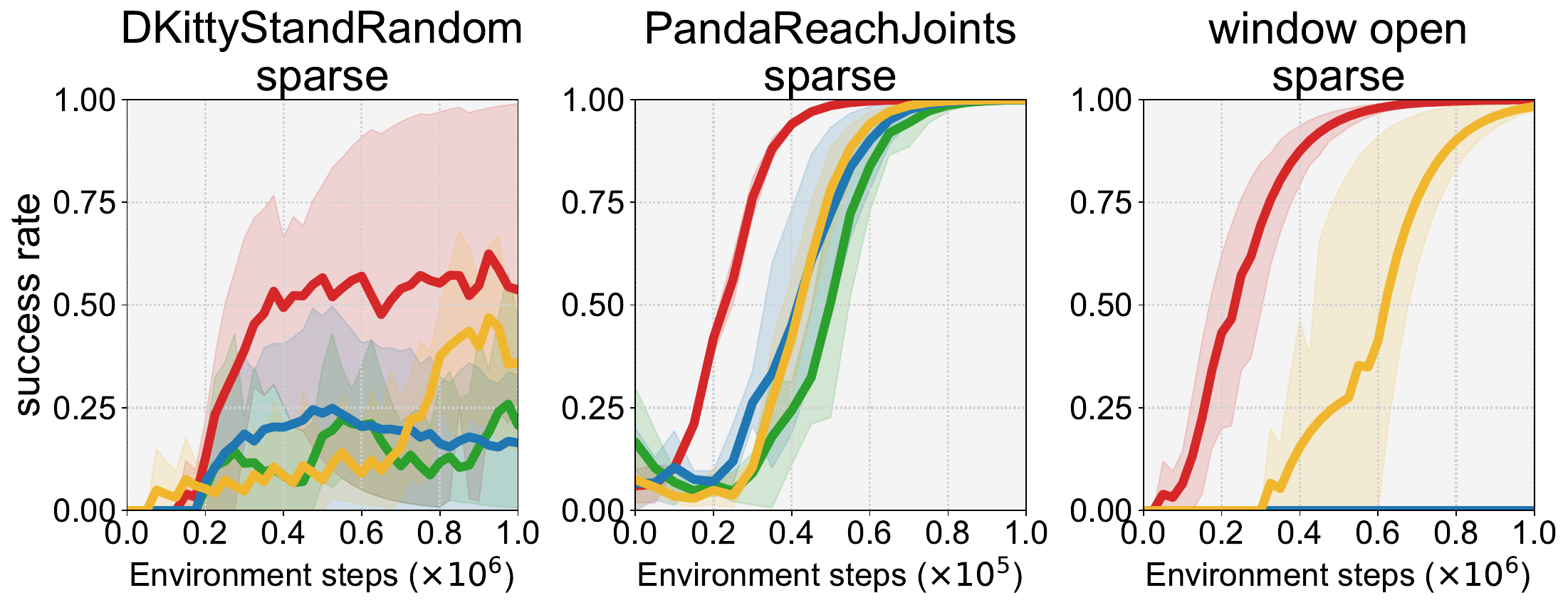}
        \\
        \includegraphics[height=0.55cm,keepaspectratio]{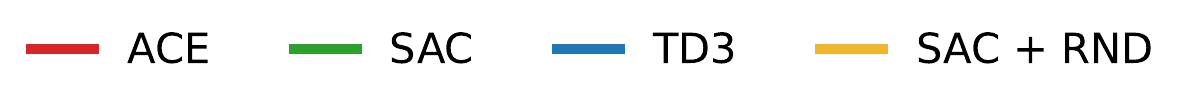}
        \vspace{-0.8em}
        \caption{\small \textbf{Sparse reward tasks.} Success rate of \ourshort, SAC, TD3, SAC+RND on sparse reward tasks from ROBEL, Panda-gym and Metaworld benchmark suites. 6 seeds. Full performance on sparse reward tasks is in Figure~\ref{fig:sparse_reward}.}
        \vspace{-0.5em}
        \label{fig:sparsereward_performance}
    \end{minipage}
\end{figure*}

\noindent\textbf{Dexterous hand manipulation.}\quad
To evaluate our method on high-dimensional tasks, we compare \ours\ with baselines on three dexterous hand manipulation tasks, including Adroit~\citep{adroit}, which involves controlling a robotic hand with up to 30 actuated degrees of freedom ($\mathcal{A}\in\mathbb{R}^{28}$), and Shadow Dexterous Hand~\citep{shadow}, a robotic hand with 24 degrees of freedom ($\mathcal{A}\in\mathbb{R}^{20}$). Notably, tasks with Shadow Dexterous Hand include multi-goal manipulation, and we train all the algorithms without goal information on these tasks. As shown in Figure~\ref{fig:shadowhand_performance}, \ours\ consistently outperforms baselines by a significant margin on all the dexterous hand manipulation tasks. This outstanding performance underscores the effectiveness of our causal model in computing meaningful causal weights in high-dimensional action spaces, leading to a substantial improvement in exploration efficiency for these tasks.

\noindent\textbf{Hard exploration tasks with sparse rewards.}\quad
To better illustrate the effectiveness of our proposed method in improving exploration efficiency, we evaluate our approach against baselines and the efficient exploration method RND~\citep{burda2018exploration} on tasks with sparse rewards. These tasks pose significant challenges for online RL exploration, covering both complex robot locomotion (Panda-gym~\citep{panda} and ROBEL~\citep{dkitty}) and manipulation (MetaWorld~\citep{metaworld}), as illustrated in Figure~\ref{fig:sparsereward_performance}. 
Across 6 sparse reward tasks, \ours\ outperforms the efficient exploration method RND and surpasses all other baselines by a significant margin, particularly excelling in tasks where the baselines completely fail to learn. These results showcase the superior sample efficiency of our method and further support the versatility of \ours\ in various challenging exploration tasks.

\vspace{-0.5em}
\subsection{Ablation Studies}
\begin{figure*}
    \centering
    \begin{minipage}[t]{0.48\textwidth}
        \includegraphics[height=3.2cm,keepaspectratio]{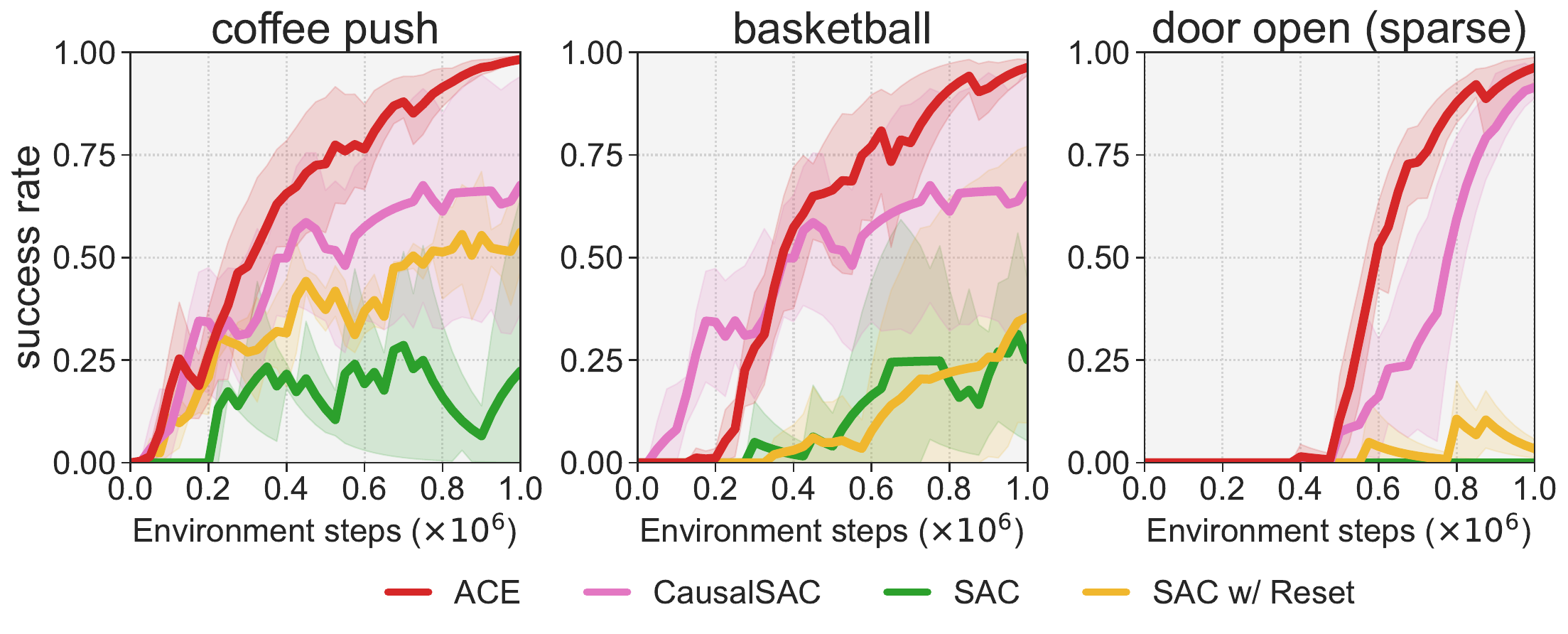}
        \vspace{-5mm}
        \caption{\small \textbf{Ablation experiments.} We ablate each component of \ours and show that each mechanism effectively combines to contribute to the overall effectiveness of \ours.}
        \label{fig:abl} 
    \end{minipage}
    \hfill
    \begin{minipage}[t]{0.48\textwidth}
        \includegraphics[height=3.3cm,keepaspectratio]{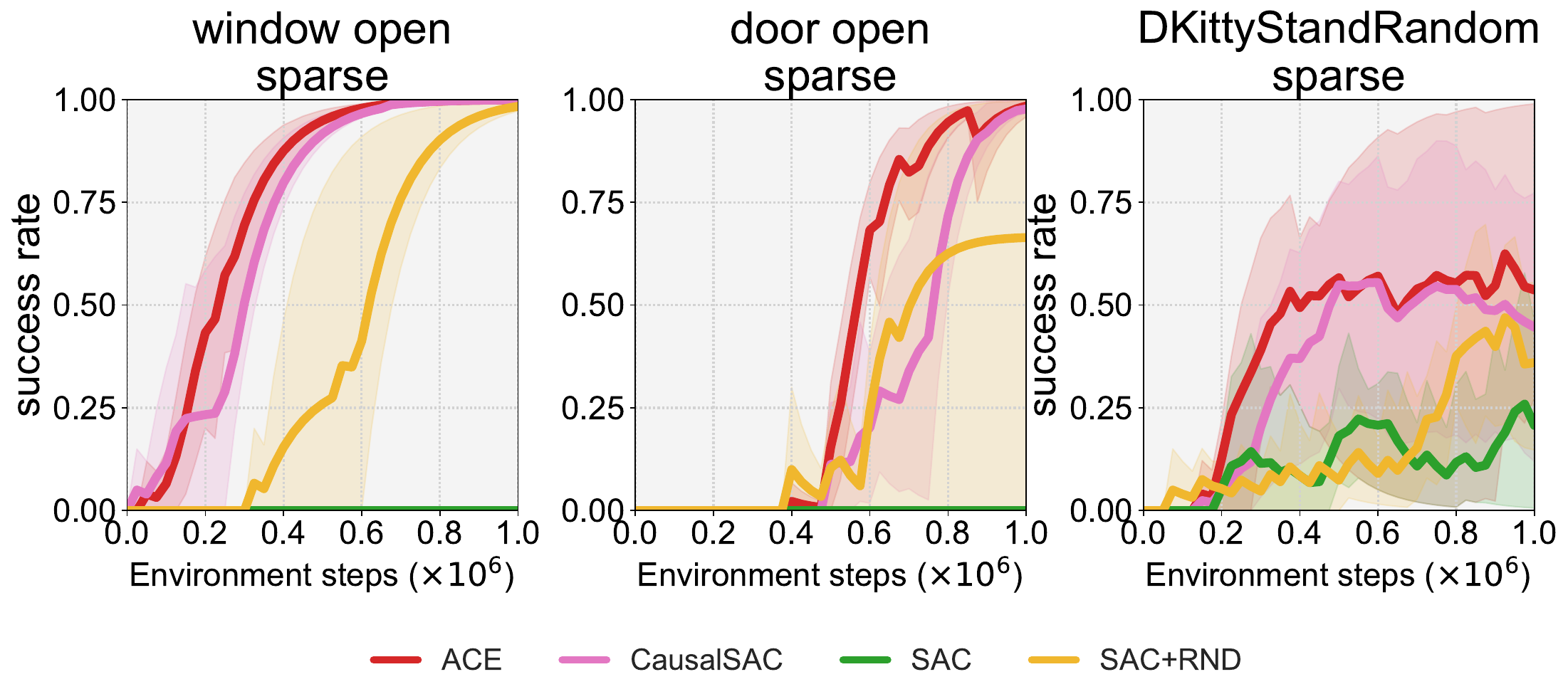}
        \vspace{-2mm}
        \caption{\small \textbf{Ablation studies on SOTA exploration techniques.} Learning curves of ACE, CausalSAC, SAC, and SOTA exploration technique RND on three sparse reward tasks.}
        \label{fig:abl} 
    \end{minipage}
\end{figure*}
\begin{figure*}[!h]
    \centering
    \begin{minipage}[t]{0.49\textwidth}
        \centering
        \includegraphics[height=2.9cm,keepaspectratio]{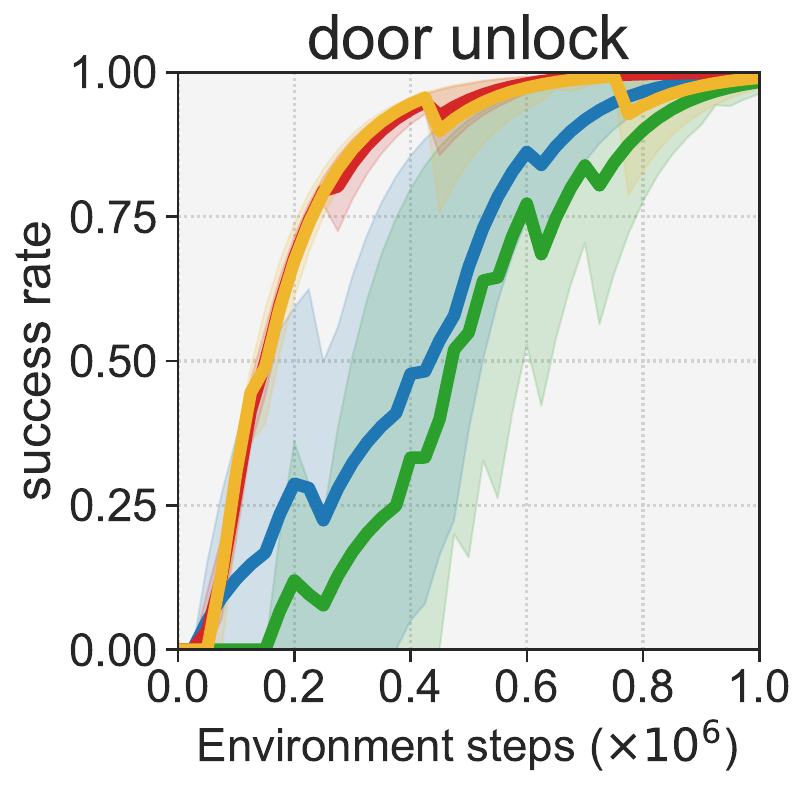}
        \hspace{2mm}
        \includegraphics[height=2.9cm,keepaspectratio]{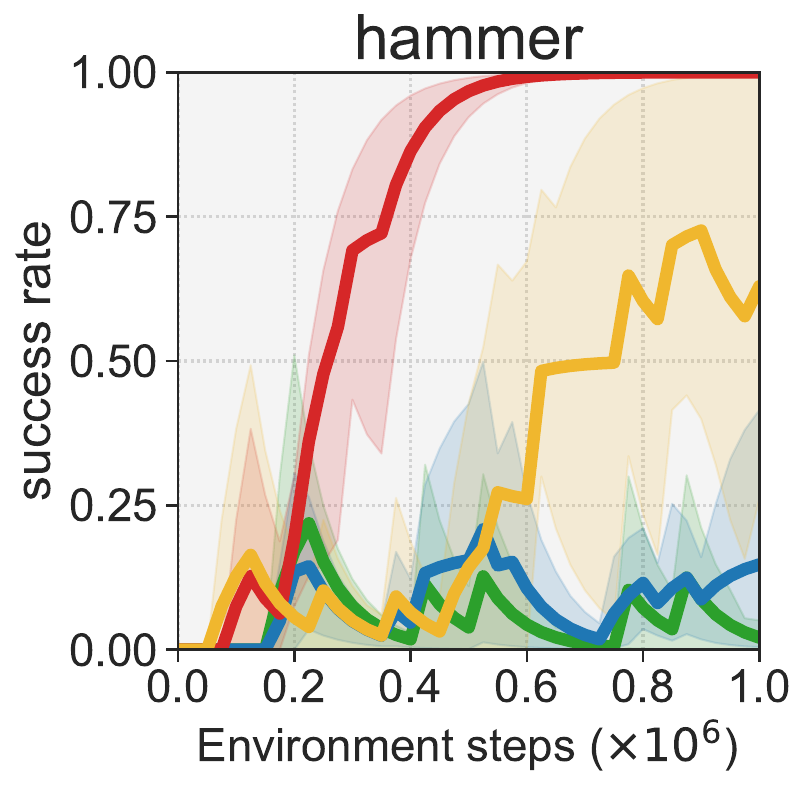}
        \\
        \includegraphics[height=0.6cm,keepaspectratio]{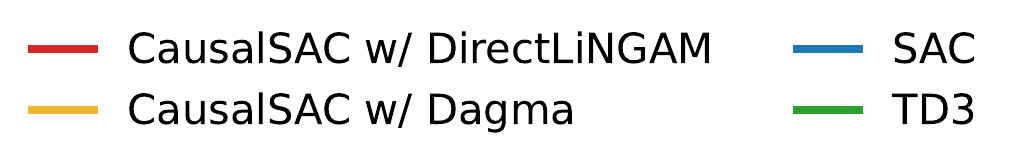}
        \vspace{-0.8em}
        \caption{\small \textbf{Different causal inference methods.} Performance curves of \textbf{CausalSAC} employing DirectLiNGAM or Dagma. Runs over 6 random seeds. }
        \vspace{-1.4em}
        \label{fig:differentcausalmodel}
    \end{minipage}
    \hfill
    \begin{minipage}[t]{0.49\textwidth}
            \centering
        \includegraphics[height=3.3cm,keepaspectratio]{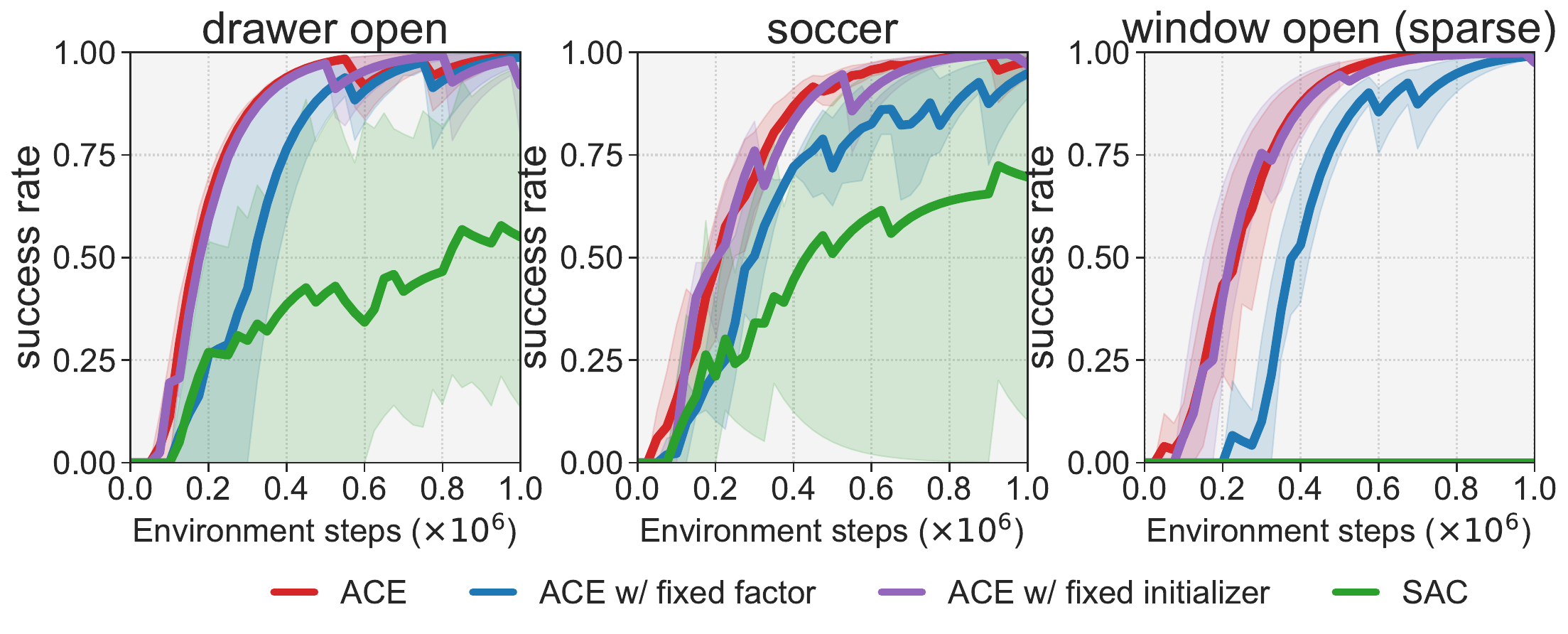}
        \vspace{-0.8em}
        \caption{\small \textbf{Ablation on reset mechanism factors.} Learning curves of \ourshort\ w/wo fixed reset factor $\eta$ and fixed initializer $\Phi$.}
        \vspace{-0.9em}
        \label{fig:abl-fixed-reset}
    \end{minipage}
\end{figure*}
To substantiate the efficacy of the design decisions in our algorithm, we conduct extensive ablation experiments.

\noindent\textbf{Effects of each mechanism.}\quad
We conduct ablation studies on the MetaWorld tasks with dense and sparse rewards to evaluate the contribution of each component to our method, including causality-aware entropy and gradient-dormancy-guided reset. Our ablation results are shown in Figure~\ref{fig:abl}.

We observe that all components of our proposed methodology significantly contribute to the final performance of \ours. 
Even without the reset mechanism, CausalSAC (vanilla ACE) demonstrates superior performance. Further, results on sparse reward tasks show that CausalSAC outperforms other SOTA exploration techniques. This highlights its potential in challenging environments.

Simultaneously, our reset mechanism also plays a crucial role in further improving performance, especially in sparse reward tasks. Interestingly, applying the gradient-dormancy-guided reset solely to SAC demonstrates improved performance and sample efficiency, suggesting that the gradient dormant phenomenon might indeed be one of the factors contributing to SAC's suboptimal performance in hard tasks.

\noindent\textbf{Different causal inference methods.}\quad
We initially opted for DirectLiNGAM due to its simplicity and efficacy in learning causal effects. However, to explore the adaptability of our framework with other score-based causal inference methods, we conducted additional experiments using Dagma~\citep{bello2022dagma}. These experiments were aimed at assessing whether different causal inference techniques could yield comparable results within our framework. The results in Figure~\ref{fig:differentcausalmodel} indicate that the integration of Dagma into our method produces outcomes that are on par with those obtained using DirectLiNGAM. This suggests that our framework is versatile and can effectively work with various causal inference methods. Notably, to eliminate the possible influence from other factors except the underlying causal inference method, we deactivate the reset mechanism in \ourshort, thus excluding its impact. We employ CausalSAC for ablations related to the causal inference method.

\begin{figure}
    \centering
    \vspace{-10mm}
    \includegraphics[height=3.2cm,keepaspectratio]{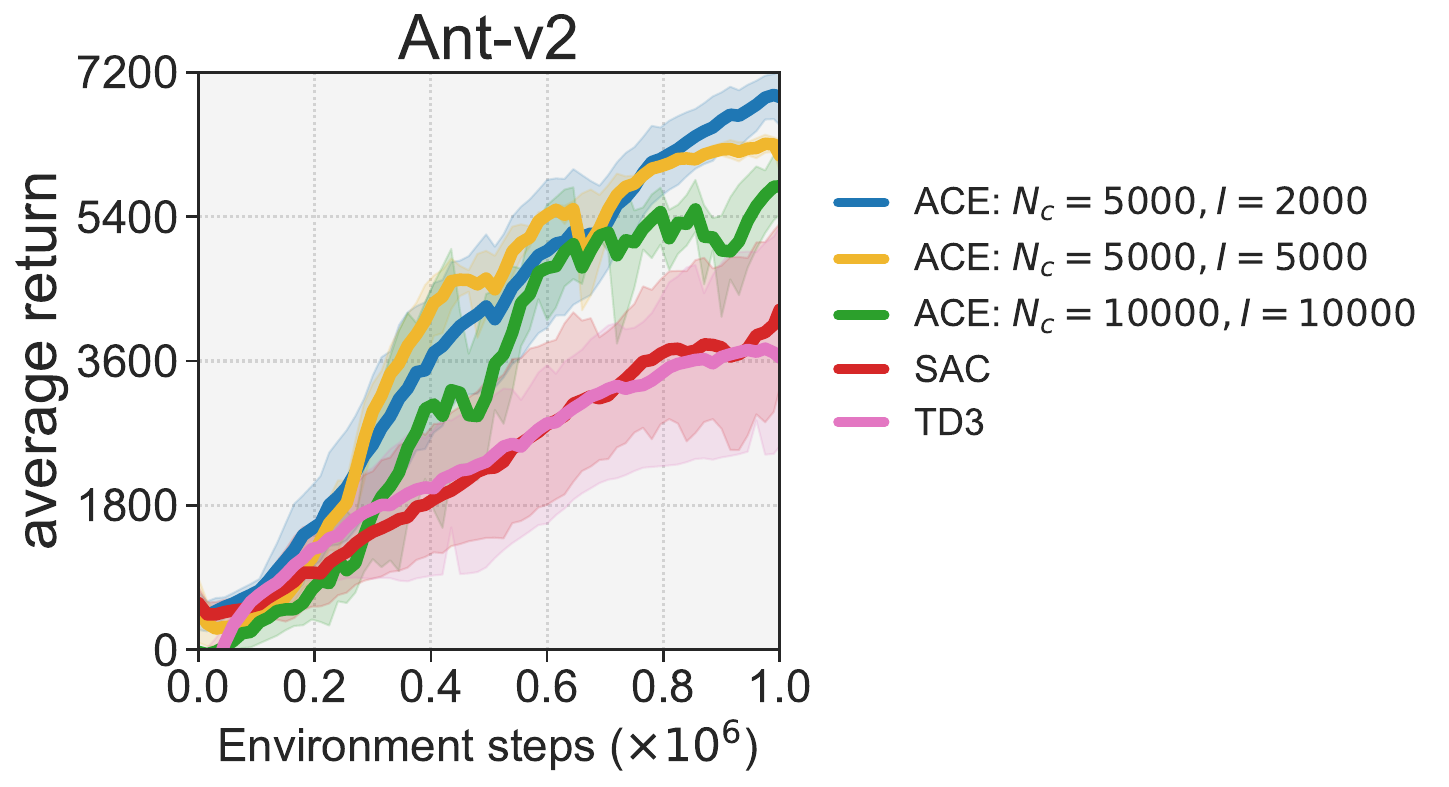}
    \vspace{-0.5em}
    \caption{\small \textbf{Hyperparameter study.} Performance curves of \ourshort\ with different hyperparameters for calculating causal weights.}
    \vspace{-1.5em}
    \label{fig:HP-study}
\end{figure}

\noindent\textbf{Ablation studies on reset mechanism factors.}
We conduct ablation studies on the fixed reset factor $\eta$ and the fixed network initializer $\phi$ across three tasks. As shown in Figure~\ref{fig:abl-fixed-reset}, fixing the reset factor (i.e., no longer guiding it by the dormancy degree) results in a significant performance drop. This highlights the importance of the proposed dormancy degree in our method. On the other hand, fixing the network initializer $\phi$ does not affect the performance of the dormancy-guided reset mechanism in ACE.

\noindent\textbf{Hyperparameter study.}\quad
The extra hyperparameters introduced by \ourshort\ are sample size for causality $N_c$ and causality computation interval $I$. The primary choices for both hyperparameters are guided by the objective of achieving a balanced trade-off between computational efficiency and algorithmic performance.
And they are sufficient to achieve strong performance throughout all our experiments. 

We conduct experiments on these two hyperparameters; refer to Figure~\ref{fig:HP-study}. We see that
reducing the causality computation interval may increase the performance yet cause more computation cost. And the performance of \ourshort\ is not highly sensitive to the hyperparameters.

\section{Related Works}
\label{sec:relate}
% \paragraph{Causal Reinforcement Learning.}
\noindent\textbf{Causal Reinforcement Learning.}\quad
In the past decades, causality and reinforcement learning have independently undergone significant theoretical and technical advancements, yet the potential for a synergistic integration between the two has been underexplored~\citep{zeng2023survey}. Recently, recognizing the substantial capabilities of causality in addressing data inefficiency and interpretability challenges within RL, there has been a surge of research in the domain of causal reinforcement learning~\citep{gershman2017reinforcement,bannon2020causality,zeng2023survey,deng2023causal, lin2024safety, ding2022generalizing}.

While existing methods in this area can be categorized based on whether causal information is explicitly given or not, our work falls into the more challenging, practical, and realistic category where the causal structure and effects are not explicitly provided. 
In causal reinforcement learning, one of the challenges lies in the characterization of shifting structures and effects from data, which might affect the performances of policy learning.
To this end, alternatives may involve incorporating the changing structures between states~\citep{luczkow2021structural} into policy learning or modeling the changes using some dynamic factors~\citep{huang2022adarl,feng2022factored}. Our approach, however, focuses on capturing changing causal effects, a nuanced facet that improves policy exploration.

% \paragraph{Exploration in RL.}
\noindent\textbf{Exploration in RL.}\quad
Efficient exploration in online RL, especially in high-dimensional environments with sparse rewards, remains a significant challenge. Exploration strategies can be broadly categorized into two major groups based on key ideas and principles. One category is uncertainty-oriented exploration~\citep{jin20, menard21b, kaufmann21}, utilizing techniques like the upper confidence bound (UCB)~\citep{chen2017ucb} to incorporate value estimate uncertainty for guiding exploration. The other category is intrinsic motivation-oriented exploration, incentivizing agents to explore by maximizing intrinsic rewards~\citep{pathak2017curiosity, burda2018exploration, sekar20, badia20}, count-based state novelty~\citep{bellemare16, tang17, ostrovski17}, or maximizing state entropy as an intrinsic reward~\citep{lee19, hazan19, mutti22, yang23}. In contrast to exploration methods that focus on sampling data, the concept of resetting the neural network has gained attention in RL as a unique approach to mitigate the loss of network expressivity. 
\citet{nikishin22} address the primacy bias by periodically reinitializing the parameters, and \citet{nikishin23} propose to temporarily freeze the current network and leverage newly initialized weights. The dormant neuron phenomenon, as discussed by \citet{dormant}, suggests resetting dormant neurons to preserve network expressivity during training. For visual RL, \citet{drm} propose a perturbation technique and exploration strategy based on the dormant neuron ratio. In contrast to these prior exploration methods, our approach combines causality-aware max-entropy RL with a reset mechanism, adjusting exploration through causality between rewards and policy, and employs a novel gradient dormancy degree for resetting. We also discuss extensive related works about maximum-entropy RL in Appendix~\ref{app:relate}.

\section{Conclusions and Discussion}
\label{sec:con}
This paper introduces insights into the consideration of the significance of various primitive behaviors throughout the policy learning process. Building upon this understanding, we propose a causal policy-reward structural model to quantify the impact of each action dimension on rewards. We introduce a causality-aware off-policy actor-critic algorithm with a novel gradient-dormancy-guided reset mechanism, achieving efficient and effective exploration and establishing a substantial lead across various domains and tasks.

Looking ahead, there are several directions for further exploration stemming from this work. First, despite our analysis of computing causal weights in the latent space and their application to DrM~\citep{drm} in Appendix~\ref{app:drm}, there is untapped potential for exploring additional applications of causality-aware exploration to enhance the sample efficiency of visual RL. Second, it is worth discussing the prospect of leveraging dynamic models in model-based RL to facilitate more efficient computation of causality between rewards and policy. Concerning time efficiency, a detailed analysis is provided in Appendix~\ref{app:efficiency}, confirming the high efficiency and cost-effectiveness of our algorithm.
We believe that the implications of causality-aware exploration extend beyond our current work, offering more effective solutions to enhance the sample efficiency of online RL.

% \clearpage
\section*{Broader Impact}
This research advances both cognition and application of Reinforcement Learning, particularly in the domain of off-policy actor-critic. Our exploration uncovers a fascinating insight: by harnessing the causal influences of each action dimension on potential rewards, we can dramatically boost the efficiency of exploration. Moreover, the resetting technique we devised is not just innovative; it echoes the renewal mechanism found in the human brain, empowering RL agents to escape the confines of local optima with remarkable agility.
This breakthrough not only deepens our understanding of the agent's decision-making process but also mirrors the dynamic and adaptive learning capabilities akin to human cognition.

The introduction of causality entropy and the resetting mechanism as flexible enhancements to existing models introduce an exhilarating possibility for the evolution of RL. However, it is worth noting that venturing into real-world environments with these RL agents brings to light a significant challenge: navigating the need for stringent safety measures to curb any risky behaviors during exploration.
\bibliography{reference}
\bibliographystyle{icml2024}

%%%%%%%%%%%%%%%%%%%%%%%%%%%%%%%%%%%%%%%%%%%%%%%%%%%%%%%%%%%%%%%%%%%%%%%%%%%%%%%
%%%%%%%%%%%%%%%%%%%%%%%%%%%%%%%%%%%%%%%%%%%%%%%%%%%%%%%%%%%%%%%%%%%%%%%%%%%%%%%
% APPENDIX
%%%%%%%%%%%%%%%%%%%%%%%%%%%%%%%%%%%%%%%%%%%%%%%%%%%%%%%%%%%%%%%%%%%%%%%%%%%%%%%
%%%%%%%%%%%%%%%%%%%%%%%%%%%%%%%%%%%%%%%%%%%%%%%%%%%%%%%%%%%%%%%%%%%%%%%%%%%%%%%
\newpage
\appendix
\onecolumn
\section{Theoretical Analyses}\label{ap-omittedproofs}
\subsection{Causal identifiability}
\label{app:proof}
We first give definitions of the Markov condition and faithfulness assumption, which will be used in our theoretical analyses.
\begin{assumption}[Global Markov Condition~\citep{spirtes2000causation,pearl2009causality}]
    The distribution $p$ over a set of variables $\rmV = (s_{1,t}, ..., s_{\mathrm{dim}\mathcal{S},t}, a_{1,t}, ..., a_{\mathrm{dim}\mathcal{A},t}, r_{t})^T$ satisfies the global Markov condition on the graph if for any partition $(\rmS, \rmA, \rmR)$ in $\rmV$ such that if $\rmA$ d-separates $\rmS$ from $\rmR$, then $p(\rmS, \rmR|\rmA) = p(\rmS|\rmA)p(\rmR|\rmA)$.
\end{assumption}

\begin{assumption}[Faithfulness Assumption~\citep{spirtes2000causation,pearl2009causality}] 
    For a set of variables $\rmV = (s_{1,t}, ..., s_{\mathrm{dim}\mathcal{S},t}, a_{1,t}, ..., a_{\mathrm{dim}\mathcal{A},t}, r_{t})^T$, 
    there are no independencies between variables that are not entailed by the Markovian Condition.
\end{assumption}

With these two assumptions, we provide the following proposition to characterize the condition of the causal relationship existence so that we are able to uncover those key actions from conditional independence relationships.

\begin{proposition} \label{appendix-prop:noind}
    Under the assumptions that the causal graph is Markov and faithful to the observations, there exists an edge from $a_{i,t}$ to $r_t$ if and only if $a_{i,t} \not\!\perp\!\!\!\perp r_t | \rvs_{t}, \rva_{-i, t}$, where $\rva_{-i, t}$ are states of $\rva_t$ except $a_{i, t}$.  
\end{proposition}
\begin{proof}
    (i) We first prove that if there exists an edge from $a_{i,t}$ to $r_t$, then $a_{i,t} \not\!\perp\!\!\!\perp r_t | \rvs_{t}, \rva_{-i, t}$. We prove it by contradiction. Suppose that $a_{i,t}$ is independent of $r_t$ given $\rvs_{t}, \rva_{-i, t}$. According to the faithfulness assumption, we get that from the graph, $a_{i,t}$ does not have a directed path to $r_t$, i.e., there is no edge between $a_{i,t}$ and $r_t$. It contradicts our statement about the existence of the edge. 
    
    (ii) We next prove that if $a_{i,t} \not\!\perp\!\!\!\perp r_t | \rvs_{t}, \rva_{-i, t}$, then there exists an edge from $a_{i,t}$ to $r_t$. Similarly, by contradiction, we suppose that $a_{i,t}$ does not have a directed path to $r_t$. From the definition of our MDP, we see in the graph that the path from $a_{i,t}$ to $r_t$ could be blocked by $\rvs_{t}$ and $\rva_{-i, t}$. According to the global Markov condition, $a_{i,t}$ is independent of $r_t$ given $\rvs_{t}$ and $\rva_{-i, t}$, which contradicts the assumption about the dependence between $a_{i,t}$ and $r_t$.
\end{proof}

We next provide the theorem to guarantee the identifiability of the proposed causal structure.

\begin{theorem} \label{appendixtheo:ident}
    Suppose  $\rvs_t$,  $\rva_t$, and $r_t$ follow the MDP model with Eq.(\ref{eq:r_func}). Under the Markov condition, and faithfulness assumption, the structural vectors $\rmB_{\rvs \to r|\rva}$ and $\rmB_{\rva \to r|\rvs}$ are identifiable.
\end{theorem}
\begin{proof}
    We prove it motivated by~\citep{huang2022adarl}. Denote all variable dimensions in the MDP by $\rmV$, with $\rmV = \{s_{1,t}, ...s_{\mathrm{dim}\mathcal{S},t}, a_{1,t},...,a_{\mathrm{dim}\mathcal{A},t}, r_t\}$, and these variables form a dynamic Bayesian network~\citep{murphy2002dynamic}. Note that our theorem only involves possible edges from state dimensions $s_{i,t} \in \rvs_{t}$ to the reward $r_t$ or from action dimensions $a_{j,t} \in \rva_{t}$ to the reward $r_t$.~\citep{huang2020causal} showed that under the Markov condition and faithfulness assumption, even with non-stationary data, for every $V_i, V_j \in \rmV$, $V_i$ and $V_j$ are not adjacent in the graph if and only if they are independent conditional on some subset of other variables in $\rmV$, i.e., $\{V_l|l\neq i, l\neq j\}$. Based on this, we can asymptotically identify the correct graph skeleton over $\rmV$. Besides, due to the property of dynamic Bayesian networks that future variables can not affect past ones, we can determine the directions as $a_{i,t} \to r_t$ if $a_{i,t}$ and $r_t$ are adjacent. So does $s_{j,t} $ and $ r_t$.
    Thus, the structural vectors $\rmB_{\rvs \to r|\rva}$ and $\rmB_{\rva \to r|\rvs}$, which are parts of the graph in $\rmV$,  are identifiable. 
    Note that, following results of~\citep{shimizu2011directlingam}, if we further assume the linearity of observations as well as the non-Gaussianity of the noise terms, we can uniquely identify 
    $\rmB_{\rvs \to r|\rva}$ and $\rmB_{\rva \to r|\rvs}$, including both causal directions and causal effects.   
\end{proof}

\subsection{Approximate dynamic programming properties}\label{sec:proof-adp-properties}
\begin{proposition}[Policy evaluation]\label{policy-evaluation}
Consider an initial $Q_0:\mathcal{S}\times\mathcal{A}\rightarrow\mathbb{R}$ with $\vert \mathcal{A}\vert < \infty$, and define $Q_{k+1}= \mathcal{T}_c^{\pi}Q_{k}$. Then the sequence $\{Q_{k}\}$ converges to a fixed point $Q^{\pi}$ as $k\rightarrow \infty$.
\end{proposition}
\begin{proof}
First, let us show that our causal policy-reward Bellman operator $\mathcal{B}$ is a $\gamma$-contraction operator in the $\mathcal{L}_\infty$ norm.

Let $Q_1$ and $Q_2$ be two arbitrary $Q$ functions, for the Bellman operator $\mathcal{T}_c$, we have,
\[
\begin{aligned}
\Vert \mathcal{T}_{c}^{\pi} Q_1 - \mathcal{T}_{c}^{\pi} Q_2\Vert_\infty=
& \max_{\rvs,\rva} \vert \gamma \mathbb{E}_{\rvs'}\big[\mathbb{E}_{\rva'\sim \pi}Q_1(\rvs',\rva') - \gamma \mathbb{E}_{\rva'\sim \pi}Q_2(\rvs',\rva')\big]\vert \\
\leq &\gamma \max_{\rvs,\rva} \mathbb{E}_{\rvs'}\vert \mathbb{E}_{\rva'\sim \pi}Q_1(\rvs',\rva') - \mathbb{E}_{\rva'\sim \pi}Q_2(\rvs',\rva')\vert\\
\leq & \gamma\max_{\rvs,\rva} \mathbb{E}_{\rvs'}\mathbb{E}_{\rva'\sim \pi}\vert Q_1(\rvs',\rva') - Q_2(\rvs',\rva')\vert\\
\leq & \gamma \max_{\rvs,\rva}\Vert Q_1- Q_2\Vert_\infty
=  \gamma \Vert Q_1 -Q_2\Vert_\infty
\end{aligned}
\]
we conclude that the Bellman operator $\mathcal{T}_c$ satisfies $\gamma$ -contraction property, which naturally leads to the conclusion that any initial Q function will converge to a unique ﬁxed point by repeatedly applying $\mathcal{T}_c^{\pi}$.
\end{proof}

\begin{proposition}[Policy improvement]\label{proof-policy-improvement}
Let $\pi_{k}$ be the policy at iteration $k$, and $\pi_{k+1}$ be the updated policies, where  $\pi_{k+1}$ is the greedy policy of the $Q$-value. 
Then for all $(\rvs,\rva)\in \mathcal{S}\times \mathcal{A}$, $\vert \mathcal{A}\vert < \infty$, we have $Q^{\pi_{k+1}}(\rvs,\rva) \geq Q^{\pi_k}(\rvs,\rva)$.
\end{proposition}
\begin{proof}
At iteration $k$, $\pi_k$ denotes the policy, and the corresponding value function is $Q^{\{\mu,\pi\}}$.
We update the policy from $\pi_k$  to $\pi_{k+1}$, where $\pi_{k+1}$  is the greedy policy w.r.t $J_{\pi_k}(\pi)$, \emph{i.e.}, $\pi_{k+1} = \arg\max_{\pi}\mathbb{E}_{\rva\sim \pi}[Q^{\pi_k}(\rvs,\rva)+\alpha\mathcal{H}_c(\pi(\rva\vert \rvs))]$. 

Since $\pi_{k+1} = \arg\max_{\pi}J_{\pi_k}(\pi)$, we have that 
$J_{\pi_k}(\pi_{k+1}) \geq J_{\pi_k}(\pi_k)$. 
Expressing $J_{\pi_k}(\pi_{k+1})$ and $J_{\pi_k}(\pi_k)$ by their definition, we have 
$\mathbb{E}_{\rva\sim \pi_{k+1}}[Q^{\pi_k}(\rvs,\rva) + \alpha \mathcal{H}_c(\pi_{k+1}(\rva\vert \rvs))] \geq \mathbb{E}_{\rva\sim \pi_k}[Q^{\pi_k}(\rvs,\rva) + \alpha \mathcal{H}_c(\pi_{k}(\rva\vert \rvs))]$.

In a similar way to the proof of the soft policy improvement~\citep{haarnoja2017reinforcement}, we come to the following inequality:

\[
\begin{aligned}
Q^{\pi_k}(\rvs_t,\rva_t) 
 =& r(\rvs_t,\rva_t) + \gamma\mathbb{E}_{\rvs_{t+1}}
 \big\{\mathbb{E}_{a_{t+1}\sim\pi_k}[Q^{\pi_k}(\rvs_{t+1},\rva_{t+1}) + \alpha \mathcal{H}_c(\pi_{k}(\rva_{t+1}\vert \rvs_{t+1})]
 \big\} 
 \\
\leq & r(\rvs_t,\rva_t) + \gamma\mathbb{E}_{\rvs_{t+1}}\{\mathbb{E}_{a_{t+1}\sim\pi_{k+1}}[Q^{\pi_{k}}(\rvs_{t+1}, \rva_{t+1}) + \alpha \mathcal{H}_c(\pi_{k+1}(\rva_{t+1}\vert\rvs_{t+1})]\}\\
& \vdots\\
\leq& Q^{\pi_{k+1}}(\rvs_t,\rva_t)
\end{aligned}
\]

Here, the inequality is obtained by repeatedly expanding $Q^{\pi_k}$ on the RHS through 
$Q^{\pi_k}(\rvs,\rva) 
 =r(\rvs,\rva) + \gamma\mathbb{E}_{\rvs'}[\mathbb{E}_{a'\sim\pi_k}[Q^{\pi_k}(\rvs',\rva')+\alpha\mathcal{H}_c(\pi_{k}(\rva'\vert\rvs')]]$ 
and applying the inequality $\mathbb{E}_{\rva\sim \pi_{k+1}}[Q^{\pi_k}(\rvs,\rva)-\omega(\rvs,\rva\vert \pi_{k+1})] \geq \mathbb{E}_{\rva\sim \pi_k}[Q^{\pi_k}(\rvs, \rva) + \mathcal{H}_c(\pi_{k}(\rva\vert\rvs)]$.  Finally, we arrive at convergence to $Q^{ \pi_{k+1}}(\rvs_t,\rva_t)$ and finish the proof.
\end{proof}

\begin{proposition}
[\textbf{Policy iteration}]
Assume $\vert \mathcal{A}\vert < \infty$, repeated application of the policy evaluation and policy improvement to any initial policy converges to a policy $\pi^*$, s.t. $Q^{\pi^*}(\rvs_t,\rva_t)\geq Q^{\pi'}(\rvs_t,\rva_t), \forall \pi'\in \Pi, \forall (\rvs_t,\rva_t) \in \mathcal{S}\times \mathcal{A} $.
\end{proposition}

\begin{proof}
Let $\Pi$ be the space of policy distributions and let $\pi_i$ be the policies at iteration $i$. 
By the policy improvement property in Proposition~\ref{policy-improvement}, the sequence $Q^{\pi_i}$ is monotonically increasing. Also, for any state-action pair $(\rvs_t,\rva_t)\in \mathcal{S}\times \mathcal{A}$, each $Q^{\pi_i}$ is bounded due to the discount factor $\gamma$. Thus, the sequence of $\pi_i$ converges to some $\pi^*$ that are local optimum. 
We will still need to show that $\pi^*$ are indeed optimal; we assume finite MDP, as typically assumed for convergence proof in usual policy iteration~\citep{sutton1988learning}. 
At convergence, we get $J_{\pi^*}(\pi^*)[\rvs]\geq J_{\pi^*}(\pi')[\rvs], \forall \pi'\in \Pi$. Using the same iterative augument as in the proof of Proposition~\ref{policy-improvement}, we get $Q^{\pi^*}(\rvs,\rva) \geq Q^{\pi'}(\rvs,\rva)$ for all $(\rvs,\rva)\in \mathcal{S}\times \mathcal{A}$. Hence, $\pi^*$ are optimal in $\Pi$.
\end{proof}

\section{Extensive Related Works}
\label{app:relate}

\paragraph{Maximum-Entropy Reinforcement Learning.}
Maximizing entropy in Reinforcement Learning (RL) aims to optimize policies for both maximizing the expected return and the expected entropy of the policy. This approach has found application in various RL contexts, ranging from inverse RL~\citep{ziebart2008maximum} to multi-goal RL~\citep{zhao2019maximum}. Guided policy search~\citep{levine2013guided} utilizes maximum entropy to guide policy learning towards high-reward regions. The incorporation of entropy regularization establishes a connection between value-based and policy-based RL~\citep{o2016combining, schulman2017equivalence}. 

In the domain of off-policy RL, soft Q-learning~\citep{haarnoja2017reinforcement} and its variants~\citep{schulman2017equivalence, grau2018soft} learn a softened value function by replacing the hard maximum operator in the Q-learning update with a softmax operator. Soft actor-critic (SAC)~\citep{haarnoja2017reinforcement} introduces a maximum entropy actor-critic algorithm, offering a balance between sample efficiency and stability. Moreover, \citet{han2021max} propose a max-min entropy framework to encourage visiting states with low entropy and maximize the entropy of these low-entropy states to enhance exploration~\citep{han2021max}. Another direction is maximum state entropy exploration. \cite{hazan2019provably} present an efficient algorithm for optimizing intrinsically state entropy objectives, and \citet{seo2021state} use state entropy as an intrinsic reward to improve exploration. Our proposed approach, falling within the spectrum of SAC variants, introduces causality-aware weighted entropy to selectively enhance exploration for different primitive behaviors, showcasing superior efficiency and effectiveness.

\paragraph{RL in multi-stage learning.} Inspired by human cognition, dividing the original task into multiple stages for policy learning has been explored in RL through heuristic methods tailored to specific tasks~\citep{jinnai2019exploration,liu2023hierarchical}. For instance, hierarchical RL~\citep{sutton1999between,pateria2021hierarchical} has implemented this idea by introducing additional subgoal spaces or leveraging the semi-Markov assumption. This allows the agent to segment tasks into different subtasks~\citep{nachum2018data}, options~\citep{machado2023temporal}, or skills~\citep{guan2022leveraging} as multiple stages, enabling policies to possess distinct exploration capabilities across various stages, ultimately enhancing the success of agents in complex continuous control tasks. 
Further, an interesting work is Human-AI shared control via Policy Dissection~\citep{li2022human}.
While both ACE and Policy Dissection draw inspiration from neuroscience, particularly in the realm of motion primitives, they operate in different paradigms. Policy Dissection fosters human-AI shared control, enabling human collaboration with RL agents in intricate environments. It establishes an interpretable interface on the agent, allowing humans to directly influence agent behavior and facilitate shared control. 

In contrast to explicit multi-stage approaches or human-guided task stage division, our proposed method \ours\ does not necessitate a clear task stage division. Instead, it can identify and prioritize actions with a high potential for reward through causal discovery dynamically, emphasizing the various importance of each action dimension in different stages for enhanced exploration and performance.

\section{Environment Setup}\label{appendix:scenarios}
\begin{figure}[H]
    \centering
    \newcommand{\img}[1]{\hspace{0.05cm}\includegraphics[width=0.15\linewidth]{{#1}}\hspace{0.05cm}}
    \begin{tabular}{@{}c@{}c@{}c@{}c@{}c@{}@{}c}
        \img{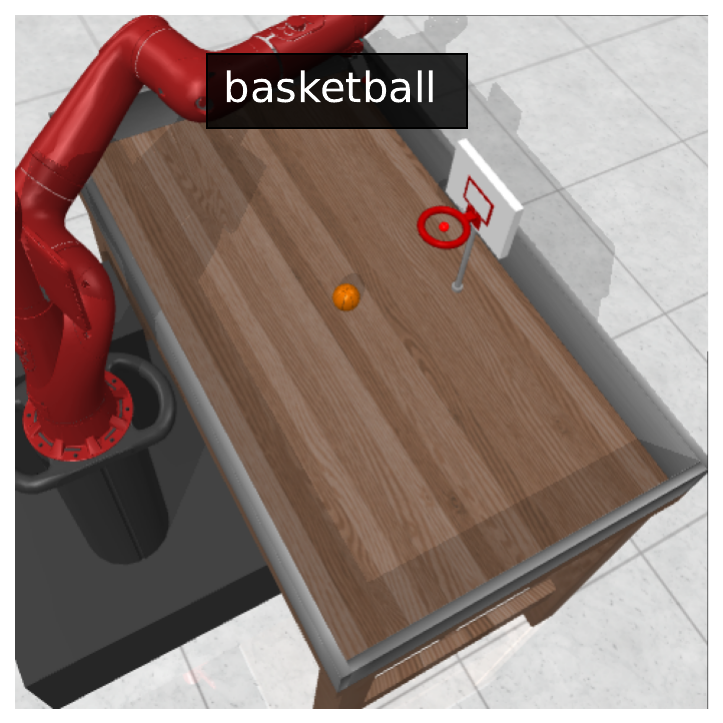} &
        \img{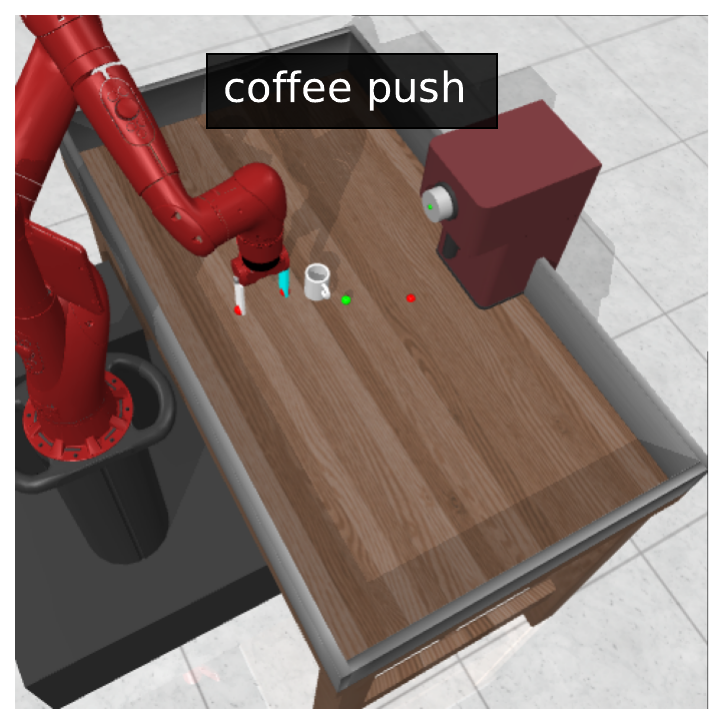} &
        \img{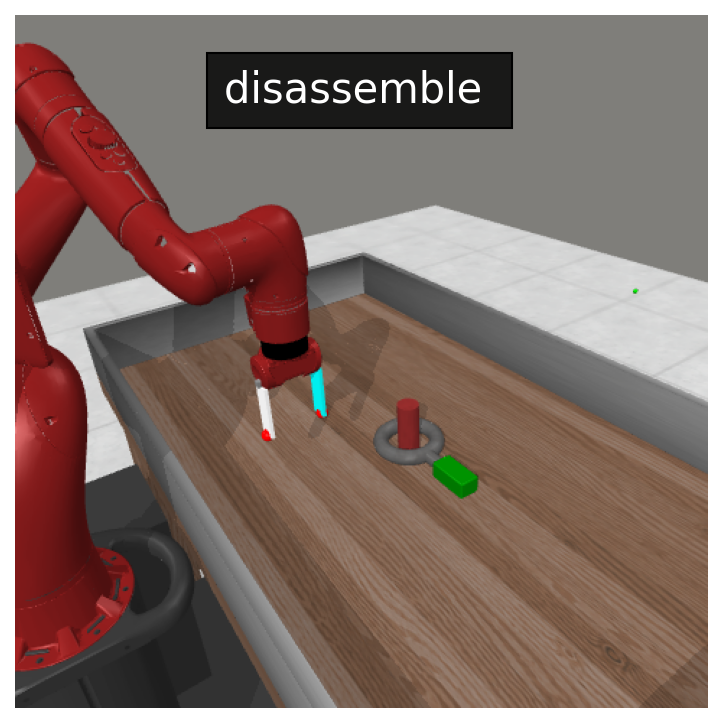} &
        \img{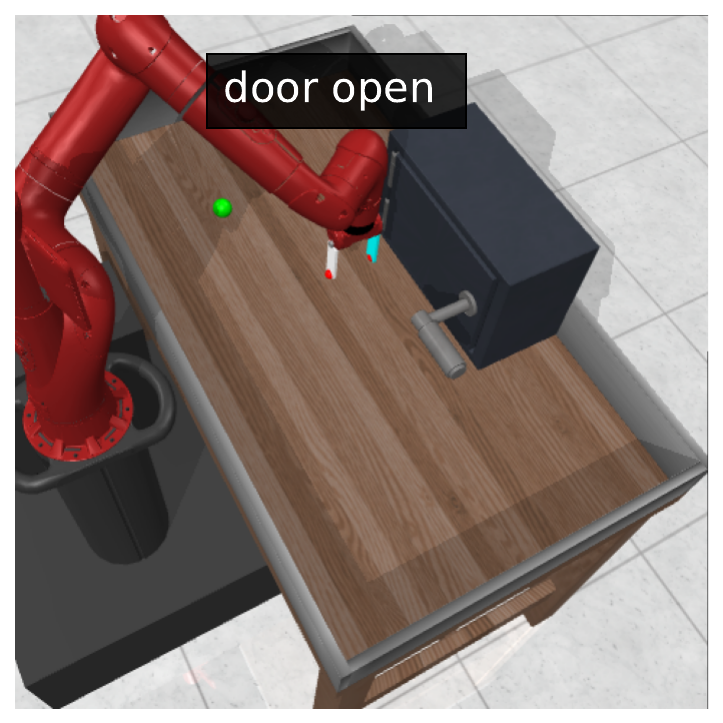} &
        \img{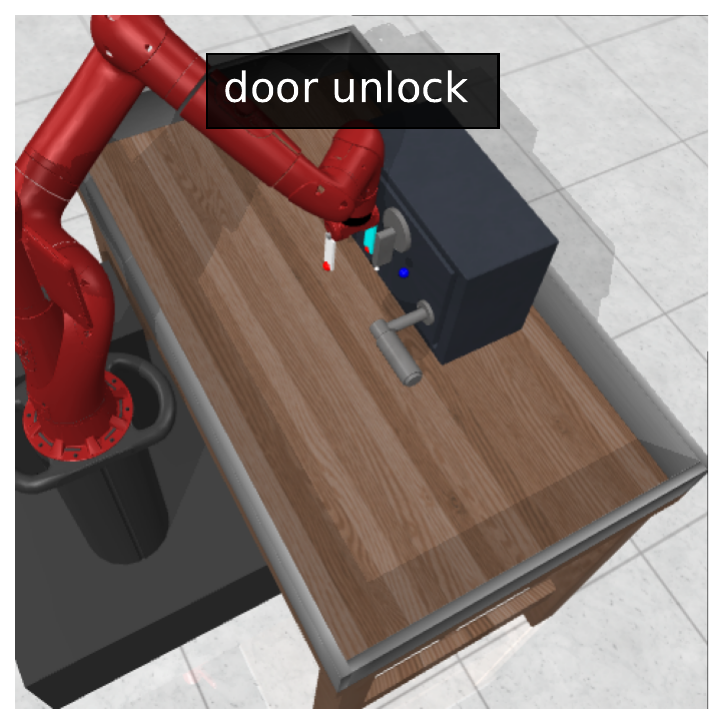} &
        \img{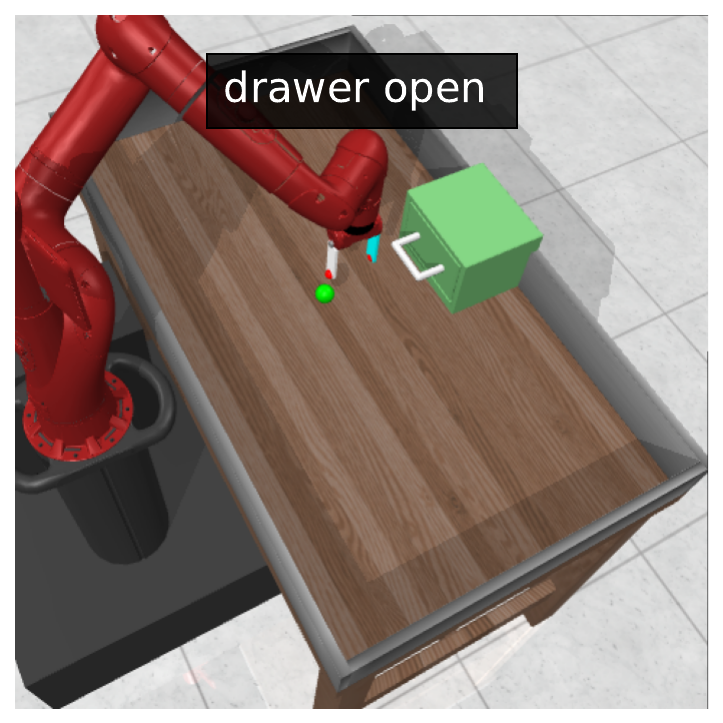} \\
        \img{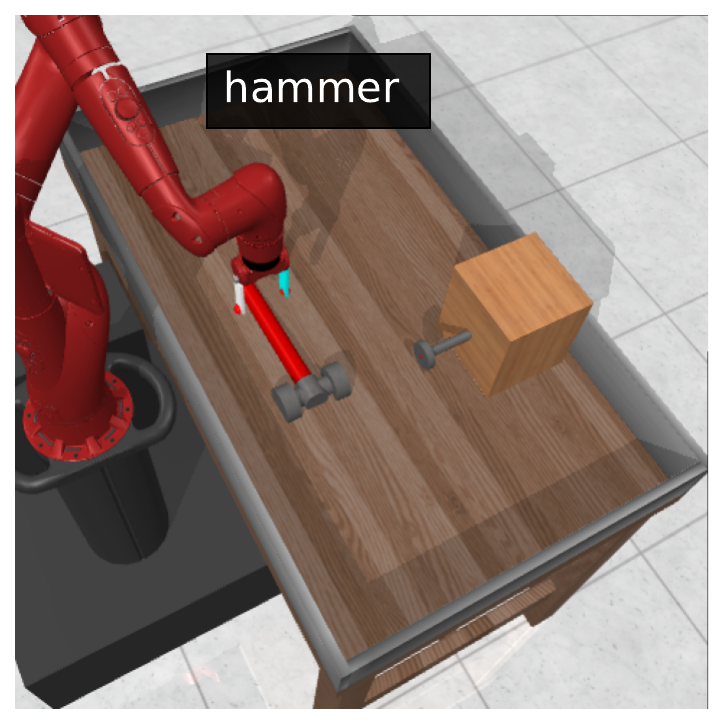} &
        \img{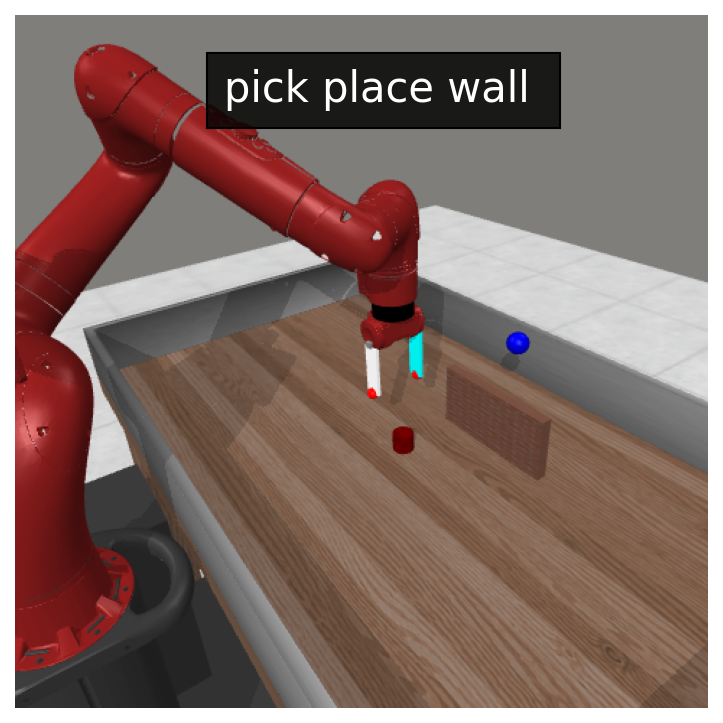}  &
        \img{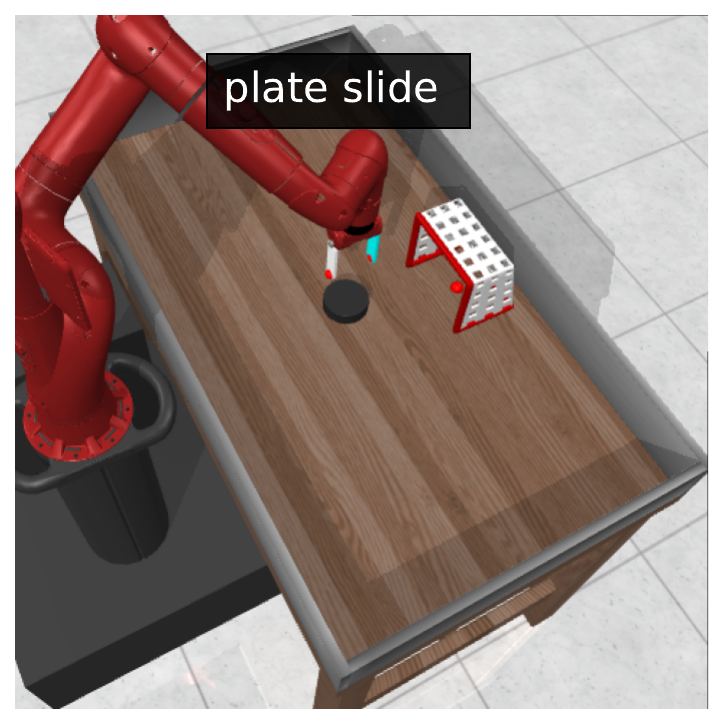}  &
        \img{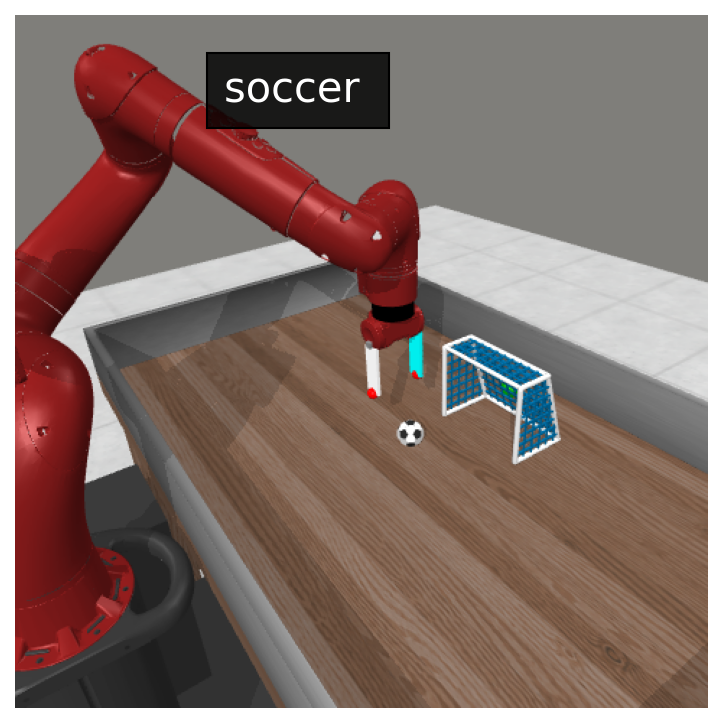}  &
        \img{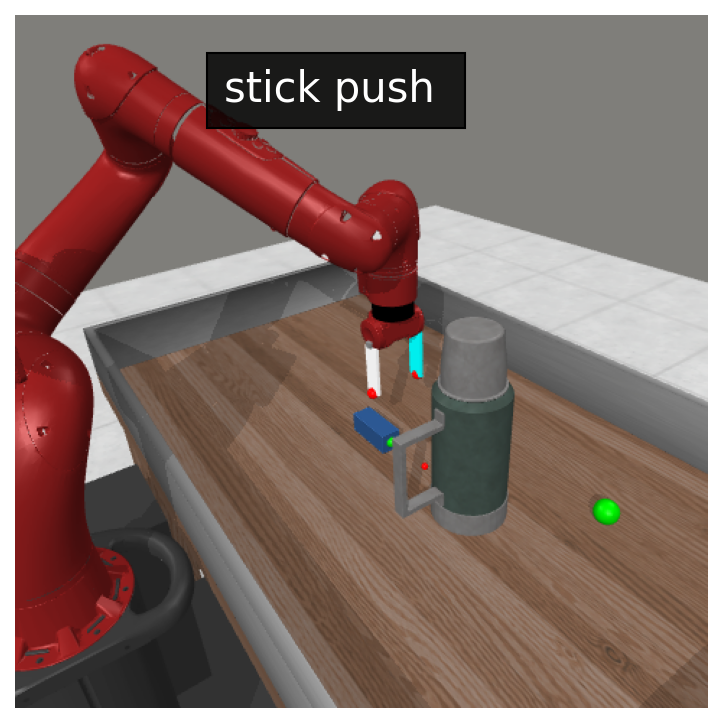}  &
        \img{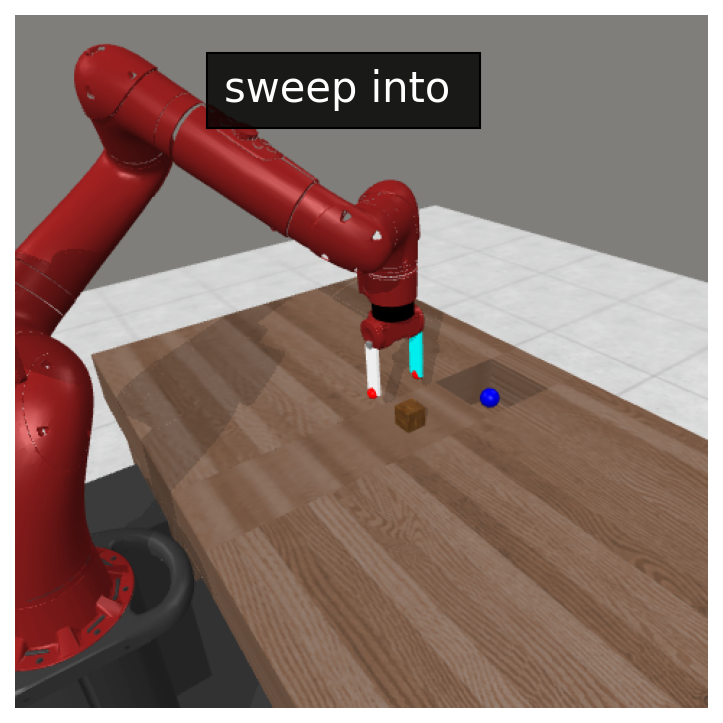} 
        \\
    \end{tabular}
    \caption{Meta-World benchmark tasks.}
    \label{fig:metawolrd-scenarios}
\end{figure}

\begin{figure}[H]
    \centering
    \newcommand{\img}[1]{\hspace{0.05cm}\includegraphics[width=0.18\linewidth]{{#1}}\hspace{0.05cm}}
    \begin{tabular}{@{}c@{}c@{}c@{}c}
        \img{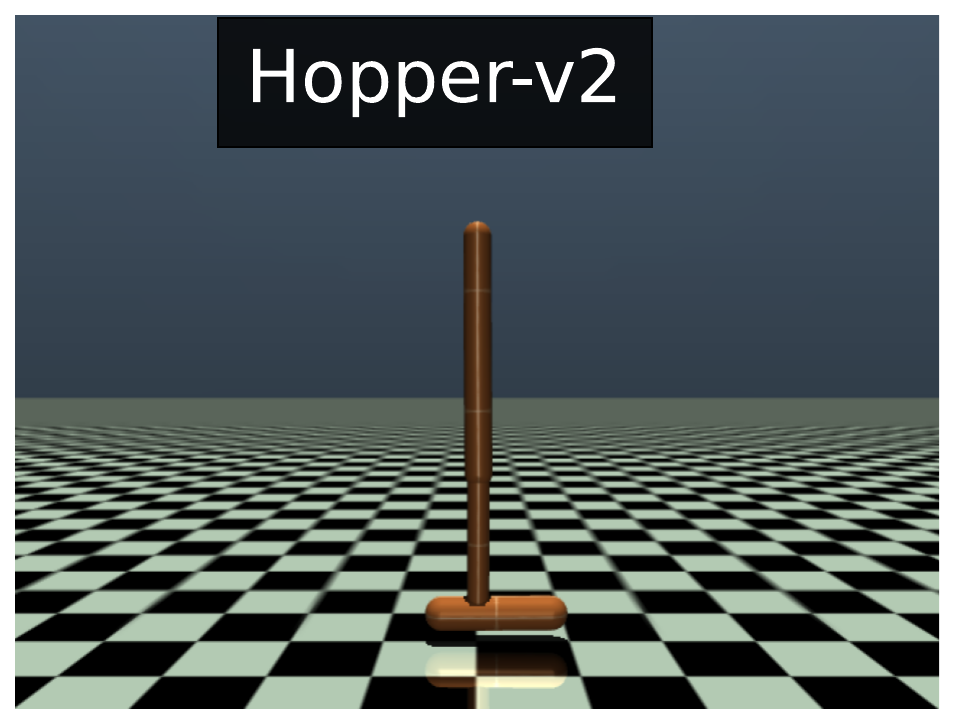} &
        \img{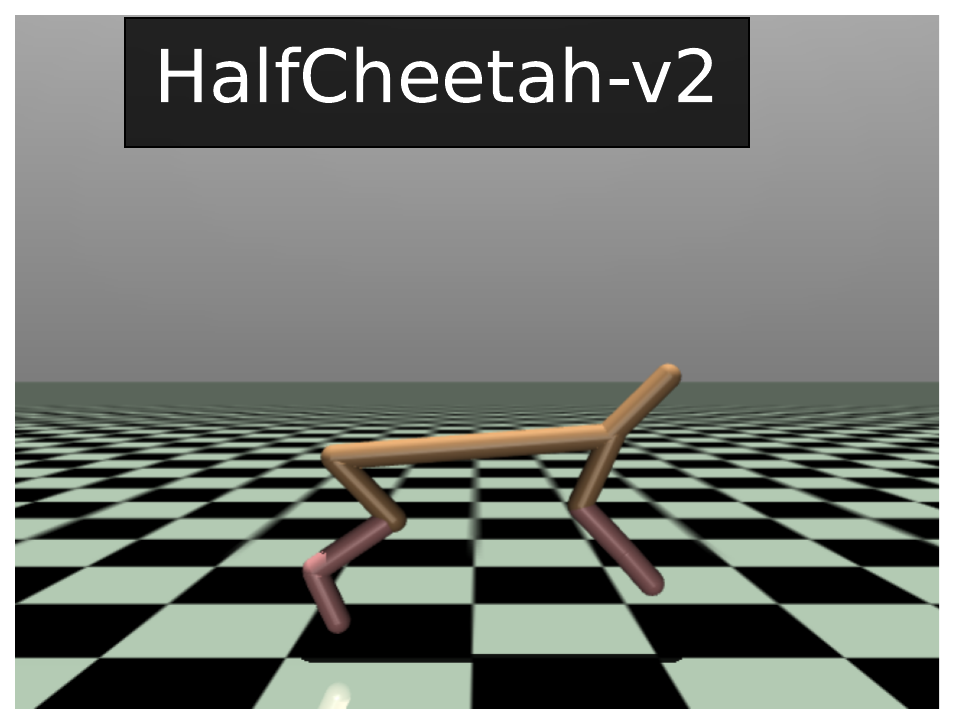} &
        \img{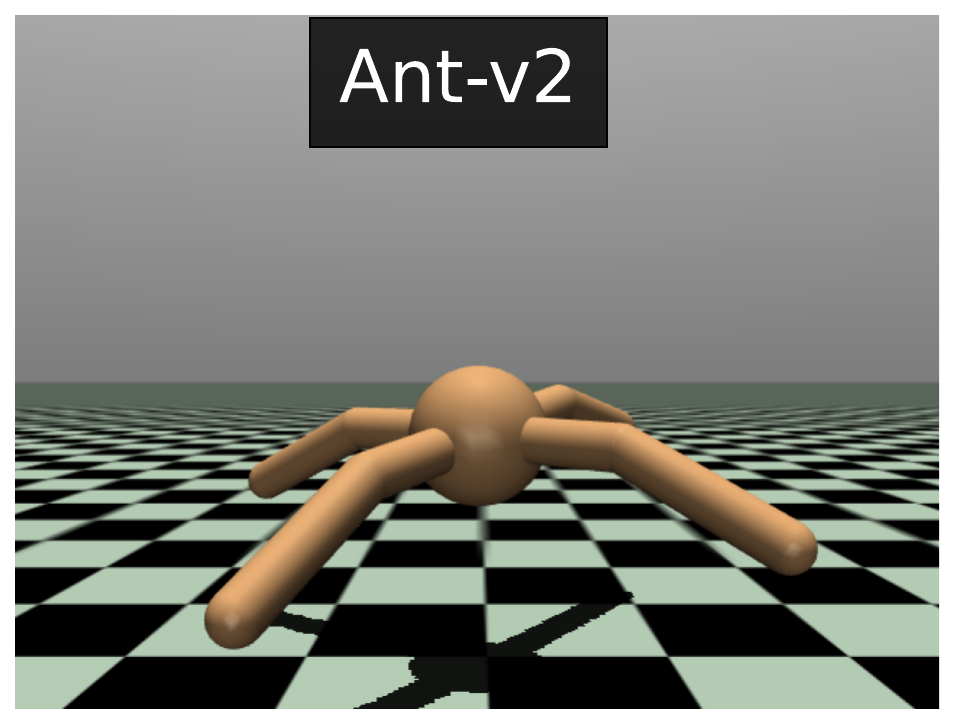} &
        \img{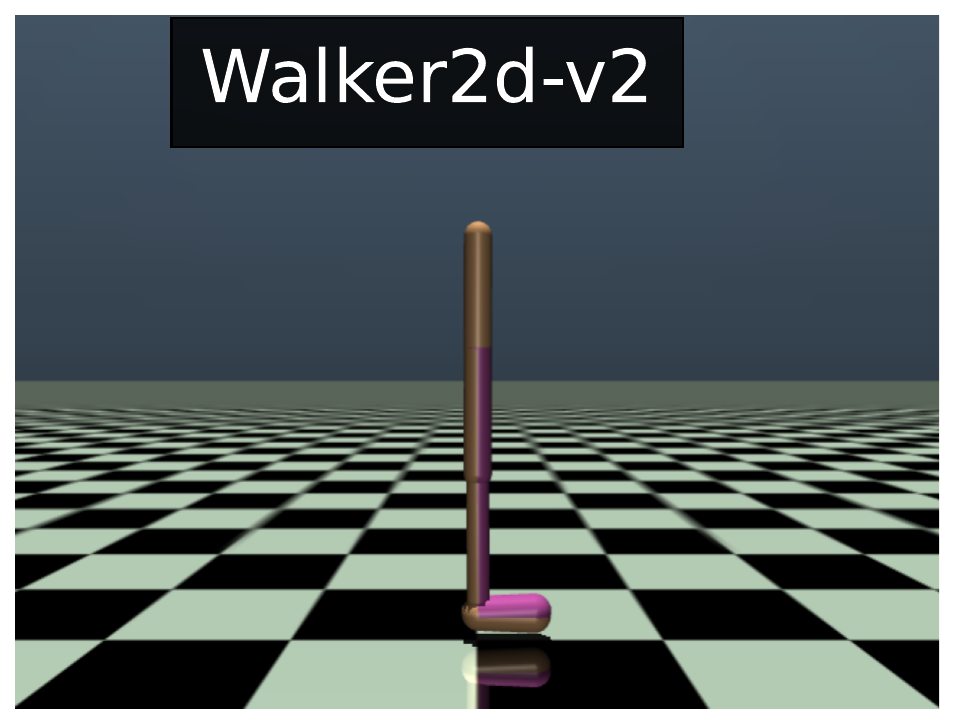}
    \end{tabular}
    \caption{MuJoCo benchmark tasks.}
    \label{fig:mujoco-scenarios}
\end{figure}

\begin{figure}[H]
    \centering
    \newcommand{\img}[1]{\hspace{0.05cm}\includegraphics[width=0.16\linewidth]{{#1}}\hspace{0.05cm}}
    \begin{tabular}{@{}c@{}c@{}c@{}c@{}c}
        \img{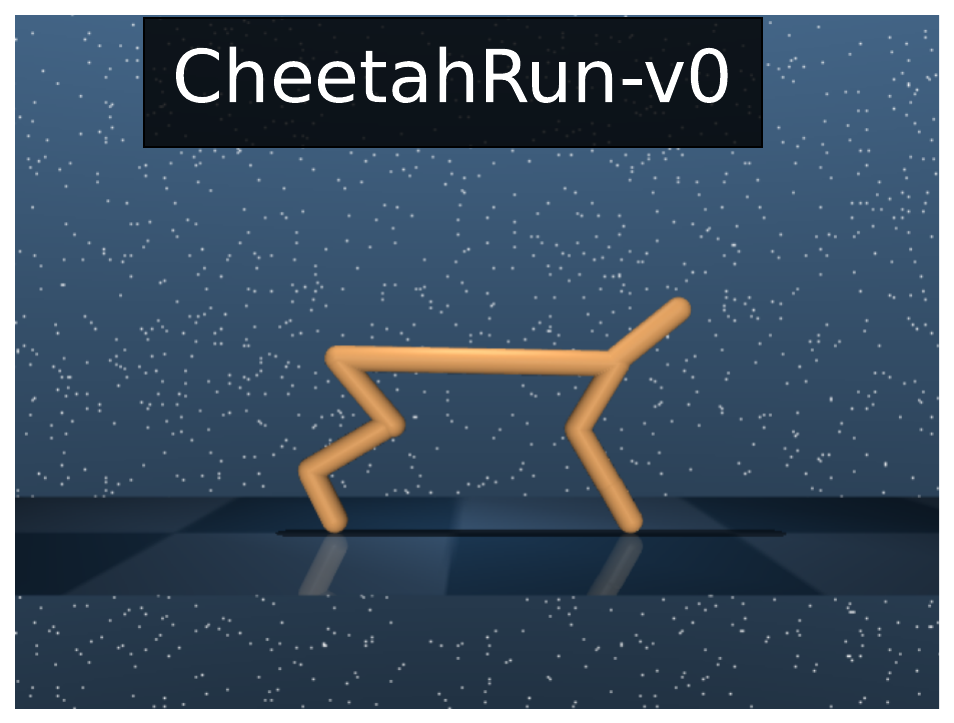} &
        \img{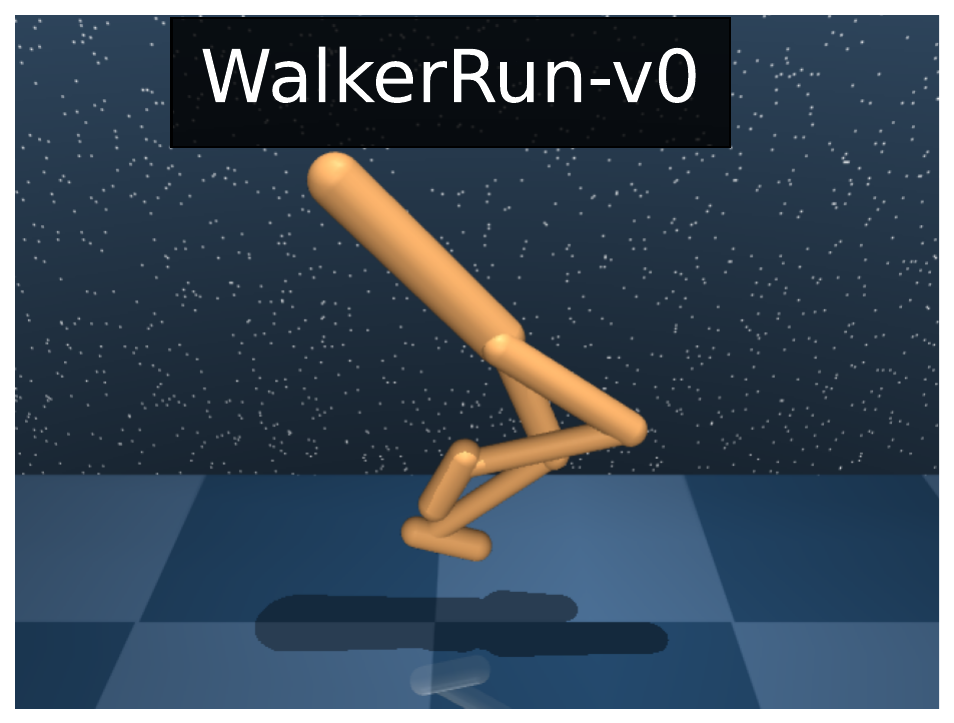} &
        \img{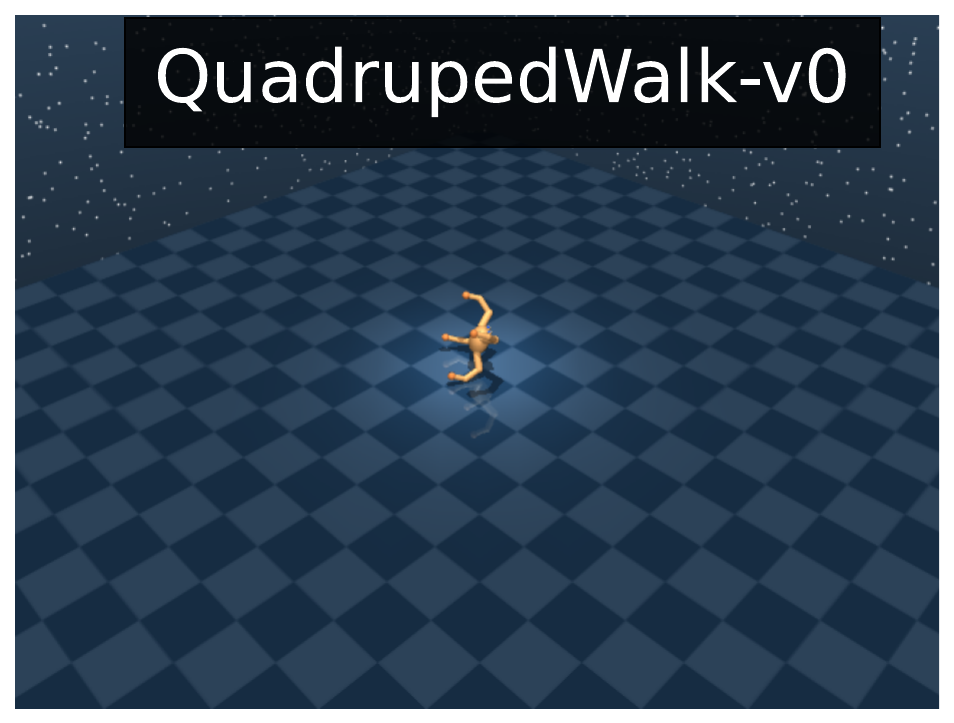} &
        \img{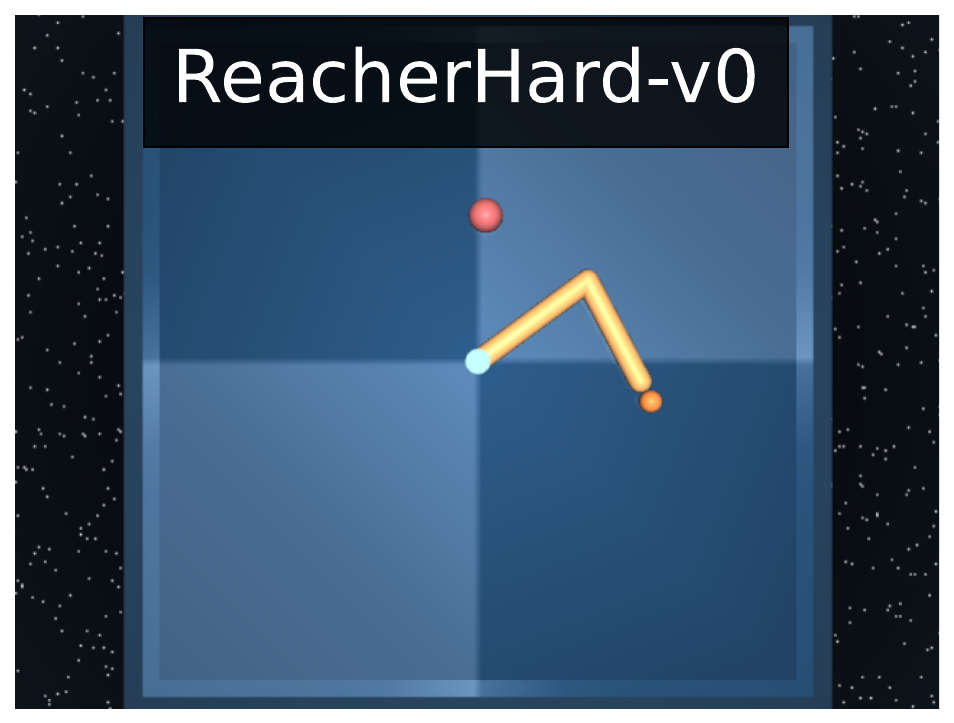}&
        \img{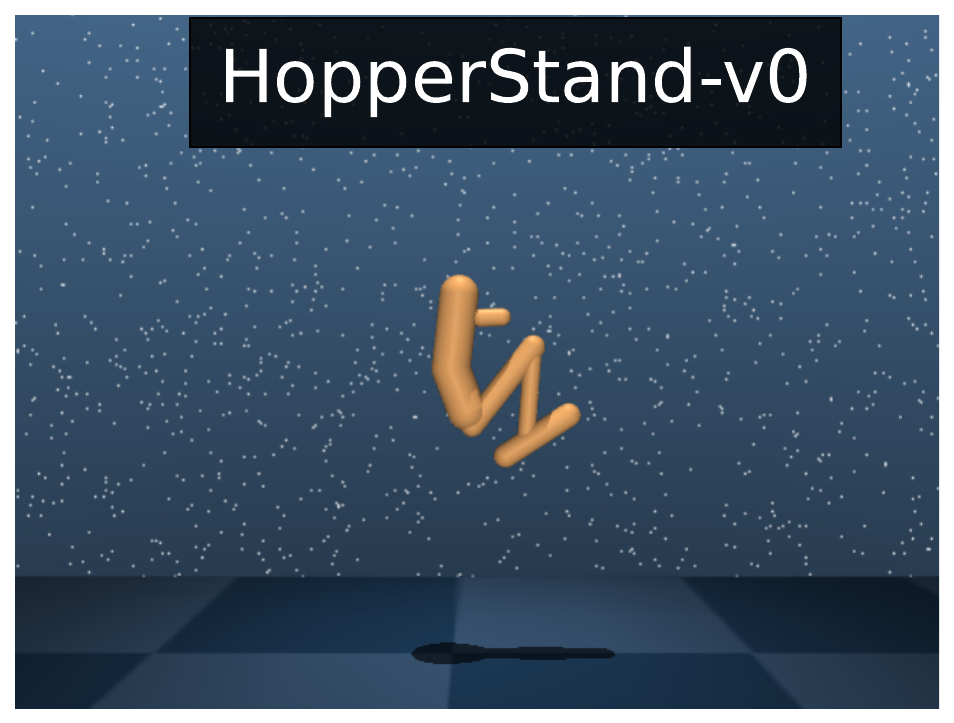}
    \end{tabular}
    \caption{DMControl benchmark tasks.}
    \label{fig:dmcontrol-scenarios}
\end{figure}

\begin{figure}[H]
    \centering
    \newcommand{\img}[1]{\hspace{0.05cm}\includegraphics[width=0.18\linewidth]{{#1}}\hspace{0.05cm}}
    \begin{tabular}{@{}c@{}c@{}c}
        \img{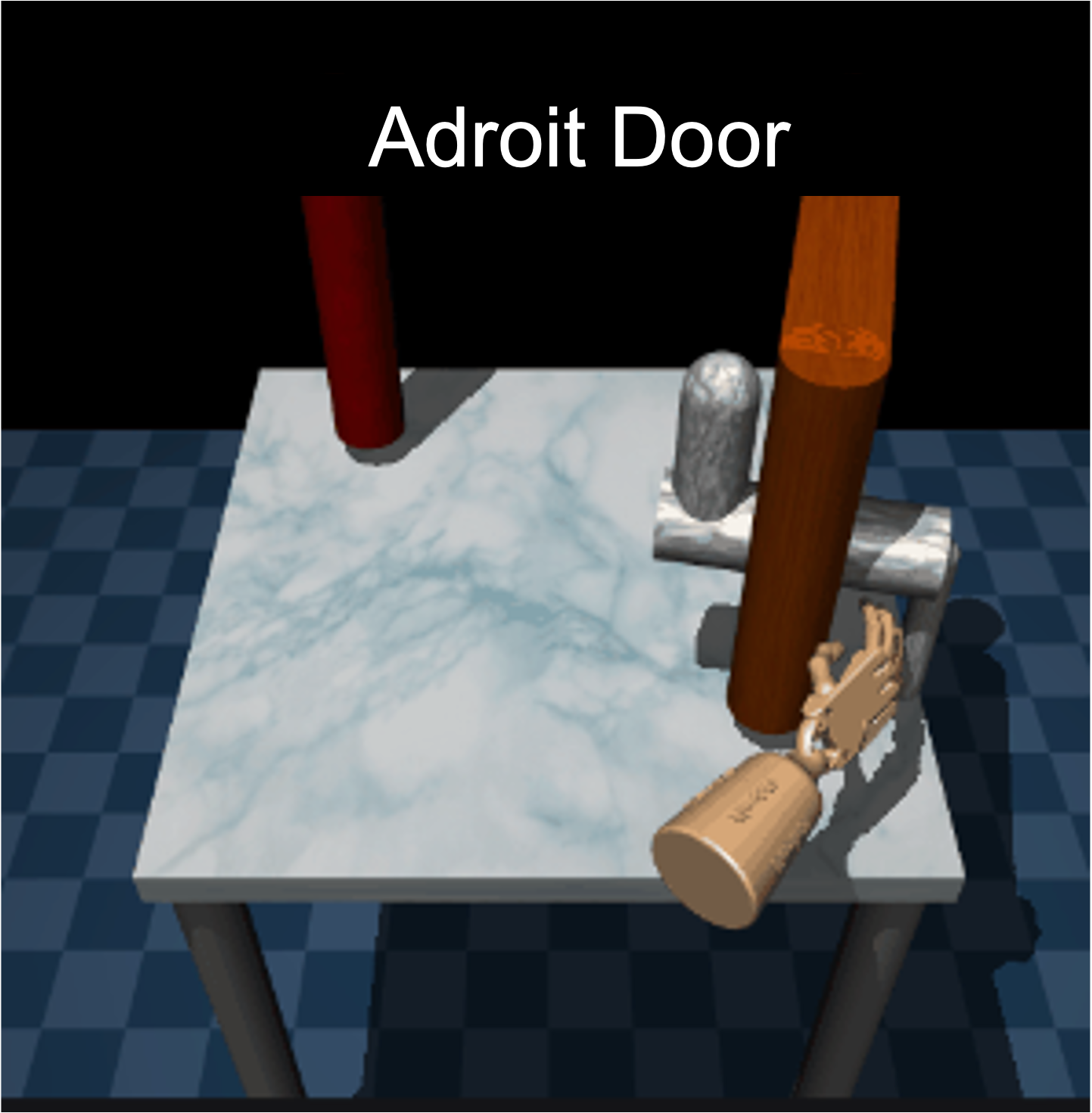} &
        \img{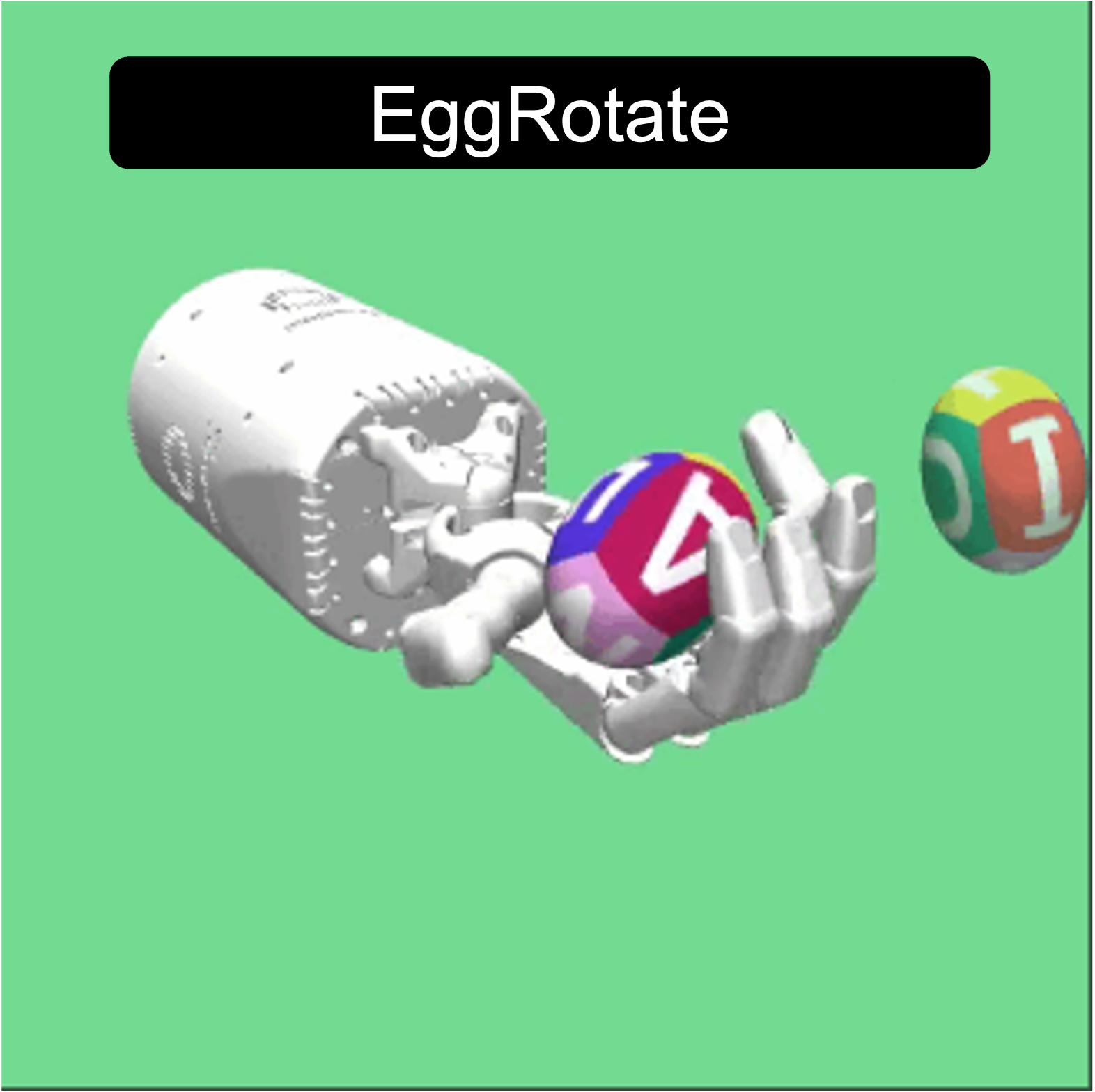} &
        \img{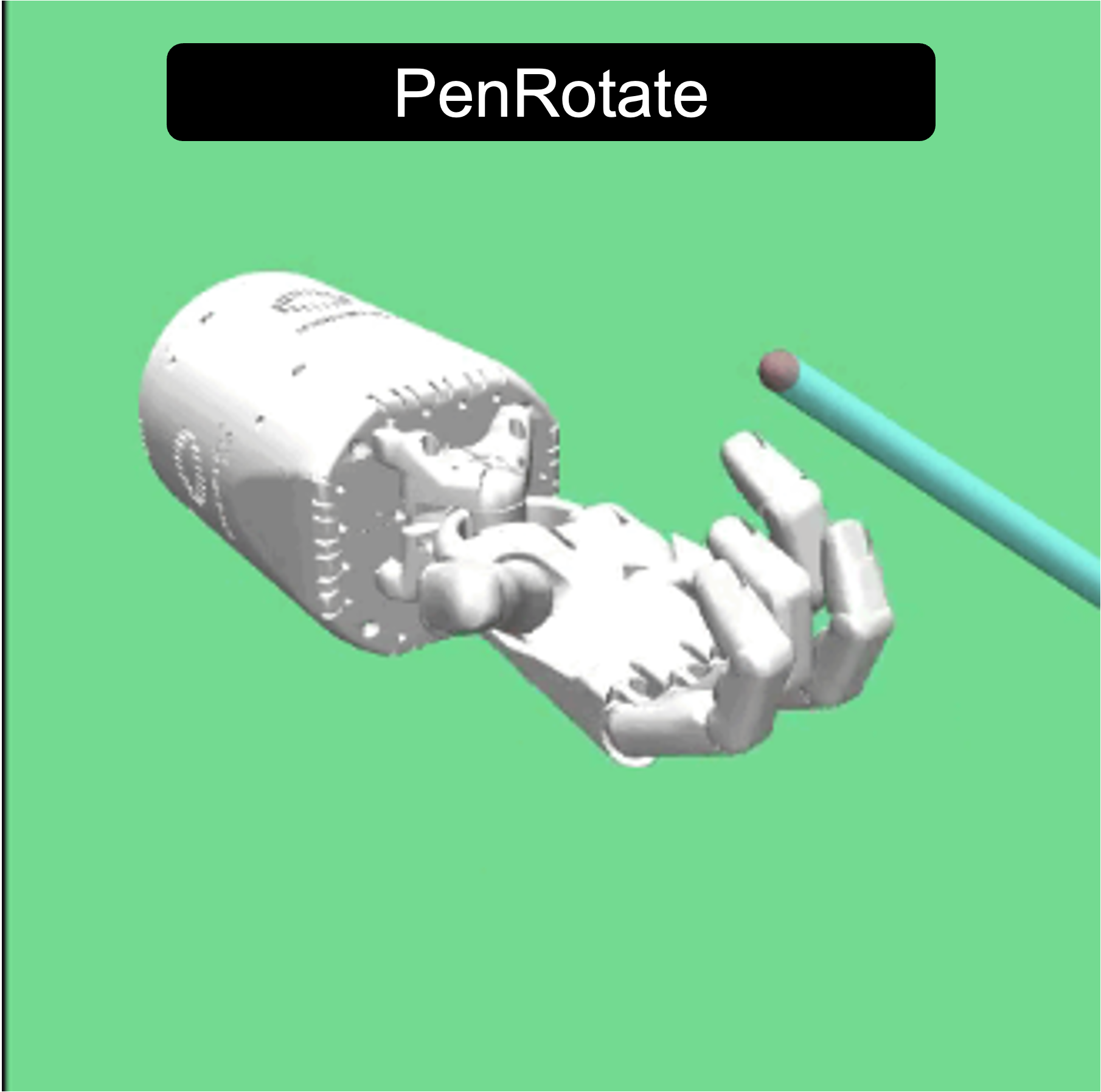} 
    \end{tabular}
    \caption{Adroit and Shadow Dexterous Hand benchmark tasks.}
    \label{fig:adroit-scenarios}
\end{figure}

\begin{figure}[H]
    \centering
    \newcommand{\img}[1]{\hspace{0.05cm}\includegraphics[width=0.18\linewidth]{{#1}}\hspace{0.05cm}}
    \begin{tabular}{@{}c@{}c@{}c}
            \img{icml2024/figures/Scenarios_door-open-v2-goal-observable.pdf} &
        \img{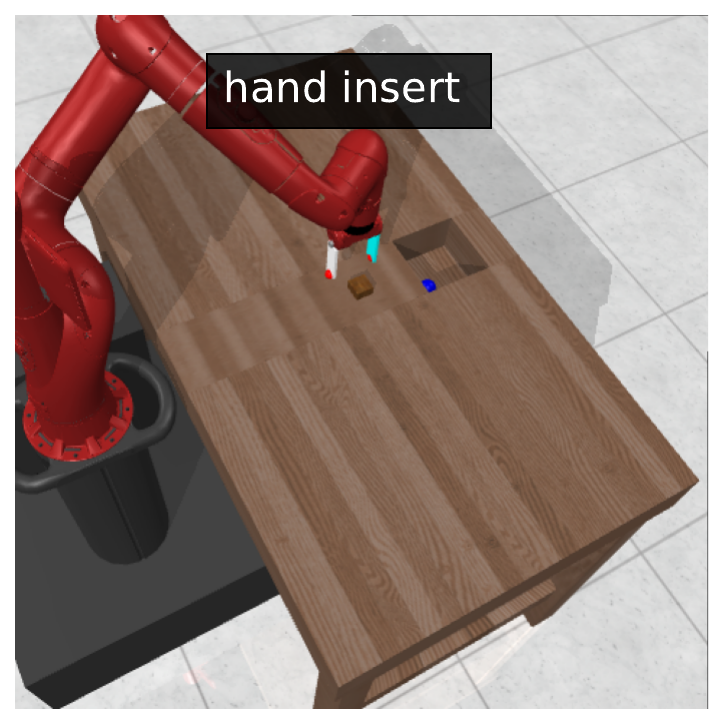} &
        \img{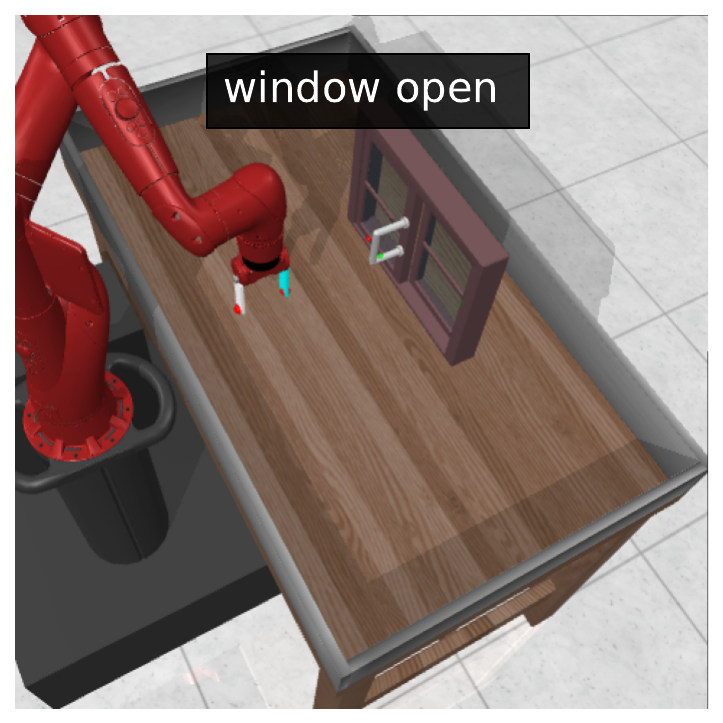} \\
        \multicolumn{3}{c}{(d) {MetaWorld benchmark tasks (sparse reward)}} \\
        {}\\   
        \img{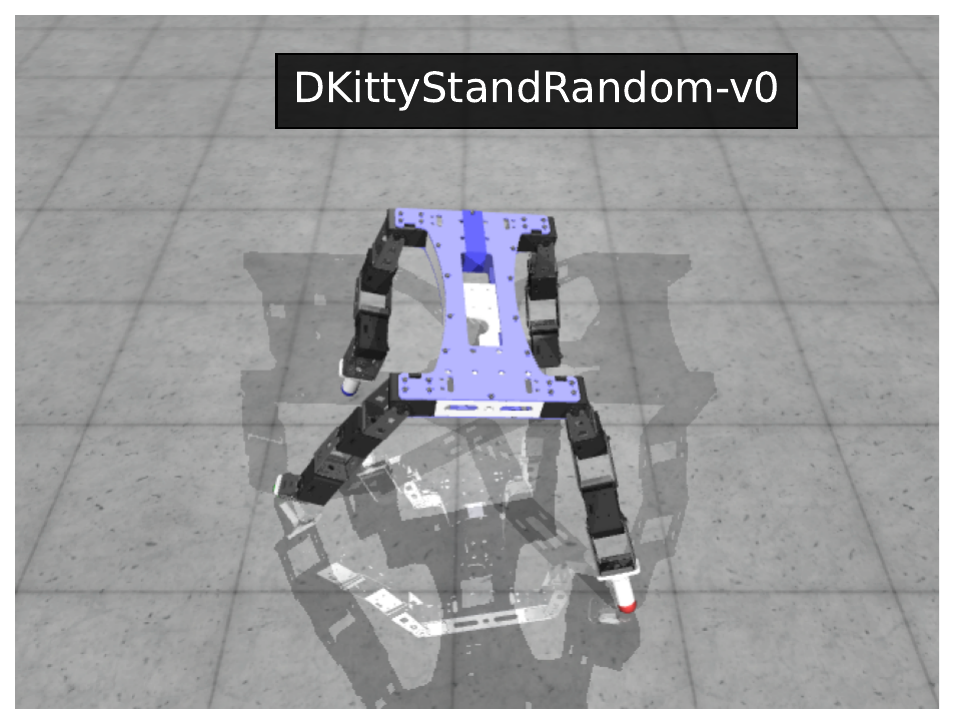} &
        \img{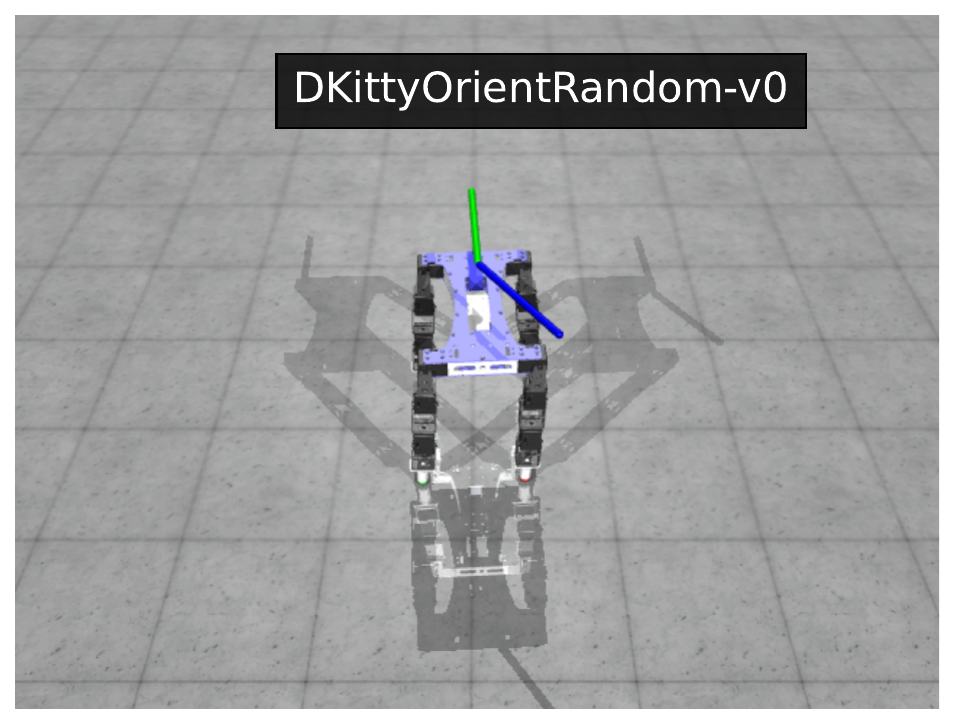} &
        \img{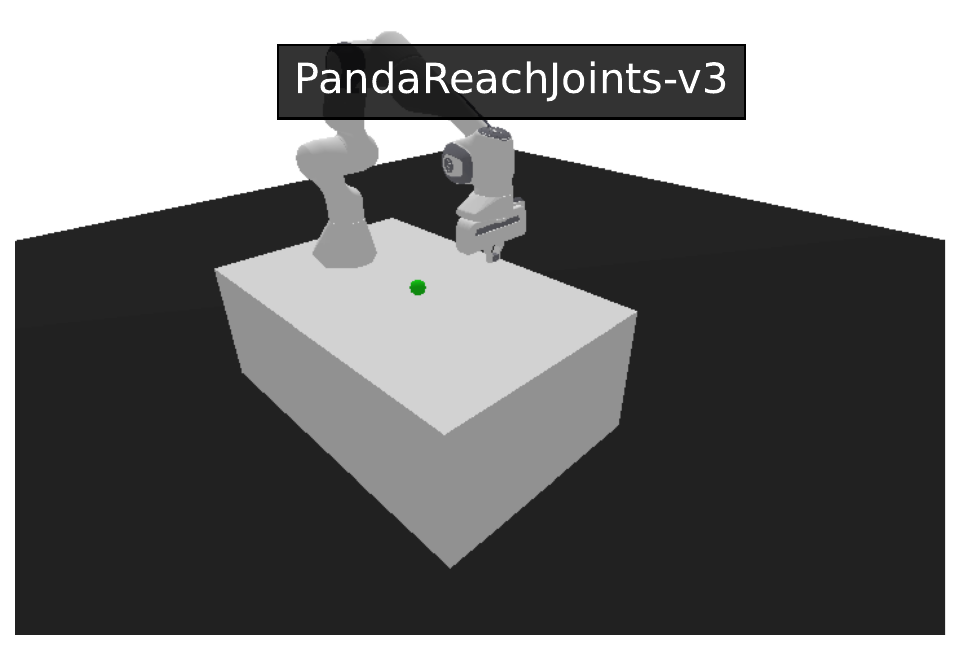} \\
        \multicolumn{2}{c}{(d) {ROBEL benchmark tasks (sparse reward)}} &
        \multicolumn{1}{c}{(e) {panda-gym benchmark tasks (sparse reward)}} \\
        {}\\   
    \end{tabular}
    \caption{ROBEL and pand-gym benchmark tasks (sparse reward).}
    \label{fig:robel-scenarios}
\end{figure}

We evaluate \ourshort\ across diverse continuous control tasks, spanning MuJoCo~\citep{todorov2012mujoco}, ROBEL~\citep{ahn2020robel}, DMControl~\citep{tassa2018deepmind}, Meta-World~\citep{yu2019meta}, Adroit~\citep{adroit}, panda-gym~\citep{panda}, Shadow Dexterous Hand~\citep{shadow}. It excels in both locomotion and manipulation tasks, both sparse reward and dense reward settings.
Visualizations of these tasks are provided in Figure~\ref{fig:metawolrd-scenarios}, Figure~\ref{fig:mujoco-scenarios}, Figure~\ref{fig:dmcontrol-scenarios}, Figure~\ref{fig:adroit-scenarios}, and Figure~\ref{fig:robel-scenarios}.

\section{More Examples of Causal Weights}
\label{sec:example}

Causal weight facilitates understanding of the agent's actions, reflecting the adaptive learning mechanisms akin to human cognition.  Here, we provide more examples for illustration.

\begin{figure*}[h]
    \centering
    \includegraphics[width=0.92\textwidth]{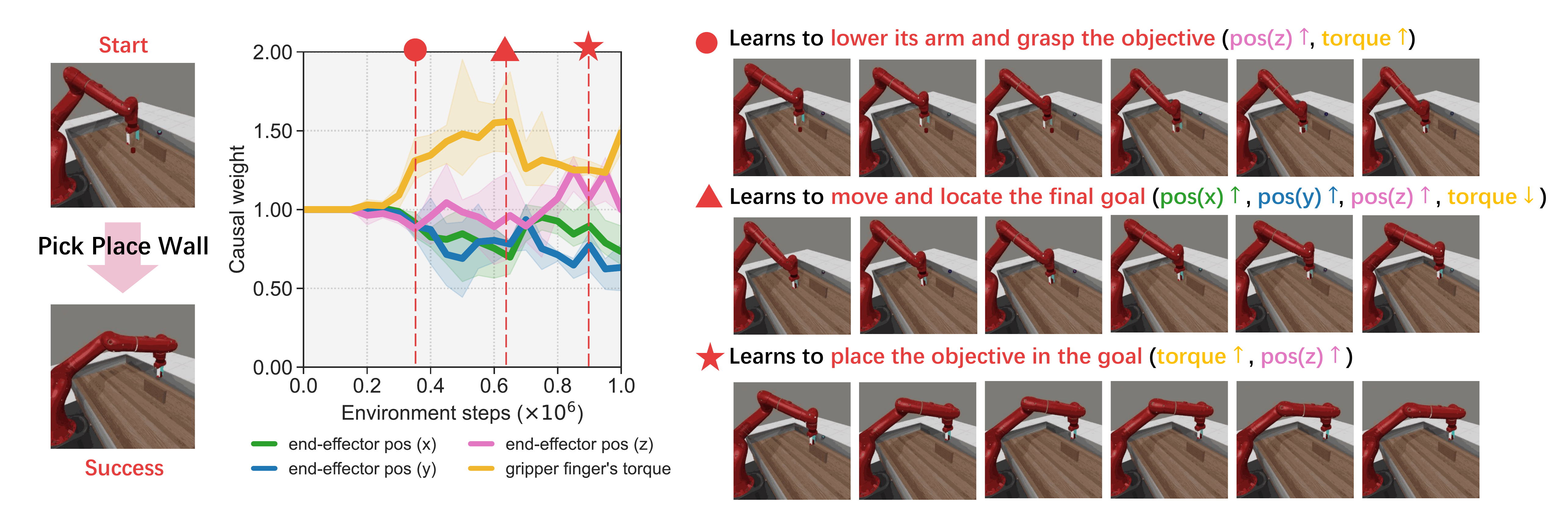}
    \vspace{-1em}
    \caption{\small \textbf{Visualization of causal weights.} We visualize the varying causal weights on the pick-place-wall task during the training stages. }
    \label{fig:causal_pick}
    \vspace{-3mm}
\end{figure*}

\begin{figure*}[h]
    \centering
    \includegraphics[width=0.92\textwidth]{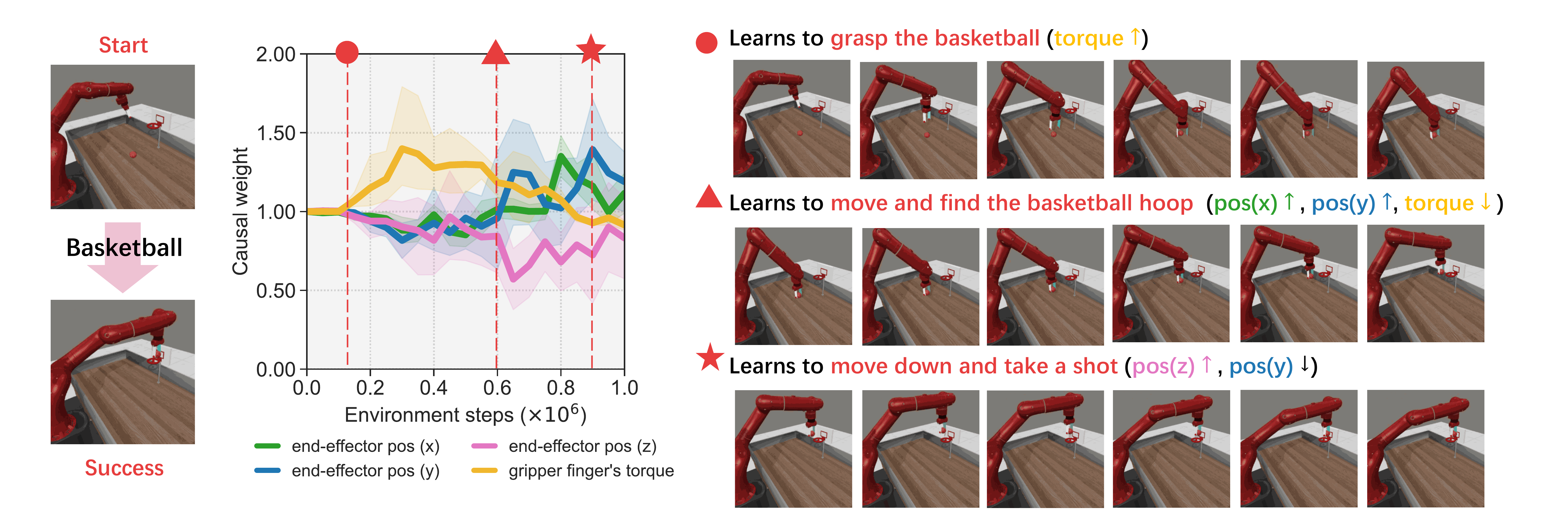}
    \vspace{-1em}
    \caption{\small \textbf{Visualization of causal weights.} We visualize the varying causal weights on the basketball task during the training stages. }
    \label{fig:causal_basketball}
    \vspace{-3mm}
\end{figure*}

\begin{figure*}[h]
    \centering
    \includegraphics[width=0.92\textwidth]{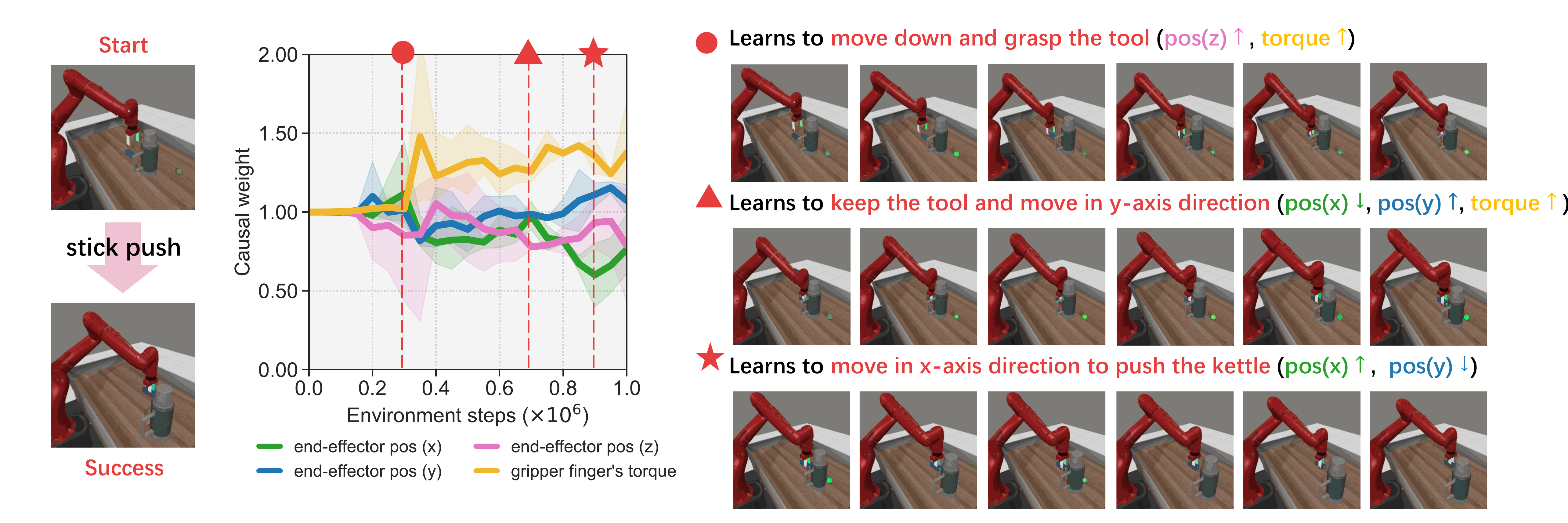}
    \vspace{-1em}
    \caption{\small \textbf{Visualization of causal weights.} We visualize the varying causal weights on the stick-push task during the training stages. }
    \label{fig:causal_stick}
    \vspace{-3mm}
\end{figure*}

\begin{figure*}[h]
    \centering
    \includegraphics[width=0.92\textwidth]{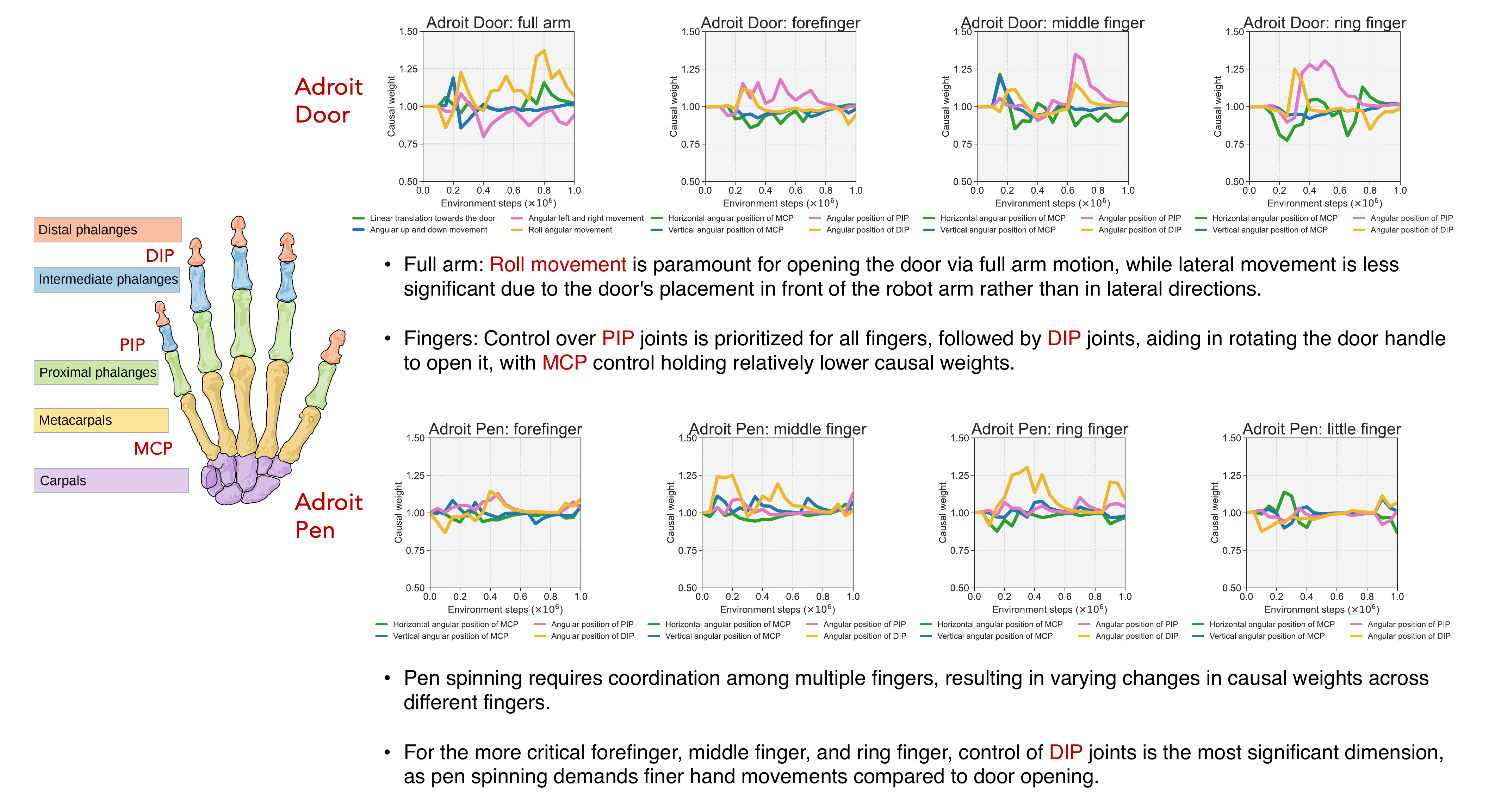}
    \vspace{-1em}
    \caption{\small \textbf{Visualizations of causal weights in Adroit Door and Pen tasks (29 action dimensions).} These curves demonstrate how causal weights control individual finger joints alongside the entire arm, with detailed explanations provided. Notably, the crucial finger joints differ across tasks. This clearly highlights the effectiveness of our causal graph calculation in high-dimensional settings involving multi-object control.}
    \label{fig:causal_stick}
    \vspace{-3mm}
\end{figure*}

\clearpage
\section{Implementation Details}
\label{sec:practicalimplementation}
\subsection{Algorithm instantiation}
Instantiating \ourshort\ involves specifying three main components: 1) effectively recognizing the causal weights of $\rva\rightarrow\rvr \vert \rvs$; 2) incorporating causal weights and the corresponding causality-aware entropy term into policy optimization. 3) periodically resetting the network based on our gradient dormancy degree.

\begin{itemize}
    \item The pseudocode of our proposed \ourshort\ is provided in Algorithm~\ref{alg:cac}.
    \item The variant named \textbf{CausalSAC} simplifies ACE by removing the reset mechanism.
    \item By incorporating the causality entropy term and the reset mechanism into the BAC backbone algorithm, we created the variant \textbf{ACE-BAC}.
\end{itemize}

\paragraph{\textcolor{myblue}{Causal discovery on $\rva\rightarrow\rvr \vert \rvs$.}} To effectively compute $\rmB_{\rva\rightarrow\rvr \vert \rvs}$, we adopt the well-regarded DirectLiNGAM method~\citep{shimizu2011directlingam}. While alternative score-based methods that simultaneously learn causal effects can also be employed, we opt for DirectLiNGAM for two main reasons: 1) Empirical validation confirms its remarkably exceptional performance, prioritizing actions with higher reward potential and aligning with human cognition in executing complex tasks. 2) Under the linearity assumption, one can straightforwardly and practically learn coefficients as causal effects. Moreover, the non-Gaussianity assumption facilitates the unique identification of the causal structure. The main implementation idea of DirectLiNGAM is as follows. In the first phase, it estimates a causal ordering for all variables of interest (i.e., state, action, and reward variables), based on the independence and non-Gaussianity characteristics of the root variable. The causal ordering is a sequence that implies the latter variable cannot cause the former one. In the second phase, DirectLiNGAM estimates the causal effects between variables using some conventional covariance-based methods such as least squares and maximum likelihood approaches. Its convergence is guaranteed theoretically under some assumptions.
Besides, we formulate a training regime wherein we iteratively adjust the causal weights for the policy at regular intervals $I$ on a local buffer $\mathcal{D}_c$ with fresh transitions to reduce computation cost.

\paragraph{\textcolor{mygreen}{Policy optimization.}} 
Given the causal weight matrix $\rmB_{\rva\rightarrow\rvr \vert \rvs}$, we could obtain the causality-aware entropy \textcolor{myred}{$\mathcal{H}_c(\pi(\cdot\vert \rvs))$}
through Eq.(\ref{eq:causal-policy-reward-entropy}). Note that to ease the computation burden of updating the causal weight matrix, we opt to conduct causal discovery with a fixed interval.

Based on the causality-aware entropy, then the $Q$-value for a fixed policy $\pi$ could be computed iteratively by applying a modified Bellman operator $\mathcal{T}_c^{\pi}$ with $\mathcal{H}_c(\pi(\cdot\vert \rvs))$ term as stated below,  
\begin{equation}
\mathcal{T}_c^{\pi} Q(\rvs_t, \rva_t) \triangleq r(\rvs_t, \rva_t) + \gamma \mathbb{E}_{\rvs_{t+1}\sim P}\left[\mathbb{E}_{\rva_t\sim \pi}[Q(\rvs_{t+1}, \rva_{t+1})+ \alpha\textcolor{myred}{\mathcal{H}_c(\pi(\rva_{t+1}\vert \rvs_{t+1}))}]\right]. 
\label{eq:bellman-op}
\end{equation}

In particular, we parameterize two $Q$-networks and train them independently, and then adopt the commonly used double-Q-techniques~\citep{van2016deep, td3, haarnoja2018soft,dac, rrs, ji2023seizing} to obtain the minimum of the $Q$-functions for policy optimization. Based on the policy evaluation, we can adopt many off-the-shelf policy optimization oracles; we chose SAC as the backbone technique primarily for its simplicity in our primary implementation of CausalSAC and \ourshort.

\paragraph{\textcolor{myyellow}{Gradient-dormancy-guided reset mechanism.}} For each reset interval, we calculate the gradient dormancy degree, initialize a random network with weights $\phi_i$ and soft reset the policy network $\pi_\theta$ and the Q network $Q_{\phi}$. 

\begin{algorithm}[t]
\caption{off-policy Actor-critic with Causality-aware Entropy~(\ours)}
\begin{algorithmic}
    \STATE \textbf{initialize:} Q network $Q_{\phi}$, policy network $\pi_\theta$, replay buffer $\mathcal{D}$; local buffer $\mathcal{D}_c$ with size $N_c$, causal weight matrix $\rmB_{\rva\rightarrow r\vert \rvs}$, perturb factor $f$; 
    
    \FOR{each environment step $t$}
        \STATE Collect data with $\pi_\theta$ from real environment
        \STATE Add to replay buffer $\mathcal{D}$ and local buffer   $\mathcal{D}_c$
    \ENDFOR
    
    \STATE
    \begin{minipage}{0.95\linewidth}
        \begin{mdframed}[linewidth=0.8pt, linecolor=myblue]
    \STATE \textcolor{myblue}{// Causal discovery }
    \IF{every $I$ environment step}
        \STATE Sample all $N_c$ transitions from local buffer $\mathcal{D}_c$
        \STATE Update causal weight matrix \textcolor{red}{$\rmB_{\rva\rightarrow r\vert \rvs}$}
    \ENDIF
    \end{mdframed}
    \end{minipage}
    \STATE

    \begin{minipage}{0.95\linewidth}
        \begin{mdframed}[linewidth=0.8pt, linecolor=mygreen]
    \STATE \textcolor{mygreen}{// Policy optimization}
    \FOR{each gradient step}
        \STATE Sample $N$ transitions $(\rvs, \rva,r,\rvs')$ from $\mathcal{D}$
        \STATE Compute causality-aware entropy \textcolor{red}{$\mathcal{H}_c(\pi(\cdot\vert \rvs))$} 
        \STATE Calculate the target $Q$ value
        \STATE Update $Q_{\phi}$ by $\min_\phi\left(\mathcal{T}_cQ_{\phi}-Q_{\phi}\right)^2$
        
        \STATE Update $\pi_\theta$ by $\max_{\theta} Q_\phi(s,a)$
    \ENDFOR
        \end{mdframed}
    \end{minipage}
    \STATE
    
    \begin{minipage}{0.95\linewidth}
    \begin{mdframed}[linewidth=0.8pt, linecolor=myyellow]
    \STATE \textcolor{myyellow}{// Reset mechanism}
    \IF {every reset interval}
        \STATE Calculate the gradient dormant degree $\beta_{\gamma}$
        \STATE Initialize a random network with weights $\phi_i$
        \STATE Soft reset $\pi_\theta$ by $\theta_t=(1-\eta) \theta_{t-1} + \eta \phi_{i}$
        \STATE Soft reset $Q_{\phi}$ by $\phi_t=(1-\eta) \phi_{t-1} + \eta \phi_{i}$
        \STATE Reset the state of the policy optimizer and Q optimizer
    \ENDIF
    \end{mdframed}
    \end{minipage}
\end{algorithmic}
\label{alg:cac}
\end{algorithm}

\subsection{Hyperparameters}
The hyperparameters used for training \ourshort\ are outlined in Table~\ref{cac-hyperparameters}. We conduct all experiments with this single set of hyperparameters.
\begin{table}[h]
    \caption{Hyperparameter settings for \ourshort.}
    \begin{center}
    \resizebox{0.7\textwidth}{!}{
        \begin{tabular}{
            >{\centering}m{0.35\textwidth}
            | c
            | c
            | c
            | c
            | c
            | c
        }
            \toprule
            \rowcolor{mylightblue}\textbf{Hyper-parameter} & \multicolumn{6}{c}{
                \textbf{Value}
            }\\
            \midrule
            $Q$-value network & \multicolumn{6}{c}{
                MLP with hidden size 512
            } \\
            \midrule
            $V$-value network & \multicolumn{6}{c}{
                MLP with hidden size 512
            } \\
            \midrule
            policy network & \multicolumn{6}{c}{
                Gaussian MLP with the hidden size 512 
            } \\
            \midrule
            discounted factor $\gamma$ & \multicolumn{6}{c}{
                0.99
            }\\
            \midrule
            soft update factor $\tau$ & \multicolumn{6}{c}{
                0.005
            }\\
            \midrule
            learning rate $\alpha$ & \multicolumn{6}{c}{
                0.0003
            }\\
            \midrule
            batch size $N$ & \multicolumn{6}{c}{
                512
            }\\
            \midrule
            policy updates per step & \multicolumn{6}{c}{
                1
            }\\
            \midrule
            value target updates interval & \multicolumn{6}{c}{
                2
            } 
            \\
            \midrule
            sample size for causality $N_c$ & \multicolumn{6}{c}{
                10000
            } 
            \\
            \midrule
            causality computation interval $I$ & \multicolumn{6}{c}{
                10000
            } \\
            \midrule
            max reset factor $\alpha_{max}$ & \multicolumn{6}{c}{
                0.8
            } \\
            \midrule
            reset interval & \multicolumn{6}{c}{
                200000
            }\\
            \midrule
            dormancy threshold $\tau$ & \multicolumn{6}{c}{
            0.025
            }\\
            \bottomrule
        \end{tabular}
        }
    \end{center}
    \label{cac-hyperparameters}
\end{table}
\clearpage
\section{More Benchmark Results}\label{app:morebenchmarkresults}
We conduct experiments on the more complex locomotion and manipulation tasks from \textbf{MuJoCo}~\citep{mujoco}, \textbf{DMControl}~\citep{dmc}, \textbf{Meta-World}~\citep{yu2019meta}, \textbf{Adroit}~\citep{adroit}, \textbf{Shadow Dexterous Hand}~\citep{shadow} for further evaluation of \ourshort and the baselines. Currently, several tasks in these benchmarks pose a formidable challenge that stumps most model-free methods.  Notably, \ourshort\ has demonstrated its effectiveness by successfully solving many of these challenging tasks. 

\subsection{Evaluation on MetaWorld benchmark tasks}\label{section:metaworld_benchmark}

\begin{figure}[H]
    \centering
    \includegraphics[height=10.5cm,keepaspectratio]{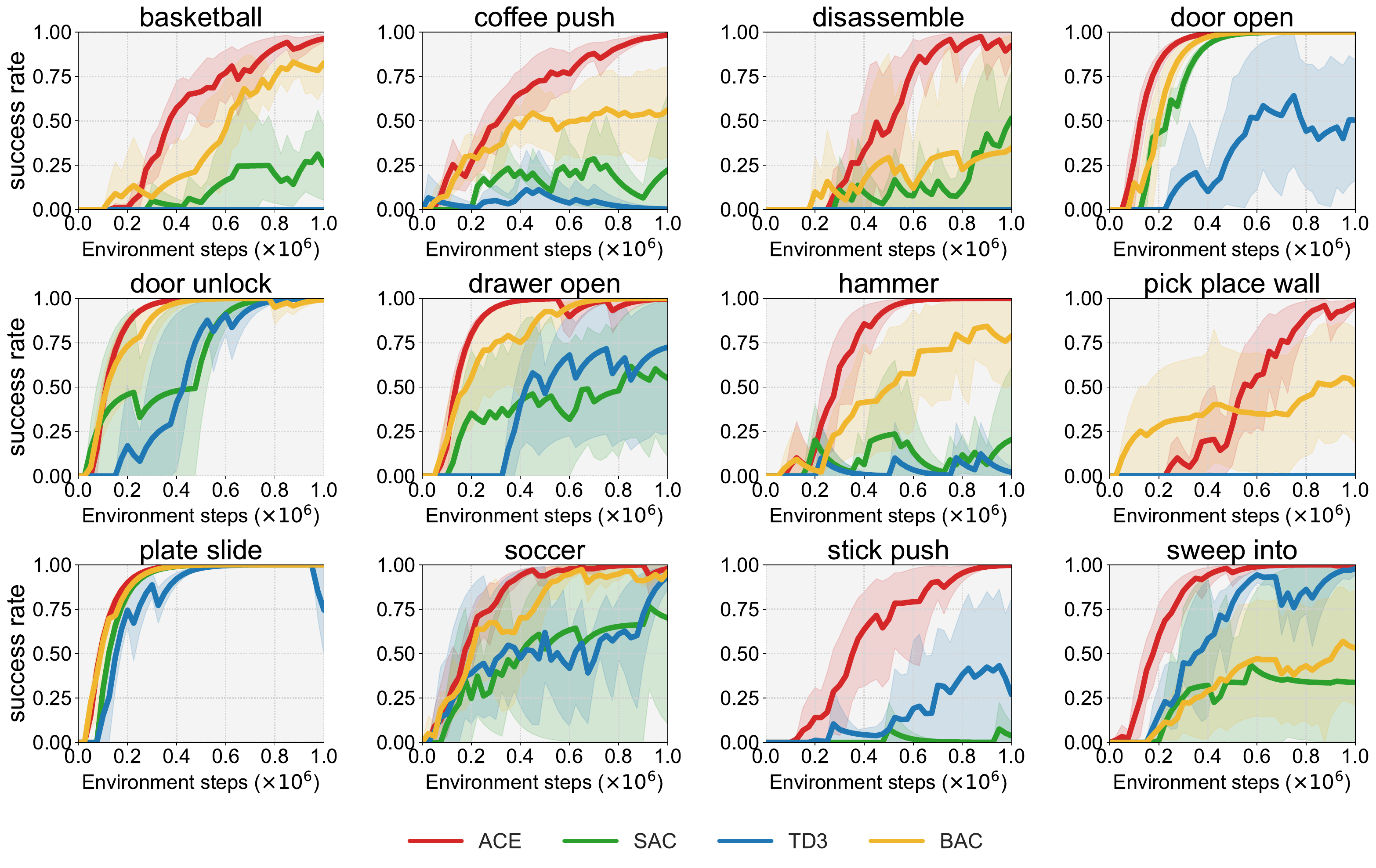} 
    \caption{\textbf{MetaWorld tasks.} Success rate of \ourshort, BAC, SAC, TD3 in MetaWorld tasks. Solid curves depict the mean of 6 trials, and shaded regions correspond to the one standard deviation. }
    \label{fig:metaworld}
\end{figure}

\subsection{Evaluation on MuJoCo benchmark tasks}\label{section:mujoco_benchmark}
\begin{figure}[H]
    \centering
    \includegraphics[height=3.8cm,keepaspectratio]{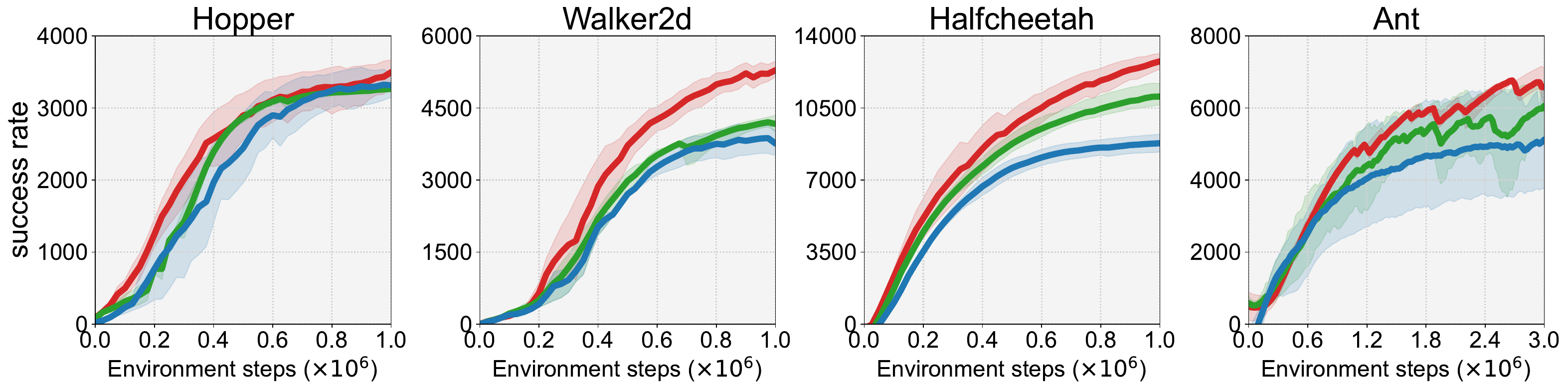} 
    \\
    \includegraphics[height=0.55cm,keepaspectratio]{icml2024/figures/main_legend.pdf}
    \caption{\textbf{MuJoCo benchmark tasks.} Training curves of \ourshort, SAC, TD3 in MuJoCo benchmark tasks. Solid curves depict the mean of 6 trials and shaded regions correspond to the one standard deviation. }
    \label{fig:mujoco}
\end{figure}

\subsection{Evaluation on DMControl benchmark tasks}\label{section:dmcontrol_benchmark}
\begin{figure}[H]
    \centering
    \includegraphics[height=3.5cm,keepaspectratio]{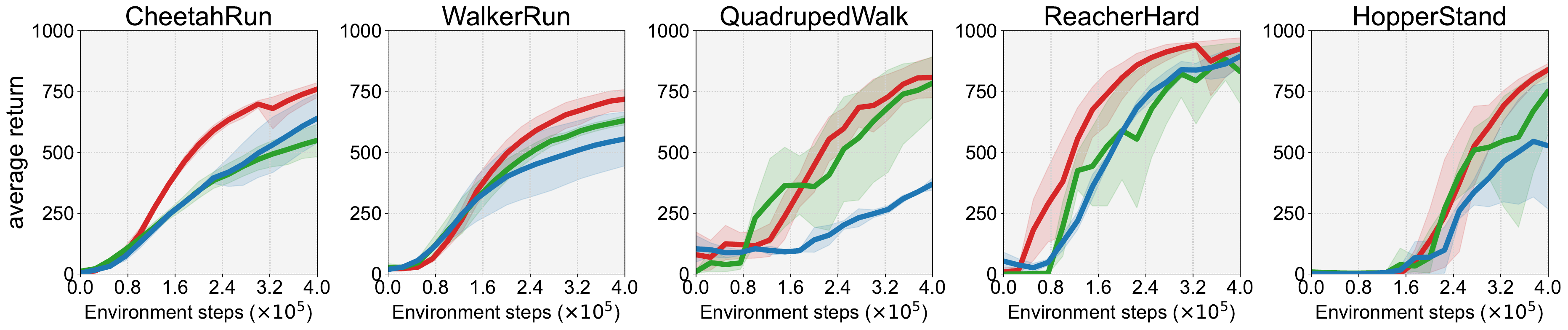} 
    \\
    \includegraphics[height=0.55cm,keepaspectratio]{icml2024/figures/main_legend.pdf}
    \caption{\textbf{DMControl benchmark tasks.} Training curves of \ourshort\ , SAC, TD3 in DMControl benchmark tasks. Solid curves depict the mean of 6 trials and shaded regions correspond to the one standard deviation. }
    \label{fig:dmcontrol}
\end{figure}

\subsection{Evaluation on Adroit and Shadow Dexterous Hand benchmark tasks}\label{section:shadowhand_benchmark}

\begin{figure}[H]
    \centering
    \includegraphics[height=3.5cm,keepaspectratio]{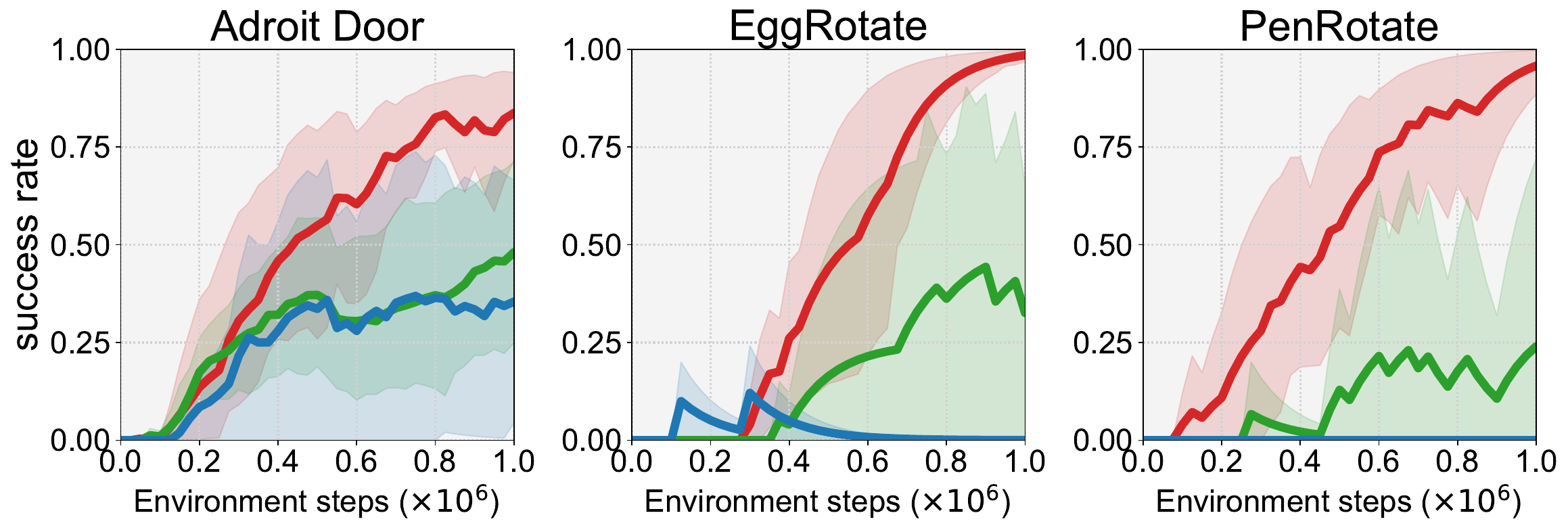} \\
    \includegraphics[height=0.55cm,keepaspectratio]{icml2024/figures/main_legend.pdf}
    \caption{\textbf{Dexterous hand tasks.} Training curves of \ourshort\ , SAC, TD3 in dexterous hand tasks. Solid curves depict the mean of 6 trials, and shaded regions correspond to the one standard deviation. }
    \label{fig:dex_hand}
\end{figure}

\section{Extensive Ablation Studies on Locomotion Tasks}
\label{apsec:abl_locomotion}
\begin{figure}[t]
    \centering
    \includegraphics[height=3.8cm,keepaspectratio]{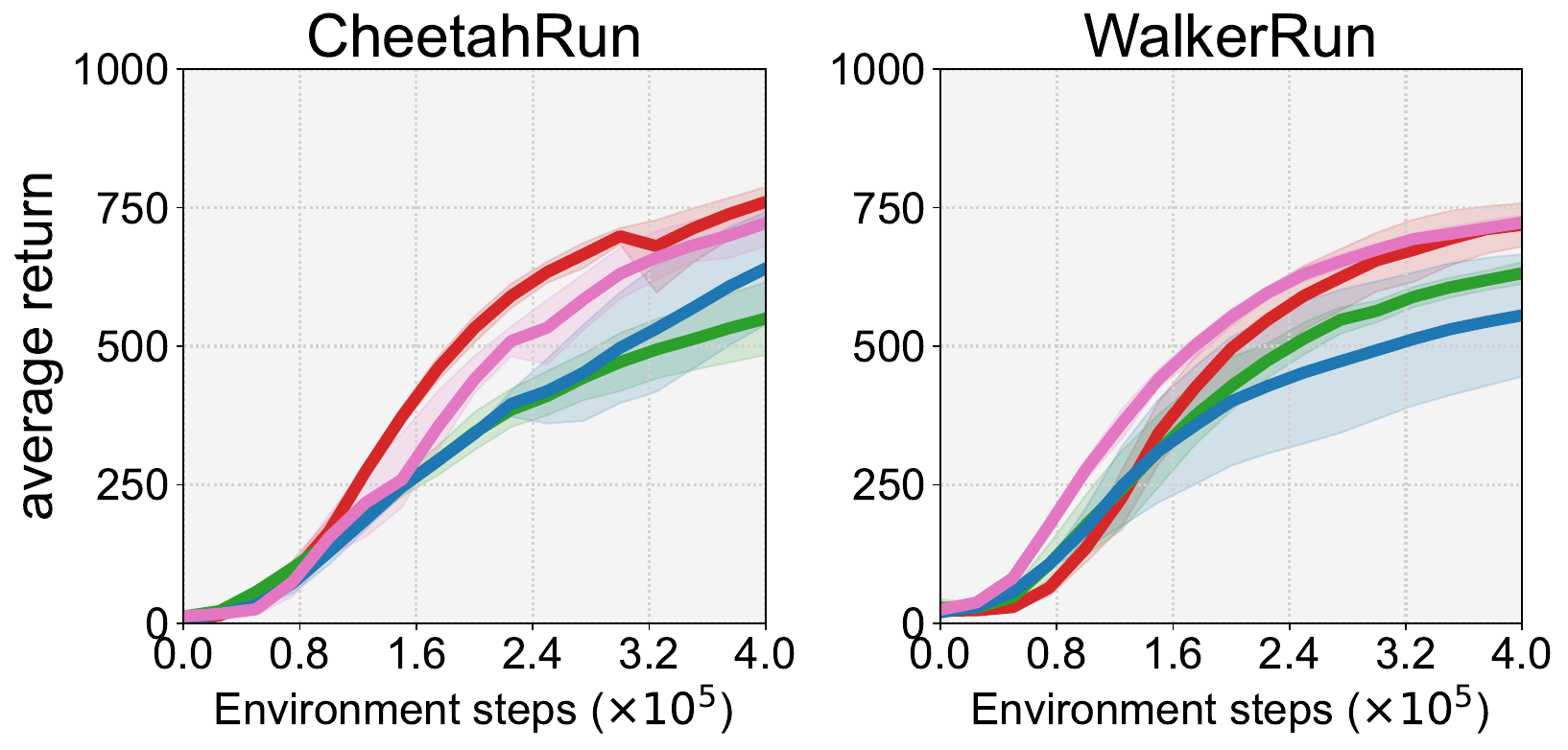} 
     \\
    \includegraphics[height=0.65cm,keepaspectratio]{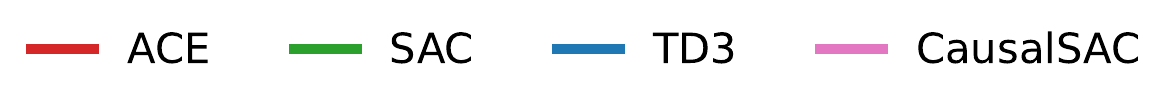}
    \caption{\textbf{Ablation of reset mechanism on locomotion tasks.} Average return of \ourshort, CausalSAC, SAC, TD3 in two locomotion tasks. CausalSAC is a reduced form of ACE by eliminating the reset mechanism of ACE.}
    \label{fig:locomotion_ace_cac_compare}
\end{figure}

We present the results of CausalSAC on a subset of locomotion tasks, demonstrating comparable or even superior performance compared to \ours. Due to the fact that the total number of steps in these tasks is only 40\% of that in manipulation tasks, the policy converges faster. We attribute this to the key reason for the less pronounced impact of our reset mechanism in these tasks. Nevertheless, it is worth noting that the causal weighted entropy we propose still exhibits significant significance in improving sample efficiency in locomotion tasks.

\section{Effectiveness in Sparse Reward Settings}\label{apsec:sparese_reward}
\begin{figure}[t]
    \centering
    \includegraphics[height=7.8cm,keepaspectratio]{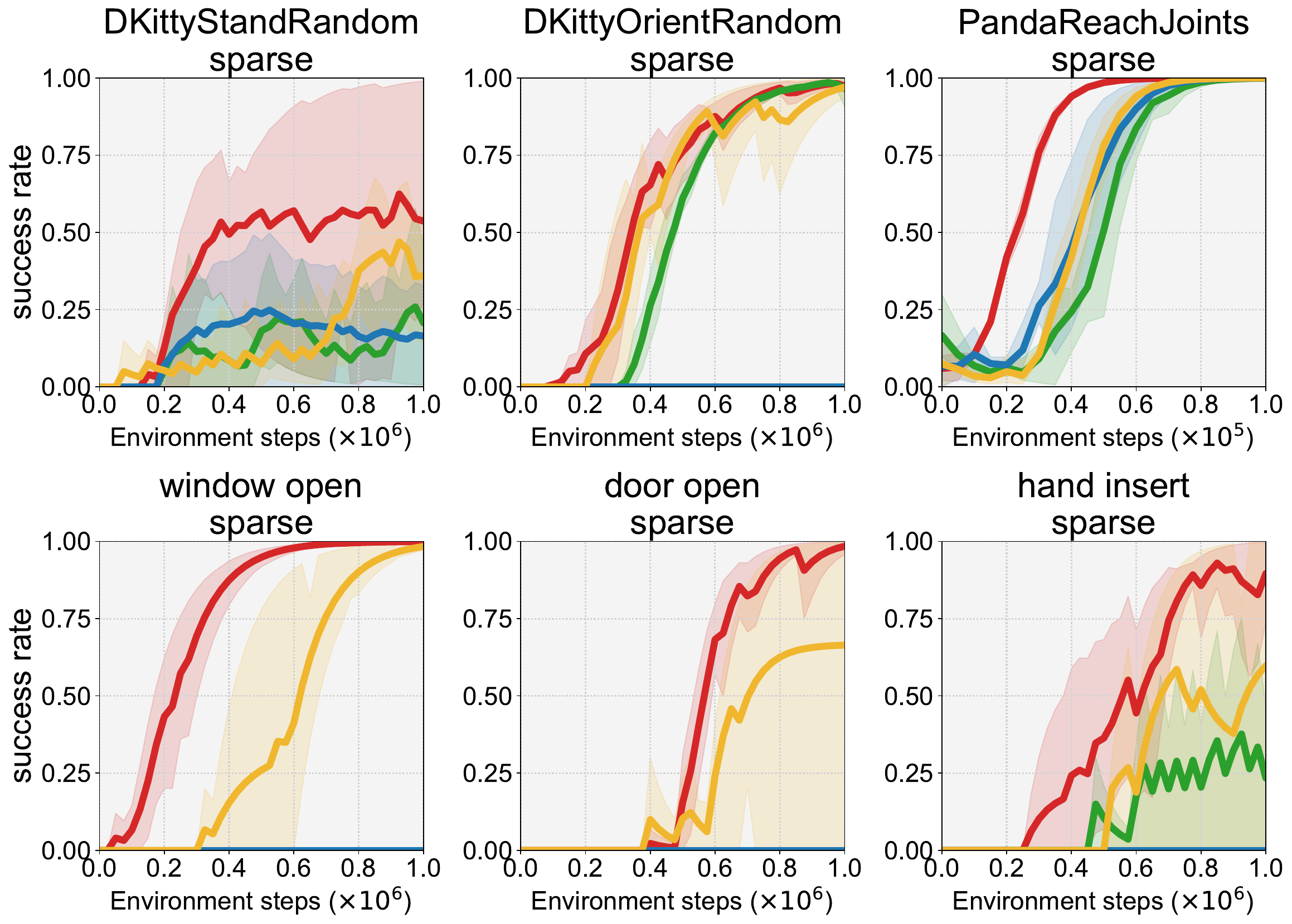} \\
    \includegraphics[height=0.6cm,keepaspectratio]{icml2024/figures/main_legend_RND.pdf}
    \caption{\textbf{Sparse reward tasks.} Training curves of \ourshort, SAC, TD3 and SAC+RND in sparse reward tasks. Solid curves depict the mean of 6 trials, and shaded regions correspond to the one standard deviation. }
    \label{fig:sparse_reward}
\end{figure}

We conduct experiments in sparse reward tasks to showcase the efficiency of \ourshort. We evaluate both robot locomotion and manipulation tasks based on the sparse reward version of benchmark tasks from MetaWorld~\citep{metaworld}, ROBEL~\citep{ahn2020robel} and panda-gym~\citep{gallouedec2021panda}. MetaWorld manipulation tasks are based on a Shawyer robot arm with end-effector control. Panda-gym manipulation tasks are based on a Franka Emika Panda robot with joint angle control.  ROBEL quadruped locomotion tasks are based on a D’Kitty robot with 12 joint positions control. As shown in Figure \ref{fig:sparse_reward}, our \ourshort\ surpasses the baselines by a large margin.

\section{Generalizability of ACE}\label{apsec:cbac_experiments}
The proposed causality-aware entropy and gradient-dormancy-guided reset mechanism are versatile and effective plug-ins, thus they could be integrated into various RL backbone algorithms and techniques.

\begin{figure}[h]
    \centering
    \includegraphics[height=3.8cm,keepaspectratio]{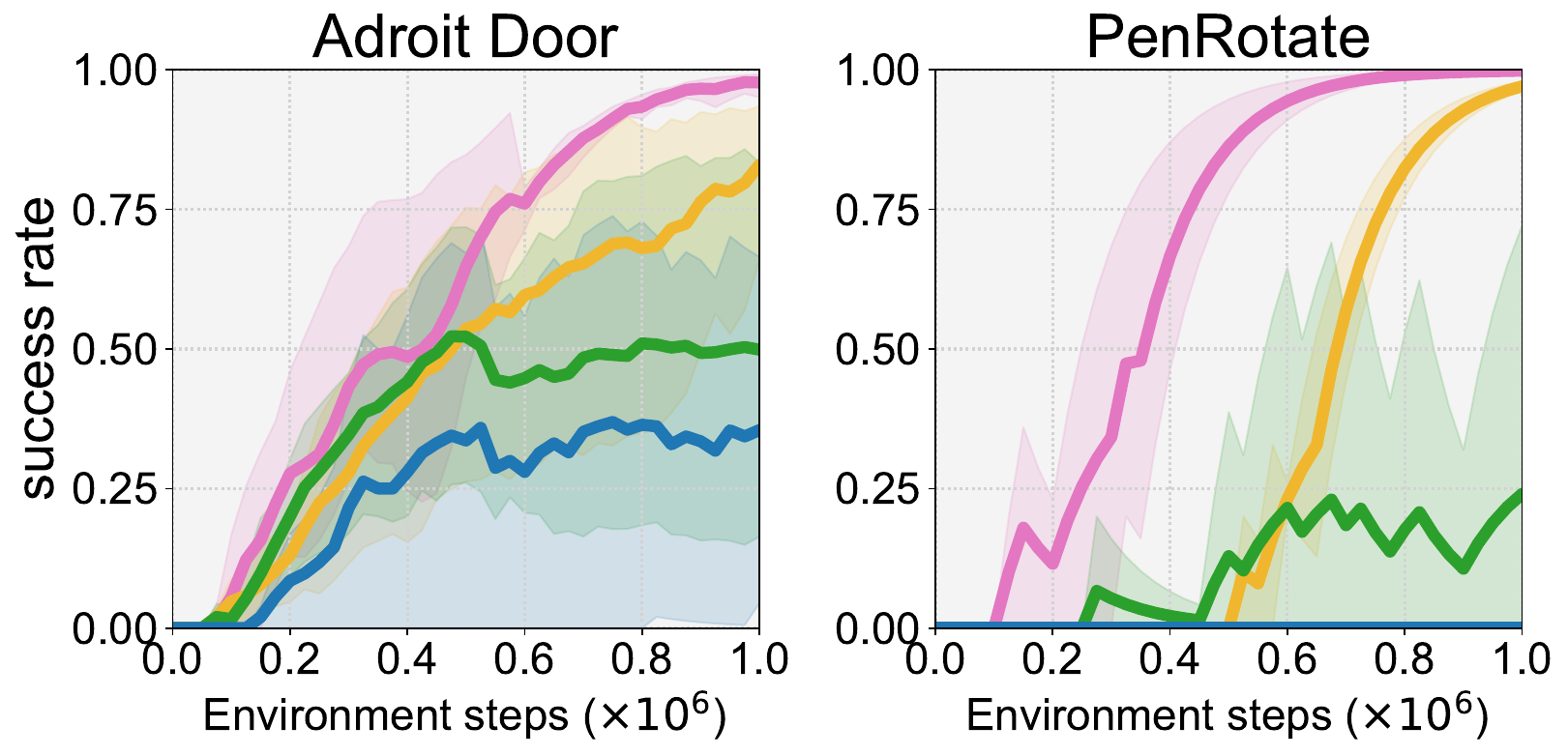} \\
    \includegraphics[height=0.6cm,keepaspectratio]{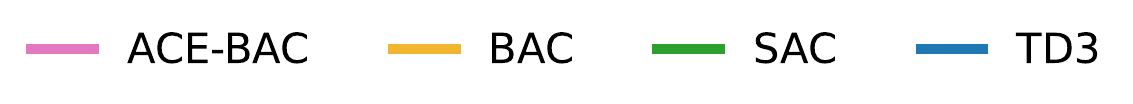}
    \caption{\textbf{Performance of ad-hoc ACE-BAC.} Training curves of ACE-BAC, BAC, SAC, and TD3 in challenging dexterous hand manipulation reward tasks. Solid curves depict the mean of 6 trials, and shaded regions correspond to the one standard deviation. }
    \label{fig:cbac-dexteroushand}
\end{figure}

\begin{figure}[h]
    \centering
    \includegraphics[height=3.8cm,keepaspectratio]{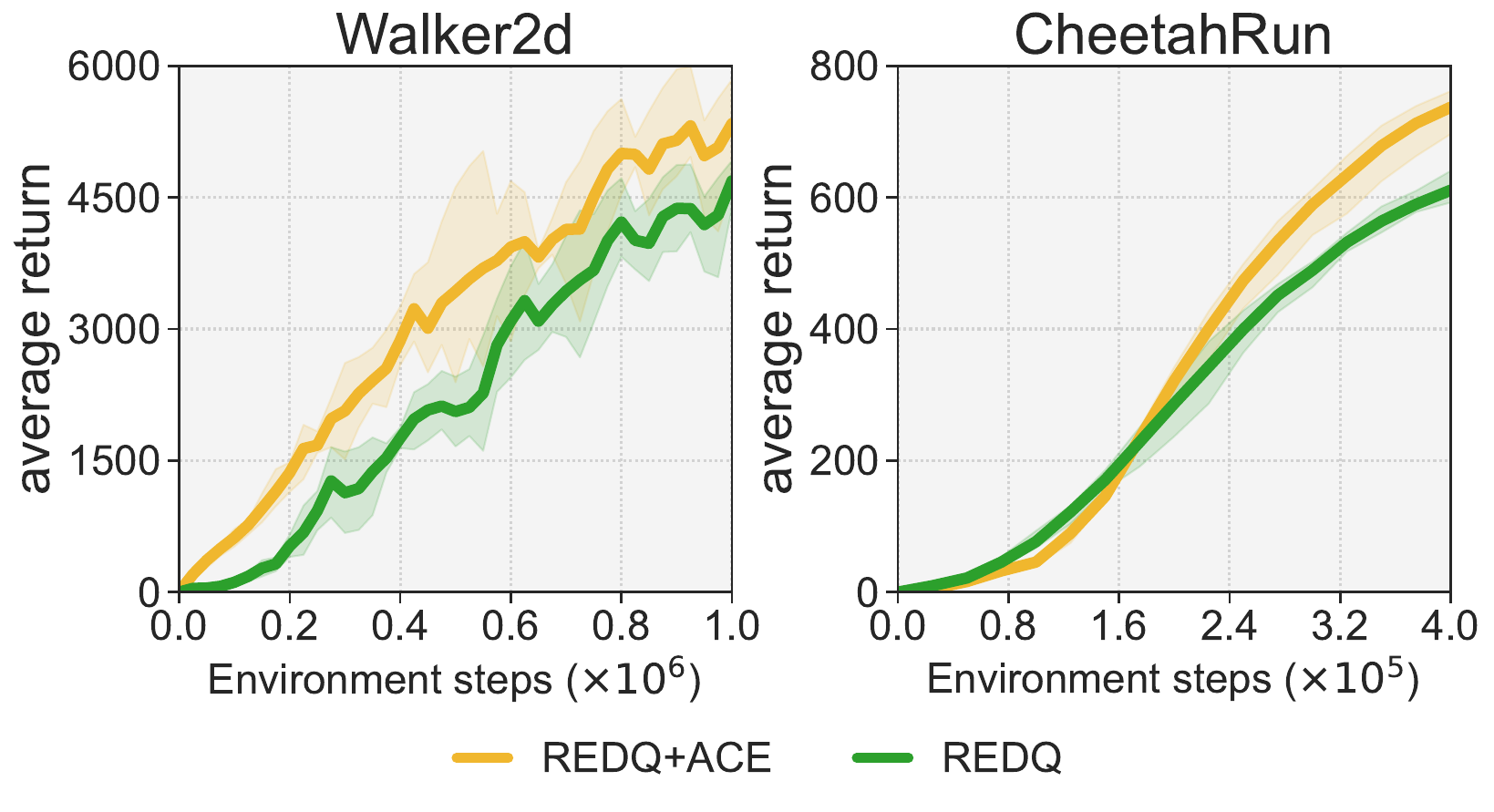} 
    \caption{\textbf{Performance of ACE with higher-UTD methods.} We applied ACE to REDQ by integrating causality-aware entropy and a dormancy-guided reset mechanism. The results show that our method significantly improves REDQ's sample efficiency and performance in multiple tasks. Solid curves depict the mean of 6 trials, and shaded regions correspond to the one standard deviation. }
    \label{fig:ace-higher-utd}
\end{figure}
\begin{table}[h]
\centering
\begin{tabular}{lcccc}
\toprule
 & halfcheetah-medium-v2 & hopper-medium-v2 & walker2d-medium-v2 & maze2d-medium-v1 \\
\midrule
CQL & 47.04 $\pm$ 0.22 & 59.08 $\pm$ 3.77 & 80.75 $\pm$ 3.28 & 86.11 $\pm$ 9.68 \\
CQL+ACE & 50.23 $\pm$ 0.43 & 60.62 $\pm$ 3.97 & 85.27 $\pm$ 3.68 & 89.61 $\pm$ 18.72 \\
\bottomrule
\end{tabular}
\caption{Performance comparison of CQL and CQL+ACE across various tasks.}
\label{table:performance_comparison}
\end{table}

\paragraph{Compatibility with Max-Entropy algorithms.} We integrate \ours into a recent BAC algorithm, which is effective in challenging failure-prone tasks. Results on the challenging dexterous hand manipulation tasks show that the ad-hoc \textbf{ACE-BAC} algorithm could outperform the BAC algorithm, refer to Figure~\ref{fig:cbac-dexteroushand}. 
\paragraph{Compatibility with high-UTD techniques.} Higher-UTD (Update-to-Data) methods are popular and effective in improving the sample efficiency and performance of reinforcement learning algorithms. We integrate \ours into the popular REDQ~\citep{redq} algorithm, which is known for its efficiency in continuous control tasks. Results on the Walker2d and CheetahRun benchmark 
tasks, shown in Figure~\ref{fig:ace-higher-utd} demonstrate that \ours significantly enhances REDQ's sample efficiency and overall performance across a range of tasks.

\paragraph{Compatibility with offline RL setting.} Excitingly, we observed that CQL with an ACE backbone outperforms CQL. This not only underscores the effectiveness and versatility of ACE but also highlights a promising direction for future exploration.

These observations demonstrate the effectiveness and generalizability of our proposed mechanisms and also shed light on the further applications of our mechanisms in enhancing existing RL algorithms.

\section{\ours\ for Visual RL Application}
\label{app:drm}
\begin{figure*}[t]
\vspace{-2mm}
    \begin{minipage}[t]{0.48\textwidth}
        \centering
        \includegraphics[width=7.5cm,keepaspectratio]{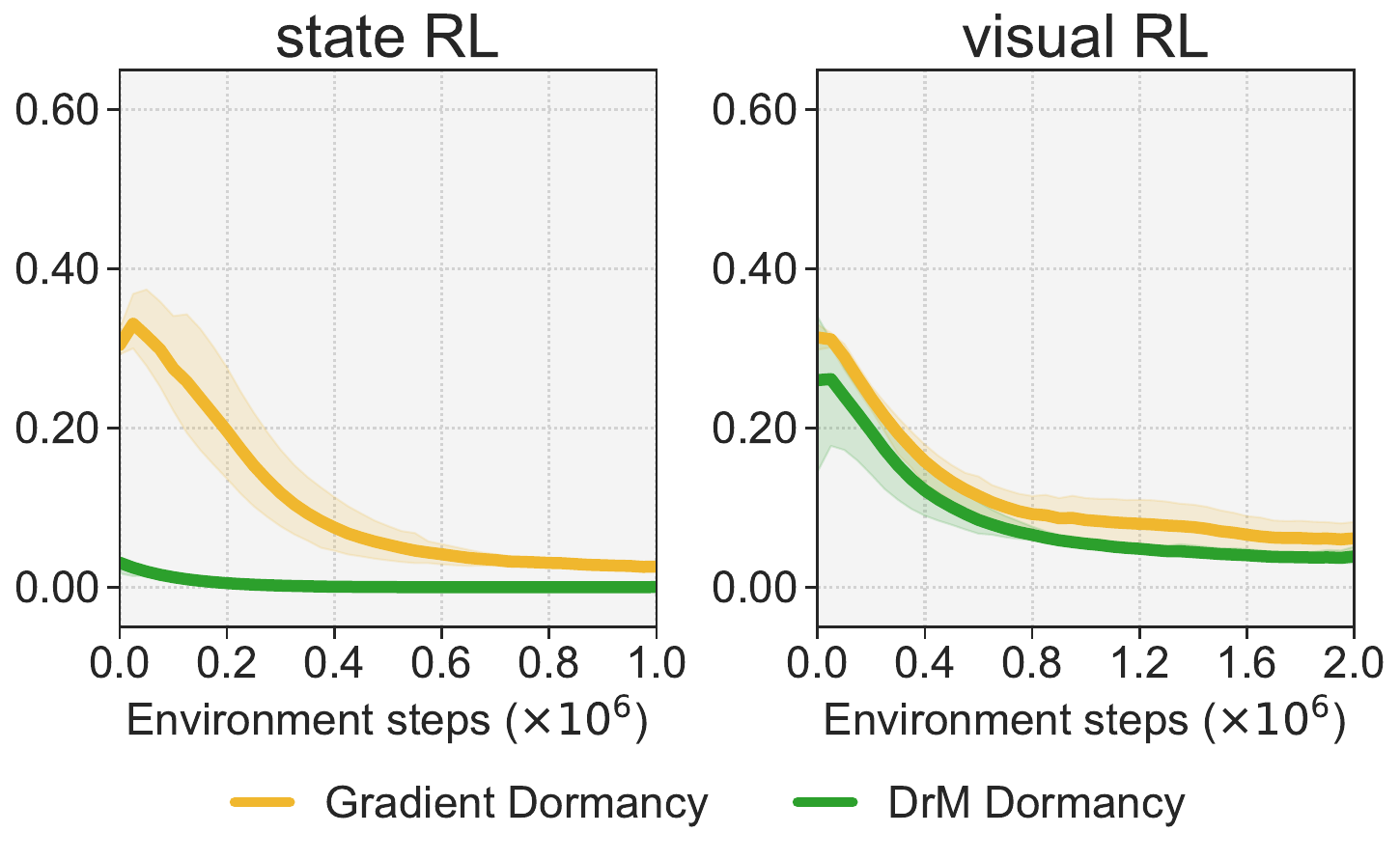}
        \\
        \caption{\small \textbf{Comparison between curves of different dormancy degrees in state-based and visual-based RL.} In state-based RL, the dormant ratio proposed by \cite{drm} consistently approaches zero, while our gradient dormancy degree undergoes notable variations.}
        \vspace{-0.6em}
        \label{fig:dormancy_compare}
    \end{minipage}
    \hfill
    \begin{minipage}[t]{0.48\textwidth}
        \centering
        \includegraphics[width=7.5cm,keepaspectratio]{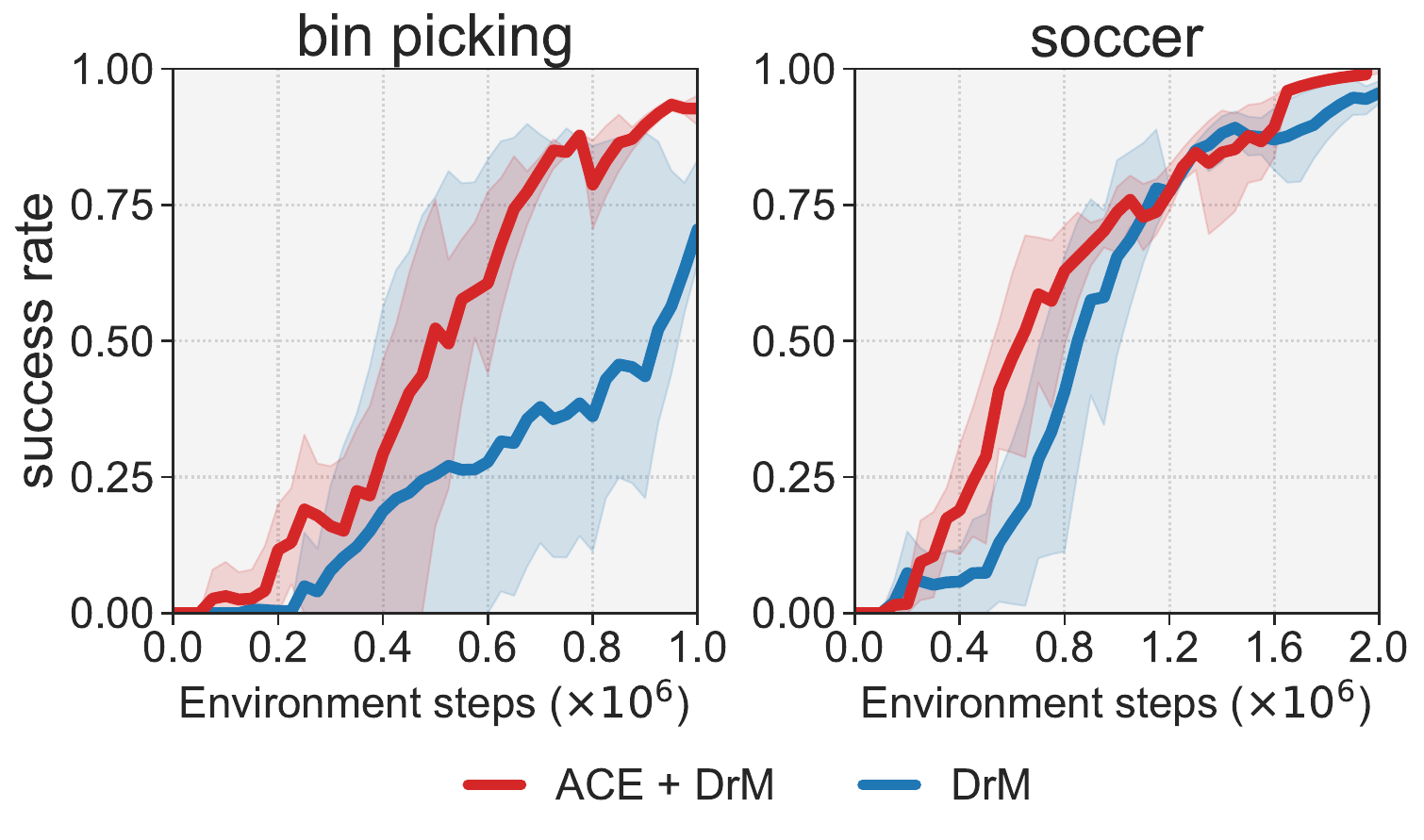}
        \caption{\small \textbf{Performance comparison of ACE + DrM and DrM.} }
        \vspace{-0.5em}
        \label{fig:visual}
    \end{minipage}
\end{figure*}
We extend our method to facilitate visual RL learning with image inputs. The challenges we address in visual RL include: 1) determining how to compute causality between rewards and policy using image inputs, as our causal model cannot directly handle high-dimensional observations. 2) applying our methodology to existing visual RL baselines that do not utilize entropy regularization.

Our proposed solutions are as follows: 1) Utilizing the features extracted from the encoder output for causality computation. Since these features have only 50 dimensions, they can be effectively processed by our causal model. Our experiments demonstrate that causal weights calculated in this manner also provide effective guidance for exploration. 2) Using causal weights to guide action sampling, applying more significant noise to action dimensions with larger causal weights to encourage exploration, and vice versa. Simultaneously, we incorporate our proposed gradient dormancy degree into the exploration schedule. Leveraging these, we make the following improvements to the state-of-the-art visual RL algorithm DrM~\citep{drm}: 1. Replacing the dormant ratio in DrM, which guides perturbation and exploration schedule, with our proposed gradient dormancy degree. 2. Utilizing causal weights to guide action sampling in DrM.

Figure~\ref{fig:dormancy_compare} depicts the dormancy curves using two distinct dormancy definitions, applied to state-based and visual-based soccer task in MetaWorld, respectively, during training with \ours\ and DrM algorithms. It is apparent that in state-based RL, our proposed gradient dormancy degree effectively reflects the network's capacity. In contrast, DrM's dormant ratio, as defined by \citet{dormant}, consistently remains close to zero in state-input tasks and fails to provide effective guidance for state-based RL exploration. In visual RL, the two dormancy curves are closely aligned, effectively guiding exploration and improving the sample efficiency of visual RL. This underscores the adaptability of our proposed gradient dormancy, suitable for any RL scenario.

As shown in Figure~\ref{fig:visual}, our refinements to DrM, leveraging causal weights and the gradient dormancy degree, effectively enhance its performance and learning efficiency. Moving forward, we will continue exploring the potential applications of causality-aware exploration in image-based RL.
\section{The effectiveness of ACE in solving hard exploration problems.}

The effectiveness of ACE is particularly evident in scenarios where the causal relationship between action and reward, conditioned on the state, can be misleading and hurt performance.

\paragraph{Addressing Dummy Actions with the Reset Mechanism.}
For scenarios involving dummy actions, the resetting mechanism helps to solve it. To be specific, such dummy actions would trap the exploration in local optima and hence the gradient dormant would be high (i.e., low policy gradients, high gradient dormancy degree). Then our reset mechanism intervenes by perturbing the network, encouraging unbiased exploration.

A concrete example is the pick-and-place-wall task, where the wall initially biases the agents’ exploration towards vertical and grasping actions. In contrast to the high dormancy degree and poor performance observed with all other baselines, ACE overcomes this challenge with its dormancy-guided reset mechanism, as evidenced by the dormancy degree curves in Figure~\ref{fig: dormancy}  (y-axis=dormancy degree) and the performance data in Figure~\ref{fig:metaworld-manipulation}.

\paragraph{Handling Misleading or Imperfect Rewards.}
For scenarios with misleading or imperfect rewards, we could simply turn it into a sparse reward setting. Our experiments, as shown in Figures~\ref{fig:sparsereward_performance} and ~\ref{fig:sparse_reward}, reveal ACE's effectiveness in sparse reward environments. This advantage primarily arises from ACE's causal-aware entropy, which discerns the implicit action-reward relationship based on the state.

Furthermore, ACE's versatility also allows it to be combined with alternative intrinsic reward approaches, enhancing its applicability across various scenarios. This adaptability ensures that specific challenges, such as those involving misleading rewards.
\section{Computing Infrastructure and Computational Time}
\label{app:efficiency}
Our experiments were conducted on a server equipped with an \texttt{AMD EPYC 7763 64-Core Processor (256 threads)} and four \texttt{NVIDIA GeForce RTX 3090 GPUs}.

\begin{figure}[h]
    \centering
    \includegraphics[width=0.85\linewidth]{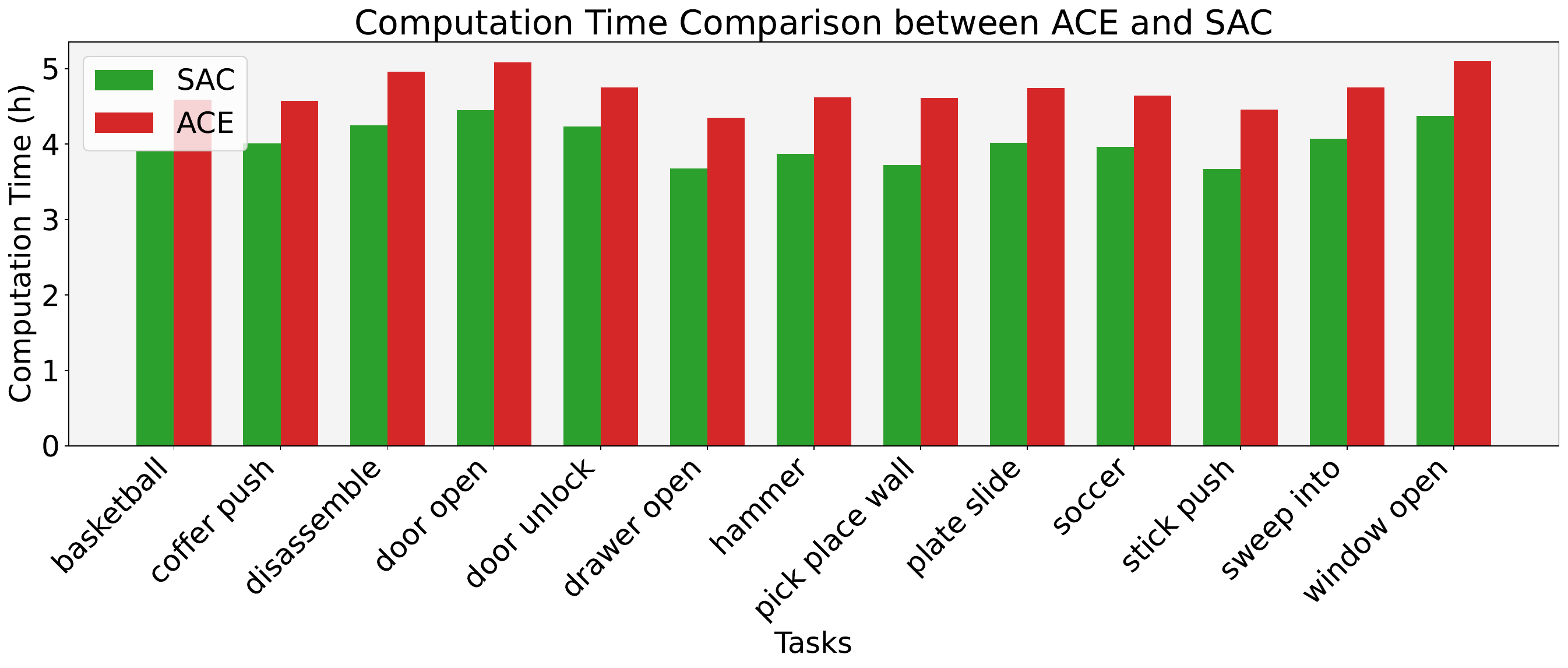}
    \caption{\textbf{Computation time comparison.} Computation time comparison between \ourshort\ and SAC in ten MetaWorld tasks, each averaged on 6 trials.}
    \label{fig:computation time}
\end{figure}

Figure~\ref{fig:computation time} presents the computational time comparison between our algorithm \ourshort\ and SAC on 12 MetaWorld benchmark tasks. Compared to SAC, the total training time of \ourshort\ only increased by an average of 0.69 hours, hence, the additional costs are acceptable. Further, for practical use, \ourshort\ requires fewer interactions for similar performance, which may lower the needed computation time in practice.

\vspace{-0.5em}
% % \section{You \emph{can} have an appendix here.}

% You can have as much text here as you want. The main body must be at most $8$ pages long.
% For the final version, one more page can be added.
% If you want, you can use an appendix like this one.  

% The $\mathtt{\backslash onecolumn}$ command above can be kept in place if you prefer a one-column appendix, or can be removed if you prefer a two-column appendix. Apart from this possible change, the style (font size, spacing, margins, page numbering, etc.) should be kept the same as the main body.
%%%%%%%%%%%%%%%%%%%%%%%%%%%%%%%%%%%%%%%%%%%%%%%%%%%%%%%%%%%%%%%%%%%%%%%%%%%%%%%
%%%%%%%%%%%%%%%%%%%%%%%%%%%%%%%%%%%%%%%%%%%%%%%%%%%%%%%%%%%%%%%%%%%%%%%%%%%%%%%

\end{document}